\PassOptionsToPackage{unicode}{hyperref}
\PassOptionsToPackage{hyphens}{url}
\PassOptionsToPackage{dvipsnames,svgnames,x11names}{xcolor}
\documentclass[
  12pt,
  dvipsnames,enabledeprecatedfontcommands]{scrartcl}
\usepackage{amsmath,amssymb}
\usepackage{lmodern}
\usepackage{iftex}
\ifPDFTeX
  \usepackage[T1]{fontenc}
  \usepackage[utf8]{inputenc}
  \usepackage{textcomp} 
\else 
  \usepackage{unicode-math}
  \defaultfontfeatures{Scale=MatchLowercase}
  \defaultfontfeatures[\rmfamily]{Ligatures=TeX,Scale=1}
\fi
\IfFileExists{upquote.sty}{\usepackage{upquote}}{}
\IfFileExists{microtype.sty}{
  \usepackage[]{microtype}
  \UseMicrotypeSet[protrusion]{basicmath} 
}{}
\makeatletter
\@ifundefined{KOMAClassName}{
  \IfFileExists{parskip.sty}{%
    \usepackage{parskip}
  }{
    \setlength{\parindent}{0pt}
    \setlength{\parskip}{6pt plus 2pt minus 1pt}}
}{
  \KOMAoptions{parskip=half}}
\makeatother
\usepackage{xcolor}
\usepackage{graphicx}
\makeatletter
\def\maxwidth{\ifdim\Gin@nat@width>\linewidth\linewidth\else\Gin@nat@width\fi}
\def\maxheight{\ifdim\Gin@nat@height>\textheight\textheight\else\Gin@nat@height\fi}
\makeatother
\setkeys{Gin}{width=\maxwidth,height=\maxheight,keepaspectratio}
\makeatletter
\def\fps@figure{htbp}
\makeatother
\providecommand{\tightlist}{%
  \setlength{\itemsep}{0pt}\setlength{\parskip}{0pt}}
\newlength{\cslhangindent}
\newlength{\csllabelwidth}
\newlength{\cslentryspacingunit} 
\newenvironment{CSLReferences}[2] 
 {
  \setlength{\parindent}{0pt}
  \ifodd #1
  \let\oldpar\par
  \def\par{\hangindent=\cslhangindent\oldpar}
  \fi
  \setlength{\parskip}{#2\cslentryspacingunit}
 }%
 {}
\usepackage{calc}

\newcommand{\CSLLeftMargin}[1]{\parbox[t]{\csllabelwidth}{#1}}
\newcommand{\CSLRightInline}[1]{\parbox[t]{\linewidth - \csllabelwidth}{#1}\break}

\usepackage{booktabs}
\usepackage{multirow}
\usepackage{subcaption}

\usepackage{graphicx}
\usepackage{supertabular, booktabs}
\usepackage{ragged2e}
\usepackage{etex}
\usepackage{marvosym}
\usepackage[notextcomp]{kpfonts}
\usepackage{nicefrac}

\usepackage{floatrow}

\usepackage{mathptmx}
\usepackage[scaled=0.86]{helvet}
\usepackage{t1enc}

\usepackage{amsfonts}
\usepackage{amssymb}
\usepackage{amsthm}
\usepackage{amsmath}
\usepackage{mathtools}

\usepackage{geometry}
\newtheorem*{reply*}{Reply}
\usepackage{enumitem}

\usepackage{float}

\usepackage[authoryear]{natbib}

\usepackage{tikz}
\usetikzlibrary{positioning,shapes,arrows}

\usepackage{booktabs}
\usepackage{longtable}
\usepackage{array}
\usepackage{multirow}
\usepackage{wrapfig}
\usepackage{float}
\usepackage{colortbl}
\usepackage{pdflscape}
\usepackage{tabu}
\usepackage{threeparttable}
\usepackage{threeparttablex}
\usepackage[normalem]{ulem}
\usepackage{makecell}
\usepackage{xcolor}
\ifLuaTeX
  \usepackage{selnolig}  
\fi
\IfFileExists{bookmark.sty}{\usepackage{bookmark}}{\usepackage{hyperref}}
\IfFileExists{xurl.sty}{\usepackage{xurl}}{} 
\hypersetup{
  pdftitle={A Bayesian approach to uncertainty in word embedding bias estimation},
  pdfauthor={Alicja Dobrzeniecka and Rafal Urbaniak},
  colorlinks=true,
  linkcolor={Maroon},
  filecolor={Maroon},
  citecolor={Blue},
  urlcolor={blue},
  pdfcreator={LaTeX via pandoc}}

\title{A Bayesian approach to uncertainty in word embedding bias
estimation}
\author{\normalsize Alicja Dobrzeniecka \\[-2mm]   \footnotesize Vrije Universiteit Amsterdam, The Netherlands \\[1mm]  \normalsize Rafal Urbaniak \\[-2mm]
\footnotesize Basis Research Institute, NYC, USA \\[-3mm]
\footnotesize The University of Gdansk,  Poland}
\date{}

\begin{document}
\maketitle

\textbf{Abstract.} Multiple measures, such as WEAT or MAC, attempt to
quantify the magnitude of bias present in word embeddings in terms of a
single-number metric. However, such metrics and the related statistical
significance calculations rely on treating pre-averaged data as
individual data points and employing bootstrapping techniques with low
sample sizes. We show that similar results can be easily obtained using
such methods even if the data are generated by a null model lacking the
intended bias. Consequently, we argue that this approach generates false
confidence. To address this issue, we propose a Bayesian alternative:
hierarchical Bayesian modeling, which enables a more
uncertainty-sensitive inspection of bias in word embeddings at different
levels of granularity. To showcase our method, we apply it to
\emph{Religion, Gender}, and \emph{Race} word lists from the original
research, together with our control neutral word lists. We deploy the
method using \emph{Google, Glove}, and \emph{Reddit} embeddings.
Further, we utilize our approach to evaluate a debiasing technique
applied to the \emph{Reddit} word embedding. Our findings reveal a more
complex landscape than suggested by the proponents of single-number
metrics. The datasets and source code for the paper are publicly
available.\footnote{\url{https://github.com/efemeryds/Bayesian-analysis-for-NLP-bias}}

\hypertarget{introduction}{%
\section{Introduction}\label{introduction}}

It has been suggested\footnote{See for instance {[}1,4,5,9,12,13{]}.}
that language models can learn implicit biases that reflect harmful
stereotypical thinking---for example, the (vector corresponding to the)
word \textit{she} might be much closer in the vector space to the word
\textit{cooking} than the word \textit{he}. Such phenomena are
undesirable at least in some downstream tasks, such as web search,
recommendations, and so on. To investigate such issues, several measures
of bias in word embeddings have been formulated and applied. Our goal is
to use two prominent examples of such measures to argue that this
approach oversimplifies the situation and to develop a Bayesian
alternative.

A common approach in natural language processing is to represent words
by vectors of real numbers---such representations are called
\emph{embeddings}. One way to construct an embedding---we will focus our
attention on non-contextual language models\footnote{One example of a
  contextualized representation is BERT. Another is GPT.}---is to use a
large corpus to train a neural network to assign vectors to words in a
way that optimizes for co-occurrence prediction accuracy. Such vectors
can then be compared in terms of their similarity---the usual measure is
cosine similarity---and the results of such comparisons can be used in
downstream tasks. Roughly speaking, cosine similarity is an imperfect
mathematical proxy for semantic similarity {[}16{]}.

One response to the raising of the issue of bias in natural language
models might be to say that there is not much point in reflecting on
such biases, as they are unavoidable. This unavoidability might seem in
line with the arguments to the effect that learning algorithms are
always value-laden {[}10{]}: they employ inductive methods that require
design-, data-, or risk-related decisions that have to be guided by
extra-algorithmic considerations. Such choices necessarily involve value
judgments and have to do, for instance, with what simplifications or
risks one finds acceptable. Admittedly, algorithmic decision making
cannot fulfill the value-free ideal, but this only means that even more
attention needs to be paid to the values underlying different techniques
and decisions, and to the values being pursued in a particular use of an
algorithm.

Another response might be to insist that there is no bias introduced by
the use of machine learning methods here since the algorithm is simply
learning to correctly predict co-occurrences based on what ``reality''
looks like. However, this objection overlooks the fact that we, humans,
are the ones who construct this linguistic reality, which is shaped in
part by the natural language processing tools we use on a massive scale.
Sure, if there is unfairness and our goal is to diagnose it, we should
do complete justice to learning it in the model used to study it. One
example of this approach is {[}3{]}, where the authors use language
models to study the shape of certain biases across a century.

However, if our goal is to develop downstream tools that perform tasks
that we care about without further perpetuating or exacerbating harmful
stereotypes, we still have good reasons to try to minimize the negative
impact. Moreover, it is often not the case that the corpora mirror
reality---to give a trivial example, heads are spoken of more often than
kidneys, but this does not mean that kidneys occur much less often in
reality than heads. To give a more relevant example, the
disproportionate association of female words with female occupations in
a corpus actually greatly exaggerates the actual lower disproportion in
the real distribution of occupations {[}6{]}.

In what follows, we focus on two popular measures of bias applicable to
many existing word embeddings, such as
\emph{GoogleNews},\footnote{GoogleNews-vectors-negative300, available at  \url{https://github.com/mmihaltz/word2vec-GoogleNews-vectors}.}
\emph{Glove}\footnote{Available at \url{https://nlp.stanford.edu/projects/glove/}.}
and
\emph{Reddit Corpus}\footnote{Reddit-L2 corpus, available at  \url{http://cl.haifa.ac.il/projects/L2/}.}:
\emph{Word Embedding Association Test} (\textsf{WEAT}) {[}9{]}, and
\emph{Mean Average Cosine Distance} (\textsf{MAC}) {[}13{]}. We first
explain how these measures are supposed to work. Then we argue that they
are problematic for various reasons---the key one being that by
pre-averaging data they manufacture false confidence, which we
illustrate in terms of simulations showing that the measures often
suggest the existence of bias even if by design it is non-existent in a
simulated dataset.

We propose to replace them with a Bayesian data analysis, which not only
provides more modest and realistic assessment of the uncertainty
involved, but in which hierarchical models allow for inspection at
various levels of granularity. Once we introduce the method, we apply it
to multiple word embeddings and results of supposed debiasing, putting
forward some general observations that are not exactly in line with the
usual picture painted in terms of \textsf{WEAT} or \textsf{MAC}.

Most of the problems that we point out generalize to any existing
approach that focuses on chasing a single numeric metric of bias. (1)
They treat the results of pre-averaging as raw data in statistical
significance tests, which in this context is bound to overestimate
significance. We show similar results can easily be obtained when
sampling from null models with no bias. (2) The word list sizes and
sample sizes used in the studies are usually small,\footnote{Depending
  on a list for {[}9{]} the range for protected words is between 13 and
  100, and for attributes between 16 and 25; for {[}13{]} the range for
  protected words is between 14 and 18, and for attributes between 11
  and 25.} (3) Many studies do not use any control predicates, such as
random neutral words or neutral human predicates for comparison.

On the constructive side, we develop and deploy our method, and the
results are, roughly, as follows. (A) Posterior density intervals are
fairly wide and the average differences between associated, different
and neutral human predicates, are not very large. (B) A preliminary
inspection suggests that the desirability of changes obtained by the
usual debiasing methods is debatable.

In Section \ref{sec:two} we describe the two key measures discussed in
this paper, \textsf{WEAT} and \textsf{MAC}, explaining how they are
calculated and how they are supposed to work. In Section
\ref{sec:challenges} we first argue in Subsection
\ref{subsec:interpretability}, that it is far from clear how results
given in terms of \textsf{WEAT} or \textsf{MAC} are to be interpreted.
Second, in Subsection \ref{subsec:problems} we explain the statistical
problems that arise when one uses pre-averaged data in such contexts, as
these measures do. In Section \ref{sec:bayesian} we explain the
alternative Bayesian approach that we propose. In Section
\ref{sec:results} we elaborate on the results that it leads to,
including a somewhat skeptical view of the efficiency of debiasing
methods, discussed in Subsection \ref{subsec:rethinking}. Finally, in
Section \ref{sec:related} we spend some time placing our results in the
ongoing discussions.\footnote{\textbf{Disclaimer:} throughout the paper
  we will be mentioning and using word lists and stereotypes we did not
  formulate, which does not mean we condone any judgment made therein or
  underlying a given word selection. For instance, the Gender dataset
  does not recognize non-binary categories, and yet we use it without
  claiming that such categories should be ignored.}

\hypertarget{two-measures-of-bias-weat-and-mac}{%
\section{Two measures of bias: WEAT and
MAC}\label{two-measures-of-bias-weat-and-mac}}

\label{sec:two}

The underlying intuition is that if a particular harmful stereotype is
learned in a particular embedding, then certain groups of words will be
systematically closer to (or further from) each other. This gives rise
to the idea of protected groups---for example, in guiding online search
completion or recommendation, female words might require protection in
that they should not be systematically closer to stereotypically female
job names, such as ``nurse'', ``librarian'', ``waitress'', and male
words require protection in that they should not be systematically
closer to toxic masculinity stereotypes, such as ``tough'', ``never
complaining'' or ``macho''.\footnote{However, for some research-related
  purposes, such as the study of stereotypes across history {[}3{]},
  embeddings which do not protect certain classes may also be useful.}

The key role in the measures to be discussed is played by the notion of
cosine distance (or, symmetrically, by cosine similarity). These are
defined as follows:\footnote{Here, ``\(-\)'' stands for point-wise
  difference, ``\(\cdot\)'' stands for the dot product operation, and
  \(\lVert a\rVert = \sqrt{(a \cdot a)}\).}
\(^{\!\!\! , \,}\)\footnote{Note that this terminology is slightly
  misleading, as mathematically cosine distance is not a distance
  measure, because it does not satisfy the triangle inequality, as
  generally
  \(\mathsf{cosineDistance}(A,C) \not \leq \mathsf{cosineDistance}(A,B)+ \mathsf{cosineDistance}(B,C)\).
  We will keep using this mainstream terminology.}
\begin{align} \tag{Sim}
\mathsf{cosineSimilarity}(A,B) & = \frac{A \cdot B}{\lVert  A \rVert \,\lVert B \rVert}
\\
\tag{Distance}
\mathsf{cosineDistance}(A,B) &  = 1 - \mathsf{cosineSimilarity}(A,B).
\end{align}

One of the first measures of bias has been developed in {[}1{]}. The
general idea is that a certain topic is associated with a vector of real
numbers (the topic ``direction''), and the bias of a word is
investigated by considering the projection of its corresponding vector
on this direction. For instance, in {[}1{]}, the gender direction
\(\mathsf{gd}\) is obtained by taking the differences of the vectors
corresponding to ten different gendered pairs (such as
\(\overrightarrow{she} - \overrightarrow{he}\) or
\(\overrightarrow{girl} - \overrightarrow{boy}\)), and then identifying
their principal component.\footnote{Roughly, the principal component is
  the vector obtained by projecting the data points on their linear
  combination in a way that maximizes the variance of the projections.}
The gender bias of a word \(w\) is then understood as \(w\)'s projection
on the gender direction: \(\vec{w} \cdot gd\) (which, after normalizing
by dividing by \(\lVert w \rVert \,\lVert \mathsf{gd} \rVert\), is the
same as cosine similarity). Given a list \(N\) of supposedly gender
neutral words,\footnote{We follow the methodology used in the debate in
  assuming that there is a class of words identified as more or less
  neutral, such as \emph{ballpark, eat, walk, sleep, table}, whose
  average similarity to the gender direction (or other protected words)
  is around 0. See our list in Appendix \ref{app:custom} and a brief
  discussion in Subsection \ref{subsec:interpretability}.} and the
gender direction \(\mathsf{gd}\), the direct gender bias is defined as
the average cosine similarity of the words in \(N\) from \(\mathsf{gd}\)
(\(c\) is a parameter determining how strict we want to be):
\begin{align*}
\mathsf{directBias_c(N,gd)} & = \frac{\sum_{w\in N}\vert \mathsf{cos}(\vec{w},\mathsf{gd})\vert^c}{\vert N \vert }
\end{align*} \normalsize 

The use of projections in bias estimation has been criticized for
instance in {[}5{]}, where it is pointed out that while a higher average
similarity to the gender direction might be an indicator of bias with
respect to a given class of words, it is only one possible manifestation
of it, and reducing the cosine similarity to such a projection may not
be sufficient to eliminate bias. For instance, ``math'' and ``delicate''
might be equally similar to a pair of opposed explicitly gendered words
(\emph{she}, \emph{he}), while being closer to quite different
stereotypical attribute words (such as \emph{scientific} or
\emph{caring}). Further, it is observed in {[}5{]} that most word pairs
retain similarity under debiasing meant to minimize projection-based
bias.\footnote{In {[}1{]} another method that involves analogies and
  their evaluations by human users on Mechanical Turk is also used. We
  do not discuss this method in this paper, see its criticism in
  {[}18{]}.}

A measure of bias in word embeddings that does not proceed by
identifying bias directions (such as a gender vector), the Word
Embedding Association Test (\textsf{WEAT}), has been proposed in
{[}9{]}. The idea here is that the bias between two sets of target
words, \(X\) and \(Y\) (we call them protected words), should be
quantified in terms of the cosine similarity between the protected words
and attribute words coming from two sets of stereotype attribute words,
\(A\) and \(B\) (we will call them attributes). For instance, \(X\)
might be a set of male names, \(Y\) a set of female names, \(A\) might
contain stereotypically male-related, and \(B\) stereotypically
female-related career words. The association difference for a particular
word \(t\) (belonging to either \(X\) or \(Y\)) is:

\vspace{-2mm}

\begin{align}
\label{eq:stAB}
\mathsf{s}(t,A,B) & = \frac{\sum_{a\in A}\mathsf{cos}(t,a)}{\vert A\vert} - \frac{\sum_{b\in B}\mathsf{cos}(t,b)}{\vert B\vert}
\end{align} \normalsize \noindent then, the association difference
between \(A\) a \(B\) is:

\begin{align}
\label{eq:sXYAB}
\mathsf{s}(X,Y,A,B) & = \sum_{x\in X} \mathsf{s}(x,A,B) -  \sum_{y\in Y} \mathsf{s}(y,A,B)
\end{align}

\noindent The intention is that large values of \(s\) scores suggest
systematic differences between how \(X\) and \(Y\) are related to \(A\)
and \(B\), and therefore are indicative of the presence of bias. The
authors use it as a test statistic in some tests,\footnote{Note their
  method assumes \(X\) and \(Y\) are of the same size.} and the final
measure of effect size, \textsf{WEAT}, is constructed by taking means of
these values and standardizing: \begin{align} \label{eq:weat}
\mathsf{WEAT}(A,B) & = \frac{
\mu(\{\mathsf{s}(x,A,B)\}_{x\in X}) -\mu(\{\mathsf{s}(y,A,B)\}_{y\in Y}) 
}{
\sigma(\{\mathsf{s}(w,A,B)\}_{w\in X\cup Y})
}
\end{align}

\textsf{WEAT} is inspired by the Implicit Association Test (IAT)
{[}19{]} used in psychology, and in some applications it uses almost the
same word sets, allowing for a \emph{prima facie} sensible comparison
with bias in humans. In {[}9{]} the authors argue that significant
biases---thus measured--- similar to the ones discovered by IAT can be
discovered in word embeddings. In {[}12{]} the methodology is extended
to a multilingual and cross-lingual setting, arguing that using
Euclidean distance instead of cosine similarity does not make much
difference, while the bias effects vary greatly across embedding
models.\footnote{Interestingly, with social media-text trained
  embeddings being less biased than those based on Wikipedia.} A similar
methodology is employed in {[}4{]}. The authors employ word embeddings
trained on corpora from different decades to study the shifts in various
biases through the century.\footnote{Strictly speaking, these authors
  use Euclidean distances and their differences, but the way they take
  averages and averages thereof is analogous, and so what we will have
  to say about pre-averaging leading to false confidence applies to this
  methodology as well.}

\begin{figure}[H]

\begin{center}\includegraphics[width=0.55\linewidth]{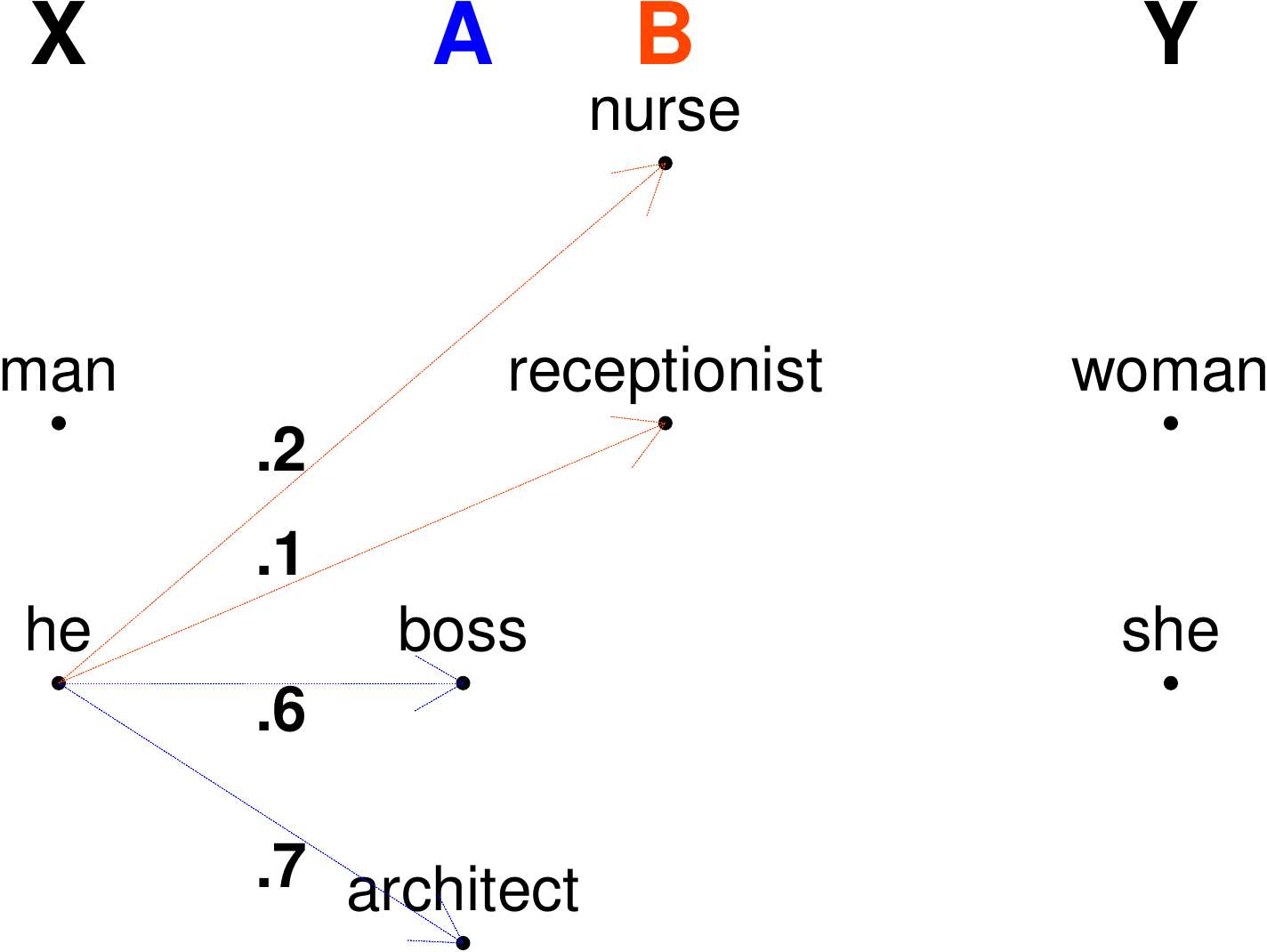} \end{center}
\caption{A simple example of a two-class set-up. Two groups of protected words, $X$ and $Y$ with two stereotypical attribute sets. An example of \textsf{WEAT} calculations follows.}
\label{eq:WEATexample}
\end{figure}

\noindent Here is an example of \textsf{WEAT} calculations for Figure
\ref{eq:WEATexample}:
\begin{align*} s_1 & = s(he,A,B)  =  \nicefrac{(.6+.7)}{2}  - \nicefrac{(.2+.1)}{2} = .65-.15= .5 \\
s_2  & = s(man,A,B) = .3 \\
s_3  & = s(woman,A,B) = -.6\\
s_4 & = s(she, A, B) = -.3\\
\mathsf{WEAT}(A,B)  & = \frac{\nicefrac{(s_1+s_2)}{2} - \nicefrac{(s_3+s_4)}{2}}{sd(\{s_1,s_2,s_3,s_4\})} \approx 1.93
\end{align*}

\textsf{WEAT} has been developed to investigate biases corresponding to
a pair of supposedly opposing stereotypes, so the question arises as to
how to generalize the measure to contexts in which biases with respect
to more than two stereotypical groups are to be measured. Such a
generalization can be found in {[}13{]}. The authors introduce Mean
Average Cosine distance (\textsf{MAC}) as a measure of bias. Let
\(T = \{t_1, \dots, t_k\}\) be a set of protected words, and let each
\(A_j\in \mathcal{A}\) be a set of attributes stereotypically associated
with a protected word where \(\mathcal{A}\). For instance, when biases
related to religion are to be investigated, they use a dataset of the
format illustrated in Table \ref{tab:religionOriginal}. The measure is
defined as follows:

\begin{table}
\footnotesize

\centering

\begin{tabular}[t]{lllr}
\toprule
protected words ($T$) & attributes & attribute set ($A_j$) & cosine distance\\
\midrule
\cellcolor{gray!15}{rabbi} & \cellcolor{gray!15}{greedy} & \cellcolor{gray!15}{jewStereotype} & \cellcolor{gray!15}{1.03}\\
church & familial & christianStereotype & 0.70\\
\cellcolor{gray!15}{synagogue} & \cellcolor{gray!15}{liberal} & \cellcolor{gray!15}{jewStereotype} & \cellcolor{gray!15}{0.79}\\
jew & familial & christianStereotype & 0.98\\
\cellcolor{gray!15}{quran} & \cellcolor{gray!15}{dirty} & \cellcolor{gray!15}{muslimStereotype} & \cellcolor{gray!15}{1.12}\\
muslim & uneducated & muslimStereotype & 0.52\\
\cellcolor{gray!15}{torah} & \cellcolor{gray!15}{terrorist} & \cellcolor{gray!15}{muslimStereotype} & \cellcolor{gray!15}{0.93}\\
quran & hairy & jewStereotype & 1.18\\
\cellcolor{gray!15}{synagogue} & \cellcolor{gray!15}{violent} & \cellcolor{gray!15}{muslimStereotype} & \cellcolor{gray!15}{0.95}\\
bible & cheap & jewStereotype & 1.22\\
\cellcolor{gray!15}{christianity} & \cellcolor{gray!15}{greedy} & \cellcolor{gray!15}{jewStereotype} & \cellcolor{gray!15}{0.97}\\
muslim & hairy & jewStereotype & 0.88\\
\cellcolor{gray!15}{islam} & \cellcolor{gray!15}{critical} & \cellcolor{gray!15}{christianStereotype} & \cellcolor{gray!15}{0.79}\\
muslim & conservative & christianStereotype & 0.45\\
\cellcolor{gray!15}{mosque} & \cellcolor{gray!15}{greedy} & \cellcolor{gray!15}{jewStereotype} & \cellcolor{gray!15}{1.15}\\
\bottomrule
\end{tabular}

\caption{Sample 15 rows of the religion dataset. The whole dataset has 15 unique protected words ($T$), and 11 unique attributes divided between 3 attribute sets ($A_1=\mathsf{jewStereotype}, A_2= \mathsf{christianStereotype}, A_3=\mathsf{muslimStereotype}$). $\mathcal{A}$ consists of these three sets, $\mathcal{A}= \{A_1, A_2, A_3\}$. The whole dataset has $15\times 11 = 165$ rows.}
\label{tab:religionOriginal}
\normalsize 
\end{table}

\begin{align*}
\mathsf{s}(t, A_j) & = \frac{1}{\vert A_j\vert}\sum_{a\in A_j}\mathsf{cosineDistance}(t,a) \\
\mathsf{MAC}(T,\mathcal{A}) & = \frac{1}{\vert T \vert \,\vert \mathcal{A}\vert}\sum_{t \in T }\sum_{A_j \in \mathcal{A}}    \mathsf{s}(t,A_j)
\end{align*}

\noindent That is, for each protected word \(t\in T\), and each
attribute set \(A_j\), they first take the mean of distances for this
protected word and all attributes in a given attribute class, and then
take the mean of thus obtained means for all the protected words and all
the protected classes.\footnote{The authors' code is available through
  their github repository at
  \url{https://github.com/TManzini/DebiasMulticlassWordEmbedding}.}

\begin{figure}[H]

\begin{center}\includegraphics[width=0.5\linewidth]{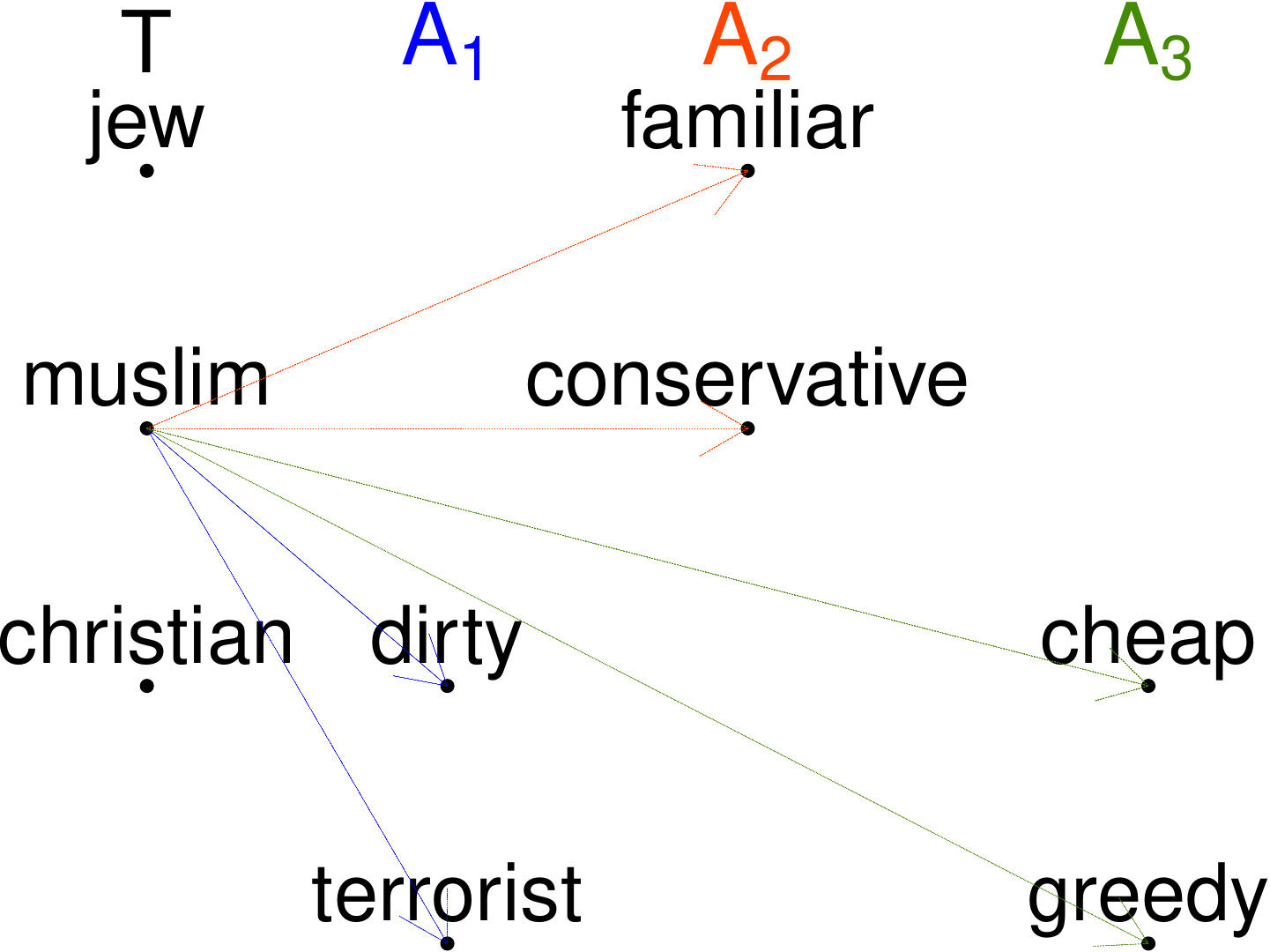} \end{center}
\caption{A small subset of the religion dataset.  To each protected word in $T$  there corresponds one class of stereotypical attributes typically associated with it (and other classes of stereotypical attributes associated with different protected words).}
\label{fig:MACexample}
\end{figure}

\noindent An example of \textsf{MAC} calculations for the situation
depicted in Figure \ref{fig:MACexample} is as follows: \begin{align*}
s_1   = s(muslim,A_1)  & = \frac{\mathsf{cos}(muslim,dirty)+\mathsf{cos}(muslim,terrorist)}{2} \\ 
s_2   = s(muslim,A_2)  & = \frac{\mathsf{cos}(muslim,familiar)+\mathsf{cos}(muslim,conservative)}{2}\\ & \vdots \\ 
\mathsf{MAC}(T,A)  & = \mathsf{mean}(\{s_i \vert i \in 1, \dots, k\})
\end{align*}

Notably, the intuitive distinction between different attribute sets
plays no real role in the \textsf{MAC} calculations. Equally well one
could calculate the mean distance of \emph{muslim} to all the
predicates, mean distance of \emph{christian} to all the predicates,
the mean distance of \emph{jew} to all the predicates, and then to take the
mean of these three means.

Having introduced the measures, first, we will introduce a selection of
general problems with this approach, and then we will move on to more
specific but important problems related to the fact that the measures
take averages and averages of averages. Once this is done, we move to
the development of our Bayesian alternative and the presentation of its
deployment.

\hypertarget{challenges-to-cosine-based-bias-metrics}{%
\section{Challenges to cosine-based bias
metrics}\label{challenges-to-cosine-based-bias-metrics}}

\label{sec:challenges}

\hypertarget{interpretability-issues}{%
\subsection{Interpretability issues}\label{interpretability-issues}}

\label{subsec:interpretability}

Table \ref{tab:religionOriginal2} contains an example of \textsf{MAC}
scores (and \(p\) values, we explain how these are obtained in
Subsection \ref{subsec:problems}) before and after deploying two
debiasing methods to the Reddit embedding, where the score is calculated
using the Religion word lists from {[}13{]}. For our purpose the details
of the debiasing method are not important: what matters is that
\textsf{MAC} is used in the evaluation of these methods.

\begin{table}[H]
\footnotesize

\centering

\begin{tabular}[t]{lllr}
\toprule
Religion Debiasing & \textsf{MAC} & a p-value \\
\midrule
\cellcolor{gray!15}{Biased} & \cellcolor{gray!15}{0.859} & \cellcolor{gray!15}{N/A} \\
Hard Debiased & 0.934 & 3.006e-07\\
\cellcolor{gray!15}{Soft Debiased ($\lambda$ = 0.2)} & \cellcolor{gray!15}{0.894} & \cellcolor{gray!15}{0.007} \\
\bottomrule
\end{tabular}

\caption{The associated mean average cosine similarity
(\textsf{MAC}) and p-values for debiasing methods for religious bias.}
\label{tab:religionOriginal2}
\normalsize 
\end{table}

The first question we should ask is whether the initial \textsf{MAC}
values lower than 1 indeed are indicative of the presence of bias?
Thinking abstractly, 1 is the ideal distance for unrelated words. But in
fact, there is some variation in distances, which might lead to
non-biased lists also having \textsf{MAC} scores smaller than 1. How
much smaller? What may attract attention is the fact that the value of
cosine distance in ``Biased'' category is already quite high (i.e.~close
to 1) even before debiasing. High cosine distance indicates low cosine
similarity between values. One could think that the average cosine
similarity equal to approximately 0.141 is not large enough to claim the
presence of a bias to start with. The authors, though, still aim to
mitigate it by making the distances involved in the \textsf{MAC}
calculations even larger. The question is, on what basis is this small
similarity still considered as proof of the presence of bias, and
whether these small changes are meaningful.

The problem is that the original paper did not employ any control group
of neutral attributes for comparison to obtain a more realistic gauge on
how to understand \textsf{MAC} values. Later on, in our approach, we
introduce such control word lists. One of them is a list of words we
intuitively considered neutral. Moreover, it might be the case that
words that have to do with human activities in general, even if
unbiased, are systematically closer to the protected words than merely
neutral words. This, again, casts doubt on whether comparing
\textsf{MAC} to the abstractly ideal value of 1 is a methodologically
sound idea. For this reason, we also use a second list with intuitively
non-stereotypical human attributes.\footnote{See Appendix
  \ref{app:custom} for the word lists.}

Another important observation is that \textsf{MAC} calculations do not
distinguish whether a given attribute is associated with a given
protected word, simply averaging across all such groups. Let us use the
case of religion-related stereotypes to illustrate. The full lists from
{[}13{]} can be found in Appendix \ref{appendix:manzini_word_lists}. In
the original paper, words from all three religions were compared against
all of the stereotypes. No distinction between cases in which the
stereotype is associated with a given religion, as opposed to the
situation in which it is associated with another one, is made. For
example, the protected word \emph{jew} is supposed to be stereotypically
connected with the attribute \emph{greedy}, while from the protected
word \emph{quran} the attribute \emph{greedy} comes from a different
stereotype, and yet the distances between these pairs contribute equally
to the final \textsf{MAC} score. This is problematic, as not all of the
stereotypical words have to be considered harmful for all 
religions. To avoid the masking effect, one should pay attention to how
protected words and attributes are paired with stereotypes.

\begin{figure}[H]

\begin{center}\includegraphics[width=0.7\linewidth]{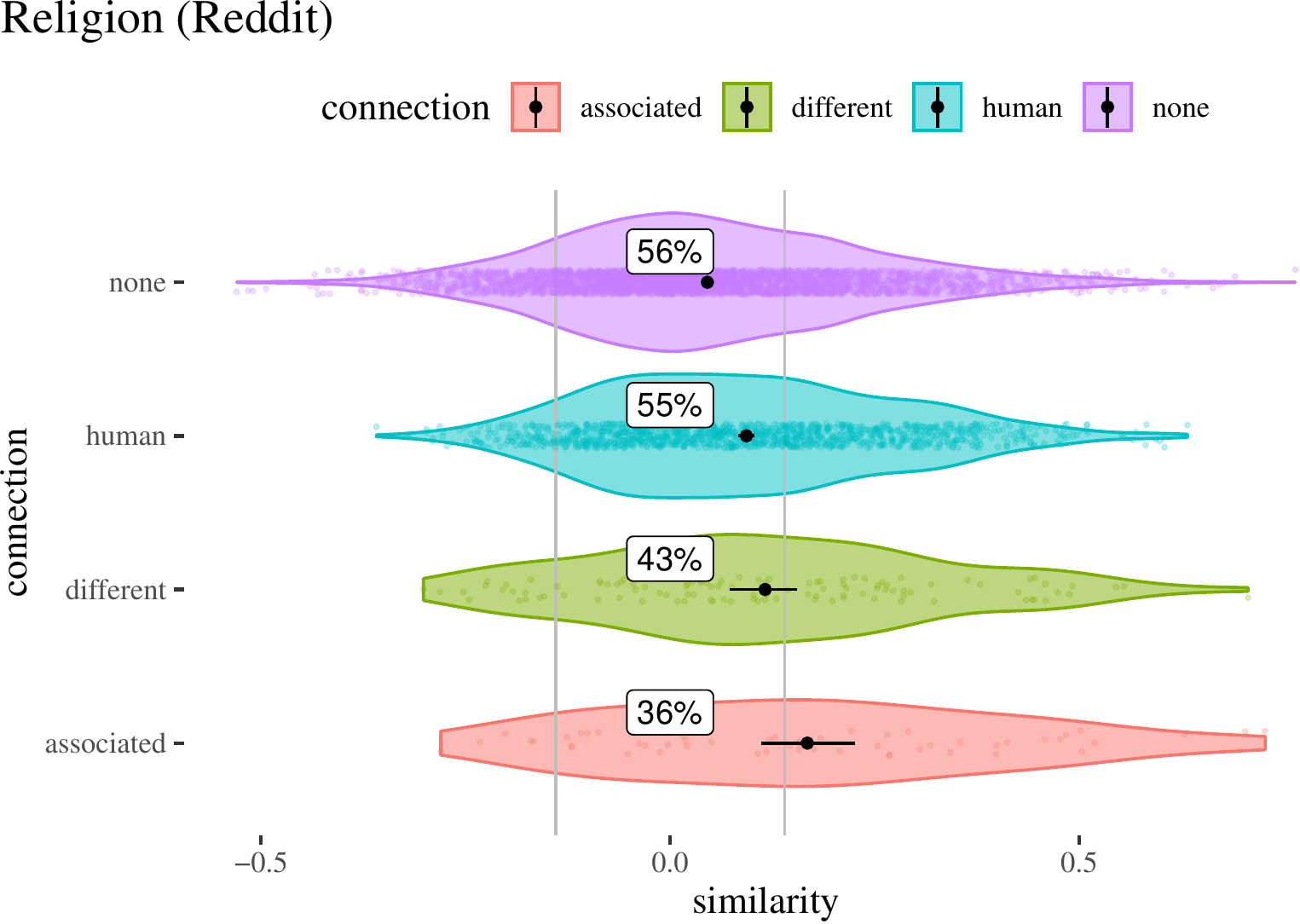} \end{center}

\caption{Empirical distributions of cosine similarities  for the Religion word list  used in  the original paper. }

\label{fig:empirical0}
\end{figure}

\begin{figure}[H]

\begin{center}\includegraphics[width=0.7\linewidth]{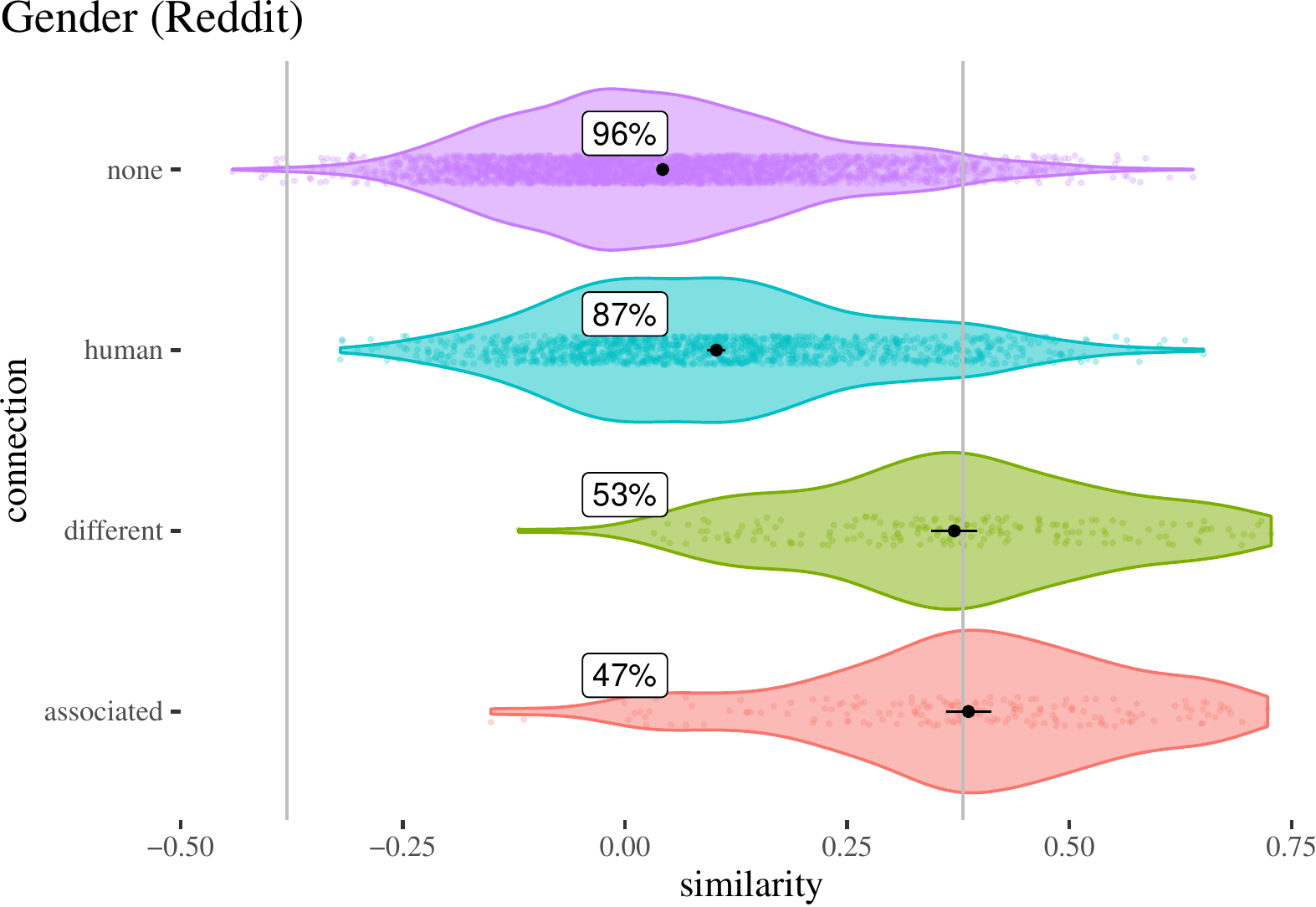} \end{center}

\caption{Empirical distributions of cosine similarities for the Gender word list used in  the original paper.  }

\label{fig:empirical2}
\end{figure}

\begin{figure}[H]

\begin{center}\includegraphics[width=0.7\linewidth]{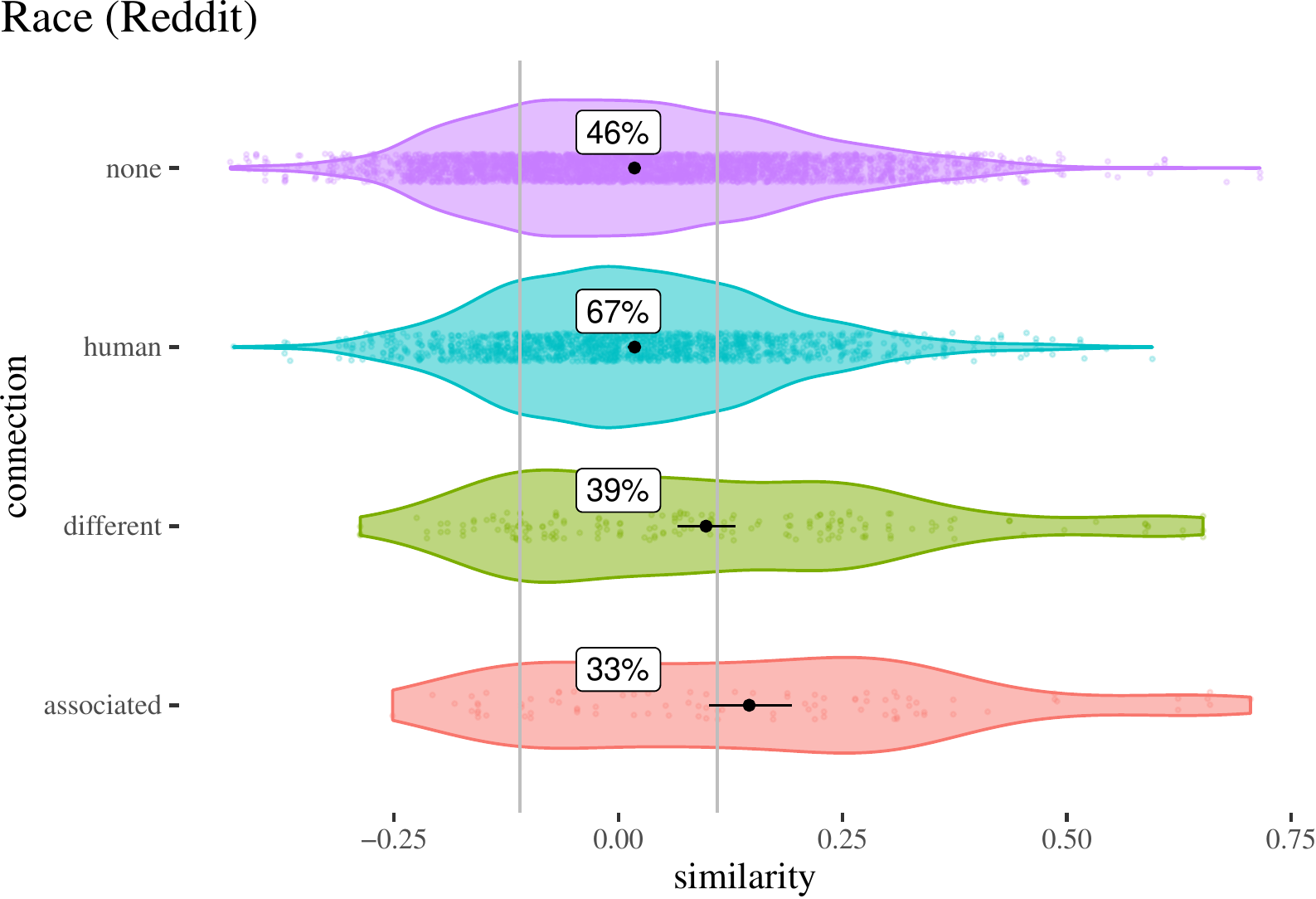} \end{center}

\caption{Empirical distributions of cosine similarities  for the Race word list used in  the original paper.  }

\label{fig:empirical3}
\end{figure}

In Figures (\ref{fig:empirical0}-\ref{fig:empirical3}) we look at the
empirical distributions, while paying attention to such divisions. The
horizontal lines represent the values of \(1 - \mathsf{MAC}\) (that is,
we now talk in terms of cosine similarity rather than cosine distance)
that the authors considered indicative of bias for stereotypes
corresponding to given word lists. For instance, in religion,
\textsf{MAC} was .859, which was considered a sign of bias, so we plot
\(0\pm (1-.859)\approx .14\) lines around similarity = 0 (that is,
distance = 1). Notice that most distributions are quite wide, and the
proportions of even neutral or human neutral words with similarities
higher than the value of \(1 - \mathsf{MAC}\) deserving debiasing
according to the authors are quite high.

Another issue to consider is the selection of attributes for bias
measurement. The word lists used in the literature are often fairly
small (5-50). The papers in the field do employ statistical tests to
measure the uncertainty involved and do make claims of statistical
significance. Yet, we will later on argue that these method are not
proper for the goal at hand. By using Bayesian methods we will show that
a more appropriate use of statistical methods leads to estimates of
uncertainty which suggest that larger word lists would be advisable.

To avoid the problem brought up in this subsection, we employ control
groups and in line with Bayesian methodology, use posterior
distributions and highest posterior density intervals instead of chasing
single-point metrics based on pre-averaged data. Before we do so, we
first explain why pre-averaging and chasing single-number metrics is a
sub-optimal strategy.

\hypertarget{problems-with-pre-averaging}{%
\subsection{Problems with
pre-averaging}\label{problems-with-pre-averaging}}

\label{subsec:problems}

\begin{figure}[H]

\begin{center}\includegraphics[width=0.9\linewidth]{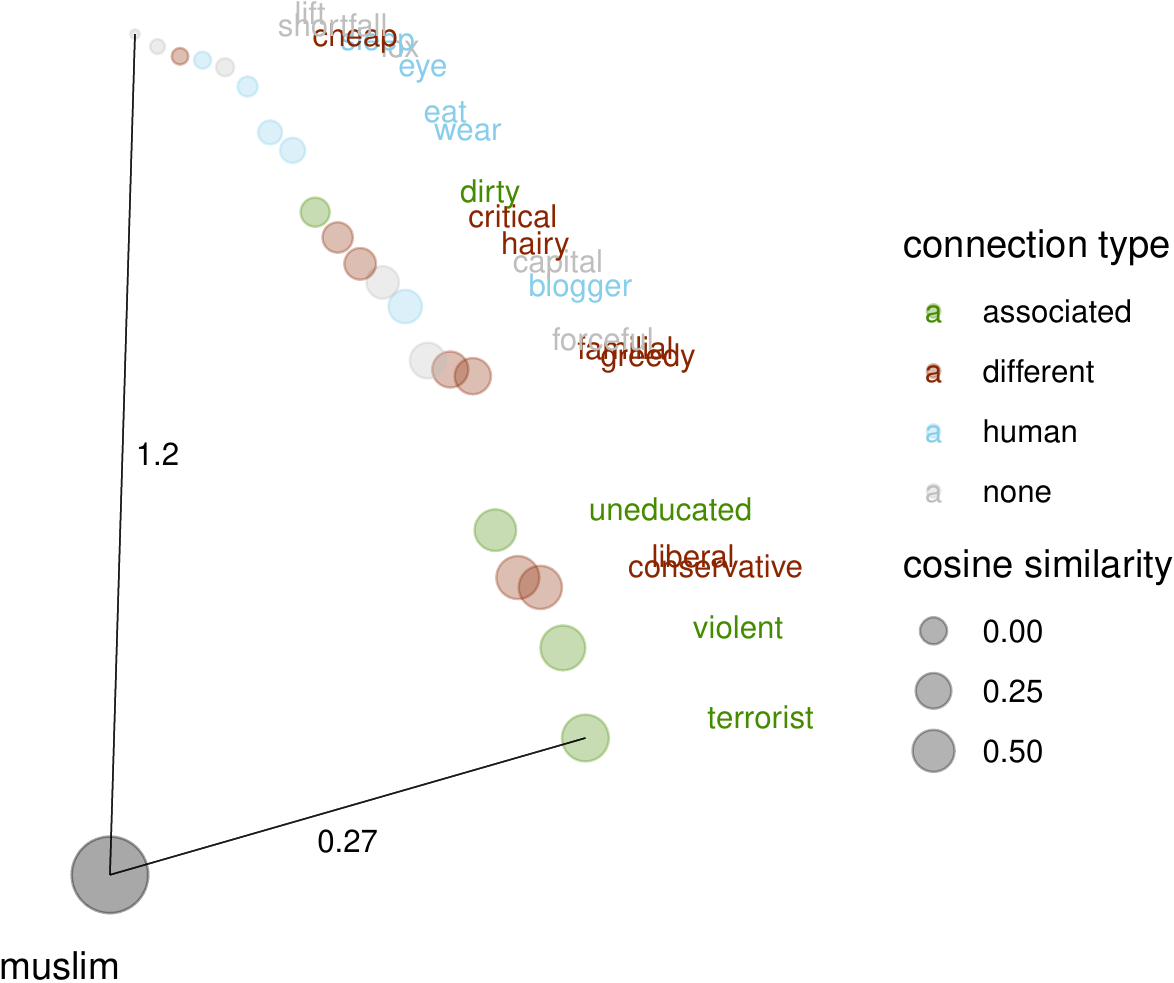} \end{center}
\caption{Actual distances for the protected word \textsf{muslim}. }
\label{fig:muslim}
\end{figure}

\begin{figure}[H]

\begin{center}\includegraphics[width=0.9\linewidth]{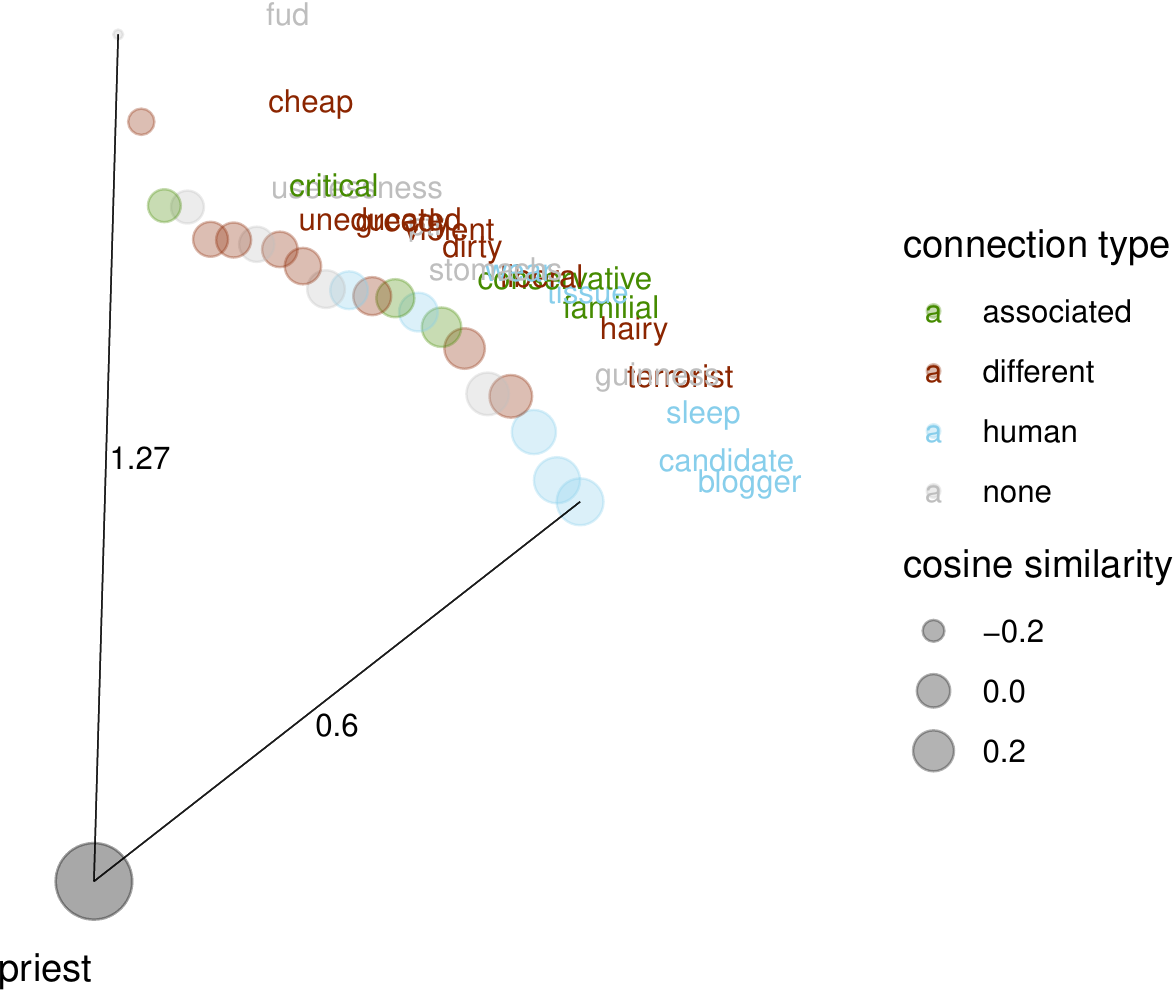} \end{center}

\caption{Actual distances for the protected word \textsf{priest}.}
\label{fig:priest}
\end{figure}

The approaches we have been describing use means of mean average cosine
similarities to measure similarity between protected words and
attributes coming from harmful stereotypes. But once we take a look at
the individual values, it turns out that the raw data variance is rather
high, and there are quite a few outliers and surprisingly dissimilar
words. This problem becomes transparent when we examine the
visualizations of the individual cosine distances, following the idea
that one of the first steps in understanding data is to look at it.
Let's start with inspecting two examples of such visualizations in
Figures \ref{fig:muslim} and \ref{fig:priest} (we also include neutral
and human predicates to make our point more transparent). Again, we
emphasize that \textbf{we do not condone the associations which we are
about to illustrate.}

As transparent in Figures \ref{fig:muslim} and \ref{fig:priest}, for the
protected word \textsf{muslim}, the most similar attributes tend to be
the ones associated with it stereotypically, but then words associated
with other stereotypes come closer than neutral or human predicates. For
the protected word \textsf{priest}, the situation is even less as
expected: the nearest attributes are human attributes, and all there
seems to be no clear pattern to the distances to other attributes.

The general phenomenon that makes us skeptical about running statistical
tests on pre-averaged data is that raw datasets of different variance
can result in the same pre-averaged data and consequently the same
single-number metric. In other words, a method that proceeds this way is
not very sensitive to the real sample variance.

Let us illustrate how this problem arises in the context of
\textsf{WEAT}. Once a particular \(s(X,Y,A,B)\) is calculated, the
question arises as to whether a value that high could have arisen by
chance. To address the question, each \(s(X,Y,A,B)\) is used in the
original paper to generate a \(p\)-value by bootstrapping. The
\(p\)-value is the frequency of how often it is the case that
\(s(X_i,Y_i,A,B)>s(X,Y,A,B)\) for sampled equally sized partitions
\(X_i, Y_i\) of \(X\cup Y\). The WEAT score is then computed by
standardizing the difference in means of means by dividing by the
standard deviation of means, see equation \eqref{eq:weat}.

\normalsize

\noindent The \textsf{WEAT} scores reported by {[}9{]} for lists of
words for which the embeddings are supposedly biased range from 2.06 to
1.81, and the reported \(p\)-values are in the range of
\(10^{-7}-10^{-2}\) with one exception for \emph{Math vs Arts}, where it
is \(.018\).

The question is, are those results meaningful? One way to answer this
question is to think in terms of null generative models. If the words
actually are samples from two populations with equal means, how often
would we see \textsf{WEAT} scores in this range? How often would we
reach the \(p\)-values that the authors reported?

Imagine there are two groups of protected words, each of size 8, and two
groups of stereotypical attributes, of the same size.\footnote{16 is the
  sample size used in the WEAT7 word list, which is not much different
  from the other sample sizes in word lists used by {[}9{]}).} Each such
a collection of samples, as far as our question is involved, is
equivalent to a sample of \(16^2\) cosine distances. Further, imagine
there really is no difference between these groups of words and the model
is in fact null. That is, we draw the cosine distances from the
\(\mathsf{Normal}(0,.08)\) distribution.\footnote{\(.08\) is
  approximately the empirical standard deviation observed in fairly
  large samples of neutral words.}

\begin{figure}[H]

\begin{center}\includegraphics[width=0.8\linewidth]{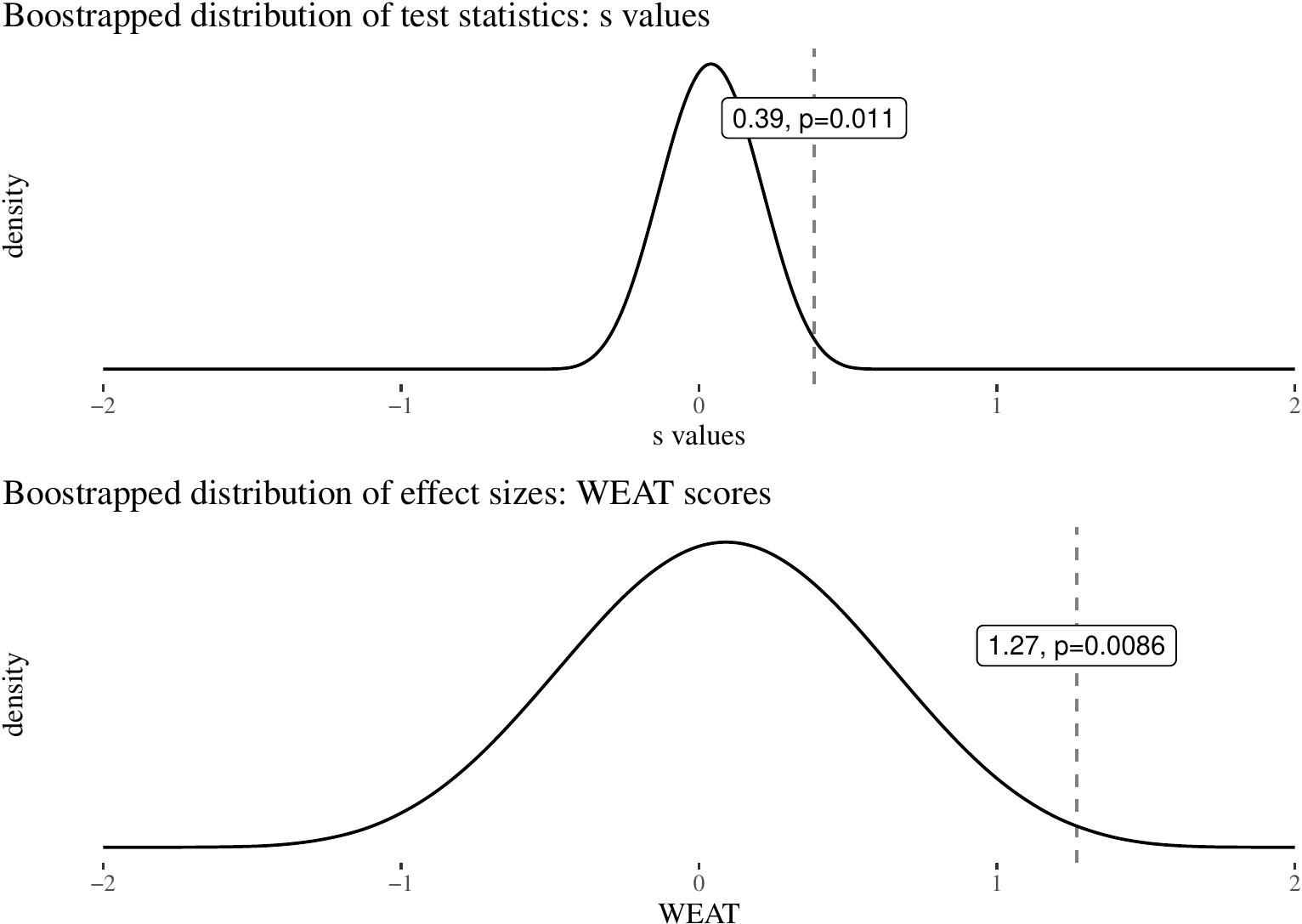} \end{center}

\caption{Bootstrapped distributions of test statistics and effect sizes in a random sample given the null hypothesis. We used a sample from the null  model with N(0,.08) and 16 protected words, and then bootstrapped from it, following the original methodology. One particular bootstrapped sample is highlighted, and discussed further in the text. }
\label{fig:caliskanCalc}
\end{figure}

In Figure \ref{fig:caliskanCalc} we illustrate one iteration of the
procedure. We draw one such sample of size \(16^2\). Then we actually
list all possible ways to split the 16 words in two equal sets (each
such a split is one bootstrapped sample) and for each of them we
calculate the \textsf{s} values and \textsf{WEAT}. What are the
resulting distributions of \(\textsf{s}\) scores and what \(p\)-values
do they lead to? What are the resulting effect sizes for each
bootstrapped sample, and how often can we get an effect size as large as
the ones reported in the original paper?

In the bootstrapped samples we would rather expect low \textsf{s} values
and low \textsf{WEAT}: after all, these are just random permutations of
random distances all of which are drawn from the same null distribution.
Let's take a look at one such a bootstrapped sample.

On purpose, we picked a rather unusual one: the observed test statistic
is 0.39 and 1.27. The bootstrapped distributions of the test statistics
and effect sizes are illustrated in Figure \ref{fig:caliskanCalc},
together with this particular example. Quite notably both (two-sided)
\(p\) values for our example are rather low (Figure
\ref{fig:caliskanCalc}). These facts might suggest that we ended up
with a situation where ``bias'' is present (albeit, due to random
noise). The reason why we picked it is that it is an example of a word
list that ends up with relatively low \(p\)-value and a relatively
unusual effect size, but nevertheless, its closer inspection shows that
even for a word list with such properties there is no clear reason to
think that the bias is present.

\begin{figure}[H]

\begin{center}\includegraphics[width=0.8\linewidth]{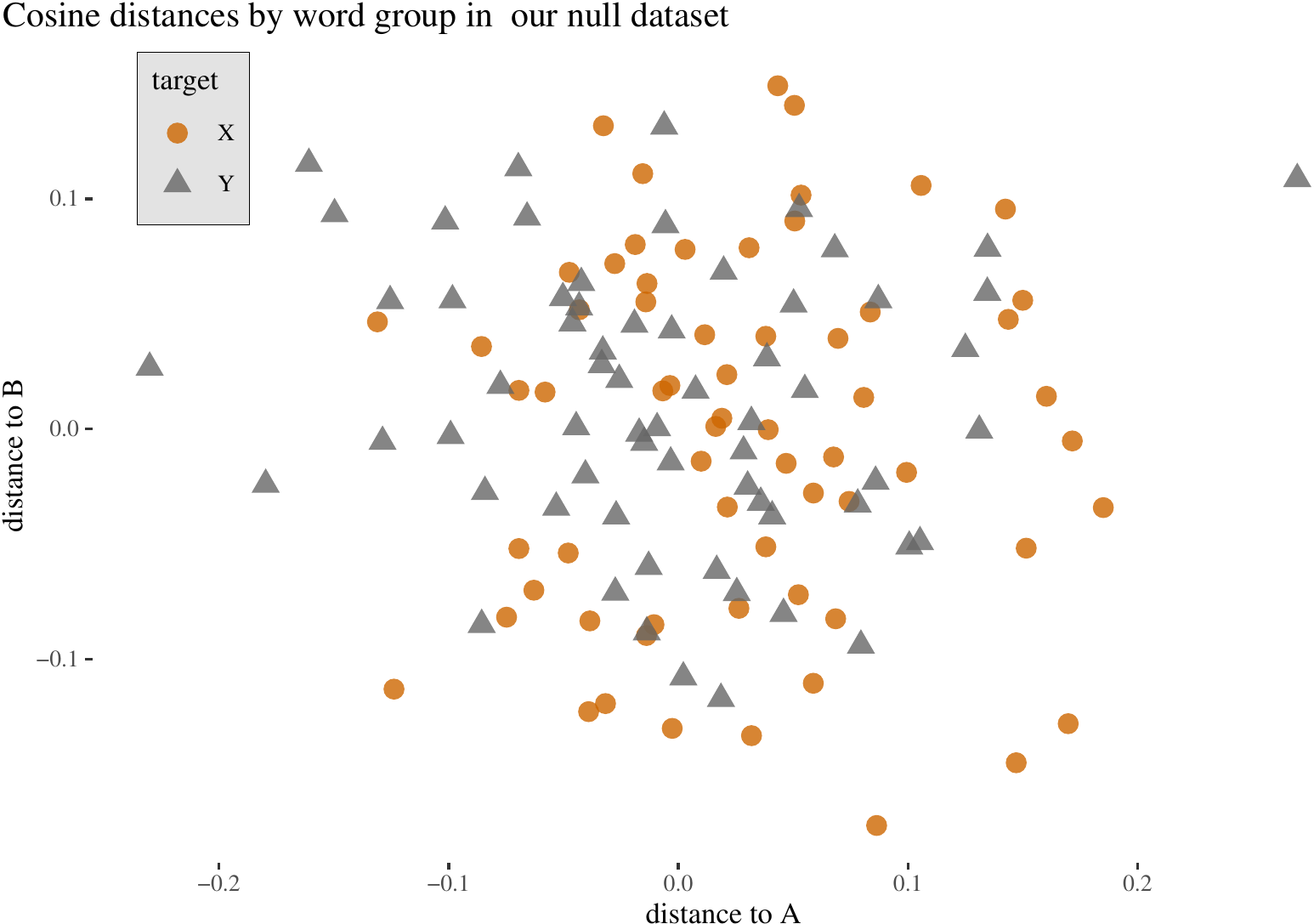} \end{center}

\caption{Cosine distances to two attribute sets by protected word groups. Observe nothing unusual except for a few outliers.}
\label{fig:caliskanDistances}
\end{figure}

At this point, we might think that we just stumbled into a bootstrapped
sample that randomly happened to display strong bias. We decide to
double-check this by visual inspection expecting exactly this: a strong,
clearly visible bias (Figure \ref{fig:caliskanDistances}).

\noindent In fact, while there might be some outliers here and there,
saying that a clear bias on which one group is systematically closer to
\(A\)s than another is definitely a stretch. What happened?

In the calculations of \textsf{WEAT} means are taken twice. The
\textsf{s}-values themselves are means and then means of
\textsf{s}-values are compared between groups. Statistical troubles
start when we run statistical tests on sets of means, for at least two
reasons.

\begin{enumerate}

\item By pre-averaging data we throw away information about sample sizes. For the former point, think about proportions: 10 out of 20 and 2 out of 4 give the same mean, but you would obtain more information by making the former observation rather than by making the latter.  And especially in this context, in which the word lists are not huge, sample sizes should matter.

\item When we pre-average, we disregard variation, and therefore pre-averaging  tends to manufacture false confidence. Group means display less variation than the raw data points and the standard deviation of a set of means of sets of means is bound to be lower than the original standard deviation in the row data. Now, if you calculate your effect size by dividing by the pre-averaged standard deviation, you are quite likely to get something that looks like a strong effect size, but the results of your calculations might not track anything interesting.
\end{enumerate}

Let us think again about the question that we are ultimately interested
in. Are the \(X\) terms systematically closer to (further from) the
\(A\) attributes (\(B\) attributes) than the \(Y\) words? But now let's
use the raw data points to try to answer these questions.

To start with, let us run two quick \(t\)-tests to gauge what the raw
data illustrated in Figure \ref{fig:caliskanDistances} tell us. First,
distances to \(A\) attributes for \(X\) words and \(Y\) words. Well, the
result is---strictly speaking---statistically significant. The
\(p\)-value is \(0.02\) (more than ten times higher than the
\(p\)-valued obtained by the bootstrapping procedure. So the sample is
in some sense unusual. But the 95\% confidence interval for the
difference in means is \([.0052, .061]\), clearly nothing that a reader
would expect given that the calculated effect size seemed quite large.
How about the distances to the \(B\) attributes? Here the \(p\)-value is
\(.22\) and the 95\% confidence interval is \([-0.03, .009]\), even less
of a reason to think a bias is present.

The difficulties are exacerbated by the fact that statistical tests are
based on bootstrapping from a relatively small data sets, which is quite
likely to underestimate the population variance. To make our point
clear, let us avoid bootstrapping and work with the null generative
model with \(\mathsf{Norm}(0,.08)\) for both word groups. We keep the
sizes the same: we have eight protected words in each group, sixteen in
total, and for each we randomly draw 8 distances from hypothetical \(A\)
attributes, and \(8\) distances from hypothetical \(B\) attributes.
Calculate the test statistic and effect size the way {[}9{]} did. Do
this 10000 times, each time calculating \textsf{WEAT} and \textsf{s}
values, and look at what the distributions of these values are on the
assumption of the null model with realistic empirically motivated raw
data point standard deviation.

\begin{figure}[H]

\begin{center}\includegraphics[width=0.8\linewidth]{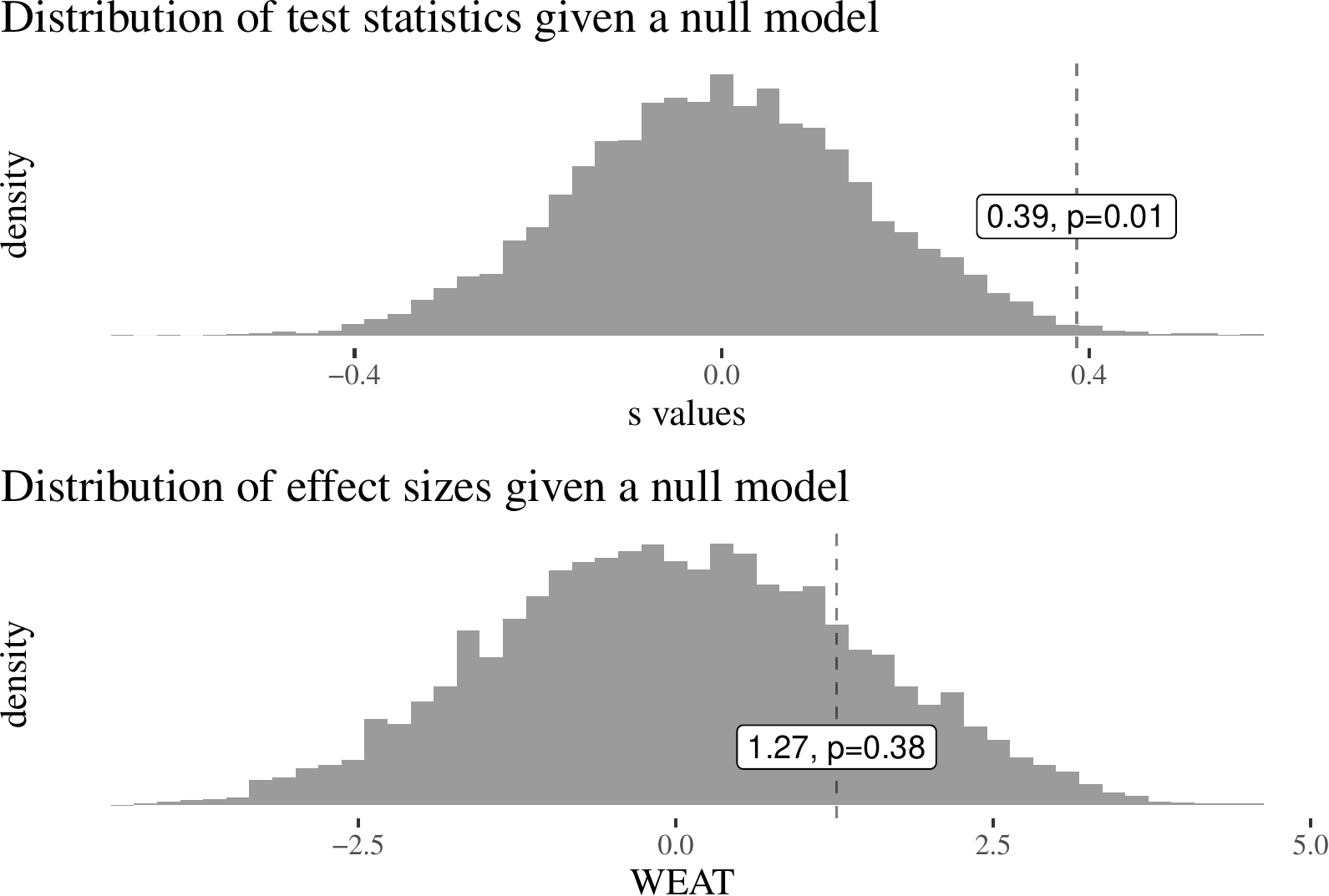} \end{center}

\caption{Distributions of test statistics and effect sizes based on 10k simulations on the assumption of a null model in which all distances come from normal distribution with $\mu =0, \sigma = .08, n=10k$. We also mark the sample we have been using as an example.}
\label{fig:ourDistances}
\end{figure}

The first observation is that the supposedly large effect size we
obtained is not that unusual even assuming a null model. Around 38\% of
samples result in \textsf{WEAT} score at least as extreme. This
illustrates the point that it does not constitute a strong evidence of
bias. Second, the distribution of \(\textsf{s}\) values is much more
narrow, which means that if we use it to calculate \(p\)-values, it is
not too difficult to obtain a supposedly significant test statistic
which nevertheless does not correspond to anything interesting happening
in the data set.

We have seen that seemingly high effect sizes might arise even if the
underlying processes actually have the same mean. The uncertainty
resulting from including the raw data point variance in considerations
is more extensive than the one suggested by the low \(p\)-values
obtained from taking means or means of means as data points. In the
section we discussed the performance of the \textsf{WEAT} measure, but
since the {[}13{]} one is a generalization thereof, including the method
of running statistical tests on pre-averaged data, our remarks,
\emph{mutatis nutandis}, apply.

What is the alternative? As we already emphasized: focusing on what the
real underlying question is and trying to answer it using a statistical
analysis of the raw data using meaningful control groups, to ensure
interpretability. Moreover, since the data sets are not too large and
since multiple evaluations are to be made, we will pursue this method
from the Bayesian perspective. Now we have to describe it.

\hypertarget{a-bayesian-approach-to-cosine-based-bias}{%
\section{A Bayesian approach to cosine-based
bias}\label{a-bayesian-approach-to-cosine-based-bias}}

\label{sec:bayesian}

\hypertarget{model-construction}{%
\subsection{Model construction}\label{model-construction}}

\label{subsec:model}

Bayesian data analysis takes prior probability distributions, a
mathematical model structure and the data, and returns the posterior
probability distributions over the parameters of interest, thus
capturing our uncertainty about their actual values. One important
difference between such a result and the result of classical
statistical analysis is that classical confidence intervals (CIs) have a
rather complicated and somewhat confusing interpretation, which has
little to do with the posterior probability distribution.\footnote{Here
  are a few usual problems. CIs are often mistakenly interpreted as
  providing the probability that a resulting confidence interval
  contains the true value of a parameter. CIs bring confusion also with
  regard to precision, it is a common mistake to interpret narrow
  intervals as the ones corresponding to more precise knowledge.
  Another fallacy is to associate CIs with likelihood and to state that
  values within a given interval are more probable than the ones outside
  it. The theory of confidence intervals does not support the above
  interpretations. CIs should be plainly interpreted as a result of
  a certain procedure (there are many ways to obtain CIs from a given set
  of data) that will in the long run contain the true value if the
  procedure is performed a fixed amount of times. For a nice survey and
  explanation of these misinterpretations, see {[}17{]}. For a
  psychological study of the occurrence of such misinterpretations, see
  {[}8{]}. In this study, 120 researchers and 442 students were asked to
  assess the truth value of six false statements involving different
  interpretations of a CI. Both researchers and students endorsed, on
  average, more than three of these statements.}

In fact, Bayesian highest posterior density intervals (HPDIs, the
narrowest intervals containing a certain ratio of the area under the
curve) and CIs end up being numerically the same only if the prior
probabilities are uniform. This illustrates that (1) classical analysis
is unable to incorporate non-trivial priors, and (2) is therefore more
susceptible to over-fitting, unless regularization (equivalent to a more
straightforward Bayesian approach) is used. In contrast with CIs, the
posterior distributions are easily interpretable and have direct
relevance to the question at hand. Moreover, Bayesian data analysis is
better at handling hierarchical models and small datasets, which is
exactly what we will we dealing with.

In standard Bayesian analysis, the first step is to understand the
data, think hard about the underlying process, and select potential
predictors and the outcome variable. The next step is to formulate a
mathematical description of the generative model of the relationships
between the predictors and the outcome variable. Prior distributions
must then be chosen for the parameters used in the model. Next, Bayesian
inference must be applied to find posterior distributions over the
possible parameter values. Finally, we need to check how well the
posterior predictions reflect the data with a posterior predictive
check.

In our analysis, the outcome variable is the cosine distances between the
protected words and attribute words. The predictor is a factor
determining whether a given attribute word is a neutral word, a human
predicate is stereotypically associated with the protected word, or
comes from a different stereotype connected with another protected word.
The idea is really straightforward: if bias is present in the embedding,
distances to associated attribute words should be systematically lower
than to other attribute words.

\begin{figure}[H]

\begin{center}\includegraphics[width=0.8\linewidth]{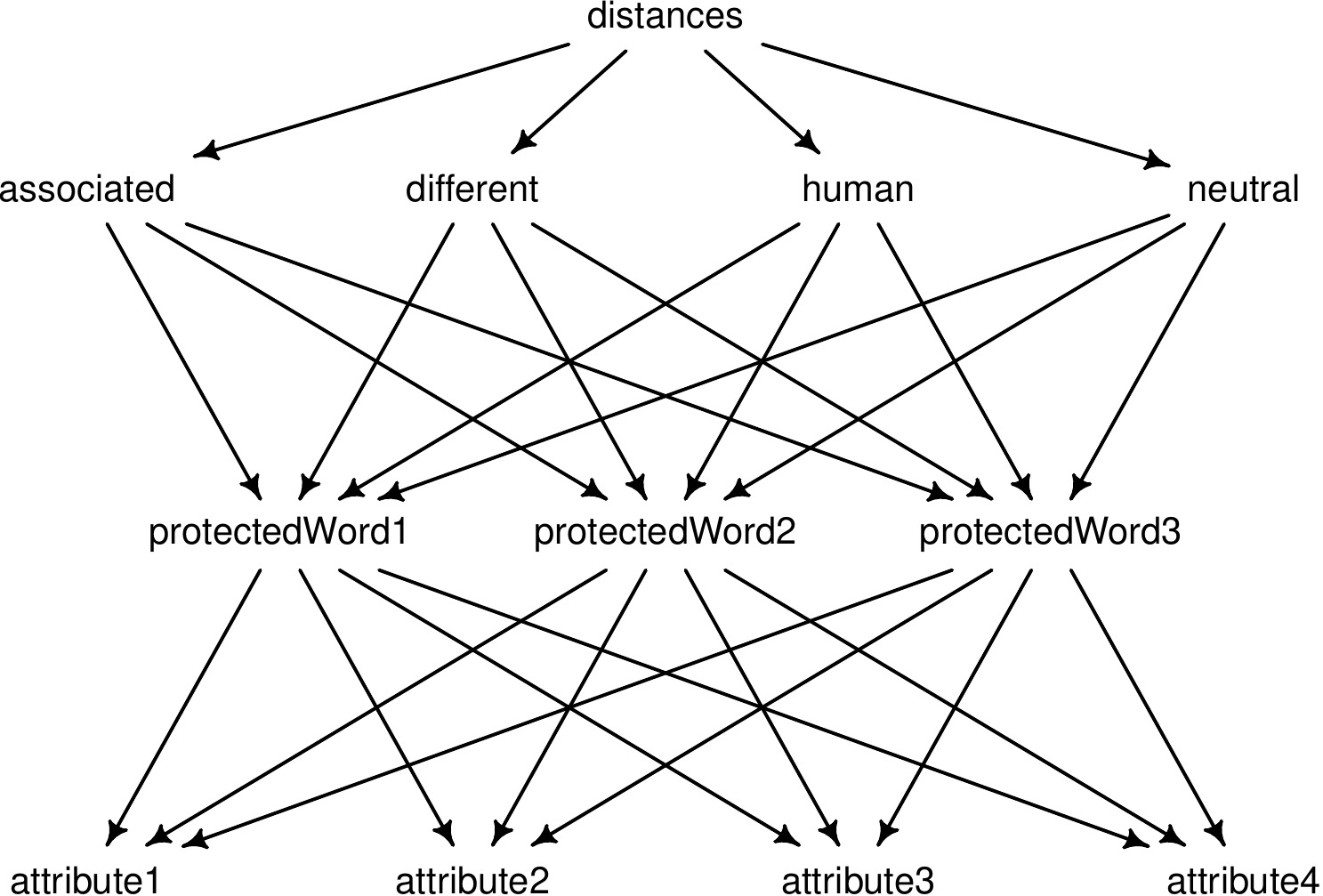} \end{center}

\caption{At a general level, we will be estimating the coefficients for distances as grouped by whether they are between protected words and attributes coming from their respective associated/different/human/neutral attribute groups. At a more fine-grained level, for each protected word we will be estimating the proximity of that word to attributes that are associated with its respective stereotype, come from a different stereotype, or come from the human/neutral attribute lists.}
\label{fig:visDag}
\end{figure}

Furthermore, conceptually there are two levels of analysis in our
approach (see Figure \ref{fig:visDag}). On the one hand, we are
interested in the general question of whether related attributes are
systematically closer across the dataset. On the other hand, we are
interested in a more fine-grained picture of the role of the predictor
for particular protected words. Learning in hierarchical Bayesian models
involves using Bayesian inference to update the parameters of the model.
This update is based on the observed data, and estimates are made at
different levels of the data hierarchy. We use hierarchical Bayesian
models in which we simultaneously estimate parameters at the protected
word level and at the global level, assuming that all lower-level
parameters are drawn from global distributions. Such models can be
thought of as incorporating adaptive regularization, which avoids
overfitting and leads to improved estimates for unbalanced datasets (and
the datasets we need to use are unbalanced).

To be more specific, the underlying mathematical model is as follows.
First, we assume that distances are normally distributed:
\begin{align*} \mathsf{distance_i} & \sim \mathsf{dnorm}(\mu_i,\sigma_i)
\end{align*} \noindent Second, for each particular protected word
\(\mathsf{pw}\) there are four parameters to be estimated. Its mean
distance to associated stereotypes \(a[pw]\), its mean distance to
attributes coming from different stereotypes, \(d[pw]\), its mean
distance to human attributes, \(h[pw]\), and its mean distance to
neutral attributes, \(n[pw]\): \begin{align*}
\mu_i & = d_{\mathsf{pw[i]}} \times \mathsf{different}_i  + a_{\mathsf{pw[i]}} \times \mathsf{associated}_i  + h_{\mathsf{pw[i]}} \times \mathsf{human}_i  + n_{\mathsf{pw[i]}}\times \mathsf{neutral}_i
\end{align*} \noindent where
\(\mathsf{different}, \mathsf{associated},\mathsf{human}\) and
\(\mathsf{neutral}\) are binary variables. This completes our
description of the simple underlying process that we would like to
investigate.

Now the priors and the hierarchy. We assume all the \(a\) parameters
come from one distribution, that is normal around a higher-level
parameter \(\bar{a}\) and so on for the other three groups of
parameters. That is, \(a_{\mathsf{pw[i]}}\) is the average distance of a
given particular protected word to attributes stereotypically associated
with it, while \(\bar{a}\) is the overall average distance of protected
words to attributes associated with them.\footnote{For a thorough
  introduction to the concepts we're using, see {[}11,15{]}.}
\begin{align*}
d_{\mathsf{pw[i]}} \sim \mathsf{Norm}(\bar{d}, \overline{\sigma_d}) &  & 
a_{\mathsf{pw[i]}} \sim \mathsf{Norm}(\bar{a}, \overline{\sigma_a}) \\
h_{\mathsf{pw[i]}} \sim \mathsf{Norm}(\bar{h}, \overline{\sigma_h}) & & 
n_{\mathsf{pw[i]}} \sim \mathsf{Norm}(\bar{n}, \overline{\sigma_n}) 
\end{align*}

According to our priors, the group means \(\bar{a}\), \(\bar{d}\),
\(\bar{h}\) and \(\bar{n}\) all come from one normal distribution with
mean equal to \(1\) and standard deviation equal to \(.3\). The standard
deviations \(\bar{\sigma_a}\), \(\bar{\sigma_d}\), \(\bar{\sigma_h}\)
and \(\bar{\sigma_n}\) to be estimated, according to our prior, come
from one distribution, exponential with rate parameter equal to \(2\).
Our priors are slightly skeptical. They do reflect our knowledge and
intuition on the probable distribution of the cosine distances in the
data. We know that the cosine distances lie in the range \(0-2\), and we
expect two randomly chosen vectors from the embedding to have rather
small similarity, so we expect the distances to be centered around
\(1\). However, we use a rather wide standard deviation (\(.3\)) to
easily account for cases where there is actually much higher similarity
between two vectors (especially in cases where the embedding is supposed
to be biased). Our priors for the standard deviations are also fairly
weak. \begin{align*}
\bar{d}, \bar{a}, \bar{h}, \bar{n} &\sim \mathsf{Norm}(1, .3)\\ 
\overline{\sigma_d}, \overline{\sigma_a},  \overline{\sigma_h},  \overline{\sigma_n}  &\sim \mathsf{Exp}(2)
\end{align*}

\hypertarget{posterior-predictive-check}{%
\subsection{Posterior predictive
check}\label{posterior-predictive-check}}

\label{subsec:posterior}

\begin{figure}[H]

\begin{center}\includegraphics[width=0.8\linewidth]{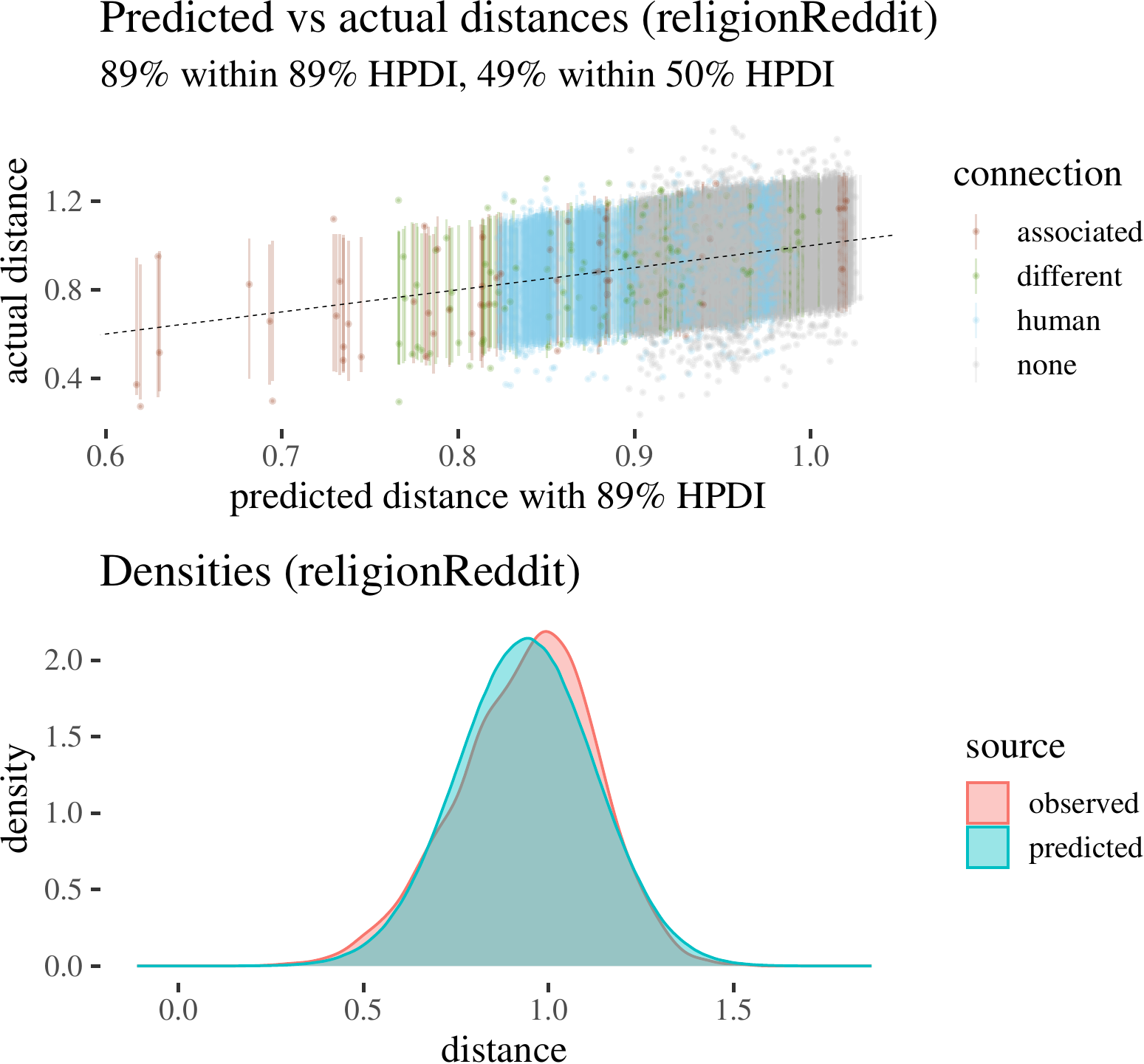} \end{center}
\caption{Example of a posterior predictive check. (Top) Actual cosine distances are plotted against mean predictions with 89\% highest posterior density intervals. Notice that 90\% of actual values fall within the 89\% HPDI and 55\% of actual values fall into 50\% HPDI, which indicates appropriate performance of the model. The left-right alignment of different colors corresponds to the fact that cosine differences between elements of different categories differ, to some extent systematically (this will be studied in the results section). (Bottom) Densities of predicted and observed distances.}
\label{fig:posteriorCheck1}
\end{figure}

A posterior predictive check is a technique used to evaluate the fit of
a Bayesian model by comparing its predictions with observed data. The
underlying principle is to generate simulated data from the posterior
distribution of the model parameters and compare them with the observed
data. If the model is a good fit to the data, the simulated data should
resemble the observed data. In Figure \ref{fig:posteriorCheck1} we
illustrate a posterior predictive check for one corpus (Reddit) and one
word list. The remaining posterior predictive checks are in section
\ref{appendix:posterior}.

\normalsize

\hypertarget{results-and-discussion}{%
\section{Results and discussion}\label{results-and-discussion}}

\label{sec:results}

\vspace{1mm}

\normalsize

\hypertarget{observations}{%
\subsection{Observations}\label{observations}}

\label{subsec:observations}

In brief, despite one-number metrics suggesting otherwise, our Bayesian
analysis reveals that insofar as the short word lists usually used in
related research projects are involved, there usually are no strong
reasons to claim the presence of systematic bias. Moreover, comparison
between the groups (including control word lists) leads to the
conclusion that the effect sizes (that is, the absolute differences
between cosine distances between groups) tend to be rather small, with
few exceptions. Moreover, the choice of protected words is crucial ---
as there is a lot of variance when it comes to the protected word-level
analysis.

In a bit more detail, the visualizations in Appendix
\ref{appendix:visualizations} show that the situation is more
complicated than merely looking at one-number summaries might suggest.
Note that the axes are sometimes in different scales to increase
visibility.

To start with, let us look at the association-type level coefficients
(illustrated in the top parts of the plots). Depending on the corpus
used and word class, there is a large variety as to posterior
densities. Quite aware of this being a crude approximation, let's
compare the HPDIs and whether they overlap for different attribute
groups.

\begin{itemize}
\item
  In Weat 7 (Reddit) there is no reason to think there are systematic
  differences between cosine distances (recall that words from Weat 7
  were mostly not available in other embeddings).
\item
  In Weat 1 (Google, Glove and Reddit) associated words are somewhat
  closer, but the cosine distance differences from neutral words are
  very low, and surprisingly it is human attributes, not neutral
  predicates that are systematically the furthest.
\item
  In Religion (Google, Glove, Reddit) and Race (Google, Glove), the
  associated attributes are not systematically closer than attributes
  belonging to different stereotypes, and the difference between neutral
  and human predicates is rather low, if noticeable. The situation is
  interestingly different in Race (Reddit) where both human and neutral
  predicates are systematically further than associated and different
  attributes---but even then, there is no clear difference between
  associated and different attributes.
\item
  For Gender (Google, Glove), despite the superficially binary nature,
  associated and opposite attributes tend to be more or less in the same
  distances, much closer than neutral words (but not closer than human
  predicates in Glove). Reddit is an extreme example: both associated
  and opposite attributes are much closer than neutral and human (around
  .6 vs.~.9), but even then, there seems to be no reason to think that
  cosine distances to associated predicates are much different from
  distances to opposite predicates.
\end{itemize}

Moreover, when we look at particular protected words, the situation is
even less straightforward. We will just go over a few notable examples,
leaving the visual inspection of particular results for other protected
words to the reader. One general phenomenon is that---as we already
pointed out---the word lists are quite short, which contributes to large
uncertainty involved in some cases.

\begin{itemize}
\item
  For some protected words the different attributes are somewhat closer
  than the associated attributes.
\item
  For some protected words, associated and different attributes are
  closer than neutral attributes, but so are human attributes.
\item
  In some cases, associated attributes are closer, but so are neutral
  and human predicates, which illustrates that just looking at average
  cosine similarity as compared to the theoretically expected value of
  1, instead of running a comparison to neutral and human attributes is
  misleading.
\item
  The only group of protected words where differences are noticeable at
  the protected word level is Gender-related words-- as in Gender
  (Google) and in Gender (Reddit) --- note however that in the latter,
  for some words, the opposite attributes seem to be a little bit closer
  than the associated ones.
\end{itemize}

\hypertarget{rethinking-debiasing}{%
\subsection{Rethinking debiasing}\label{rethinking-debiasing}}

\label{subsec:rethinking}

\begin{figure}[H]

\begin{center}\includegraphics{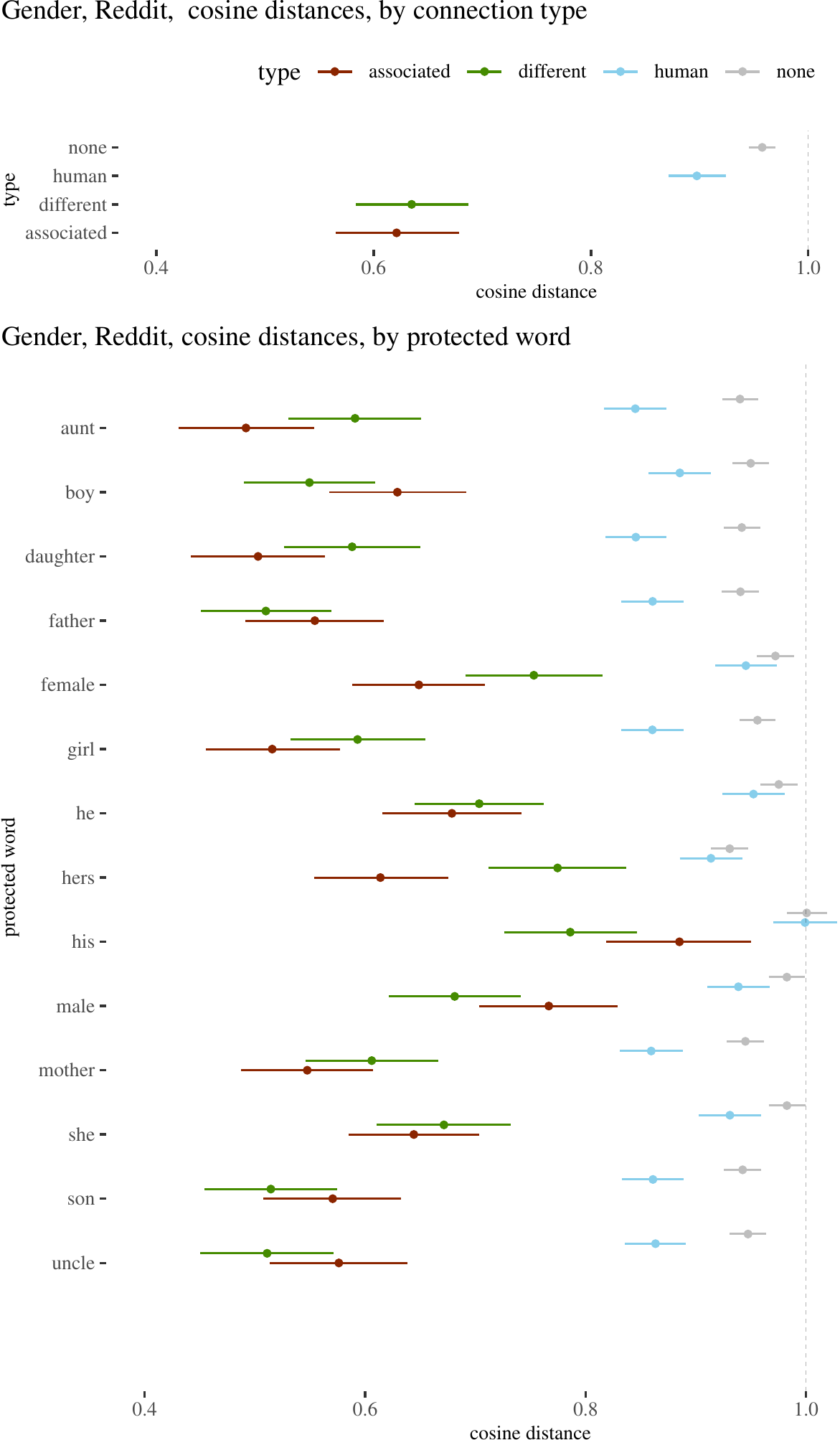} \end{center}
\caption{Mean cosine distances with 89\% highest posterior density intervals for the  gender dataset before debiasing.}
\label{fig:empiricalPriorToDebiasing}
\end{figure}

\begin{figure}[H]

\begin{center}\includegraphics{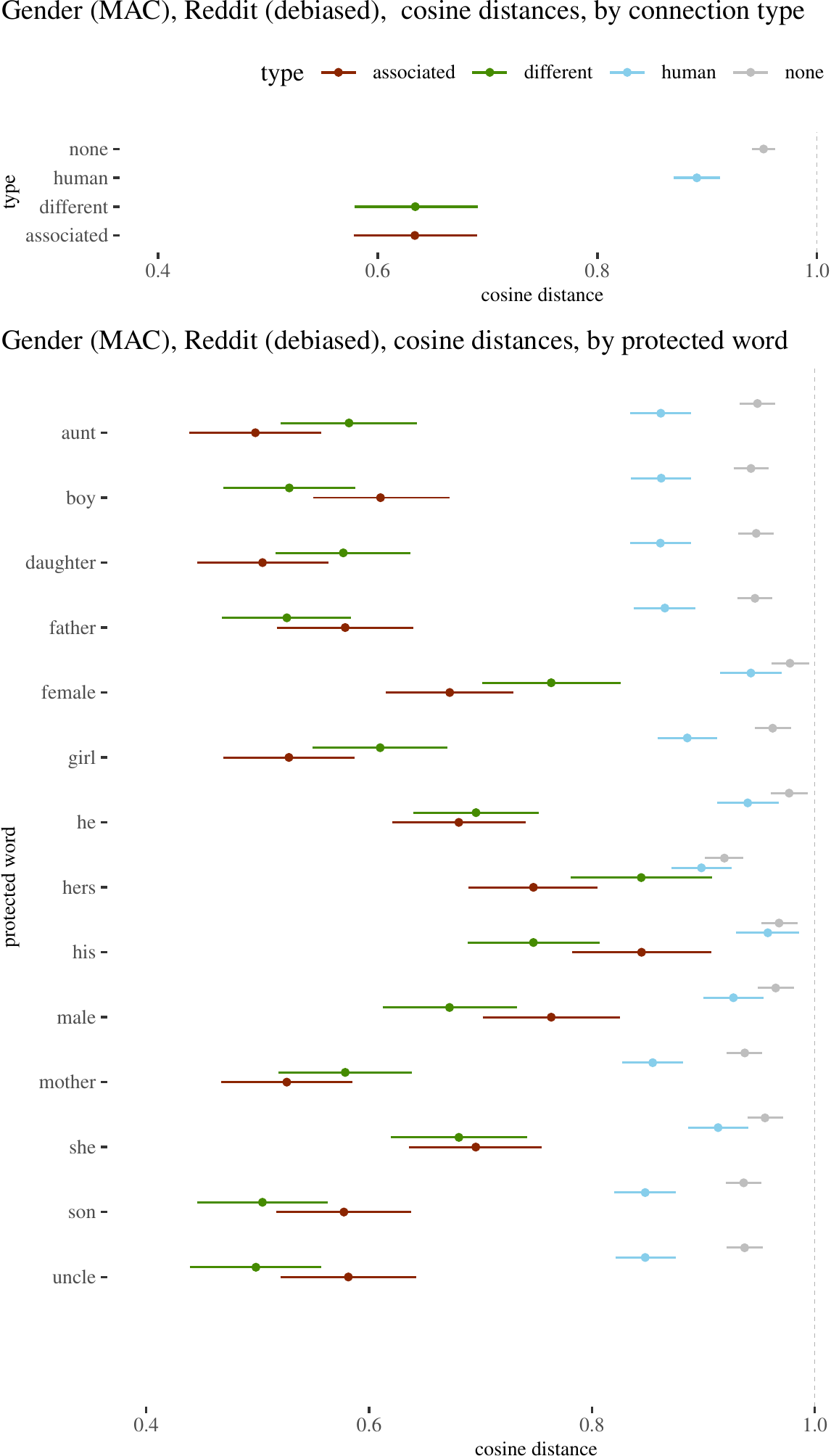} \end{center}
\caption{Mean with 89\% highest posterior density intervals for gender after debiasing.}
\label{fig:empiricalDebiased}
\end{figure}

Bayesian analyses and visualizations thereof can be also handy when it
comes to the investigation of the effect that debiasing has on the
embedding space. In Figures \ref{fig:empiricalPriorToDebiasing} and
\ref{fig:empiricalDebiased} we see an example of two visualizations
depicting the difference in means with 89\% highest posterior density
intervals before and after applying debiasing (the remaining
visualizations are in the Appendix).

\begin{itemize}
\item
  In \emph{Gender (Reddit)}, minor differences between different and
  associated predicates end up being smaller. However, this is not
  achieved by any major change in the relative positions of associated
  and different predicates with respect to protected words, but rather
  by shifting them jointly together. The only protected word for which a
  major difference is noticeable is \emph{hers}.
\item
  In \emph{Religion (Reddit)} debiasing aligns general coefficients for
  all groups together, all of them getting closer to where neutral words
  were prior to debiasing (this is true also for human predicates in
  general, which intuitively did not require debiasing). For some
  protected words such as \emph{christian}, \emph{jew}, the proximity
  ordering between associated and different predicates has been
  reversed, and most of the distances shifted a bit towards 1 (sometimes
  even beyond, such as predicates associated with the word
  \emph{quran}), but for most protected words, the relative differences
  between the coefficient did not change much (for instance, there is no
  change in the way the protected word \emph{muslim} is mistreated).
\item
  For \emph{Race (Reddit)}, general coefficients for different and
  associated predicates became aligned. However, most of the changes
  roughly preserve the structure of bias for particular protected words
  with minor exceptions, such as making the proximities of different
  predicates for protected words \emph{asian} and \emph{asia} much lower
  than associated predicates, which is the main factor responsible for
  the alignment of the general level coefficients.
\end{itemize}

In general, debiasing might end up leading to lower differences between
general level coefficients for associated and different attributes. But
that usually happens without any major change to the structure of the
coefficients for protected words, sporadic extreme and undesirable
changes for some protected words, usually with the side-effect of
changing what happens with neutral and human predicates.

We wouldn't be even able to notice these phenomena had we restricted our
attention to \textsf{MAC} or \textsf{WEAT} scores only. To be able to
diagnose and remove biases at the right level of granularity, we need to
go beyond single metric chasing.

\begin{figure}[H]

\begin{center}\includegraphics[width=0.95\linewidth]{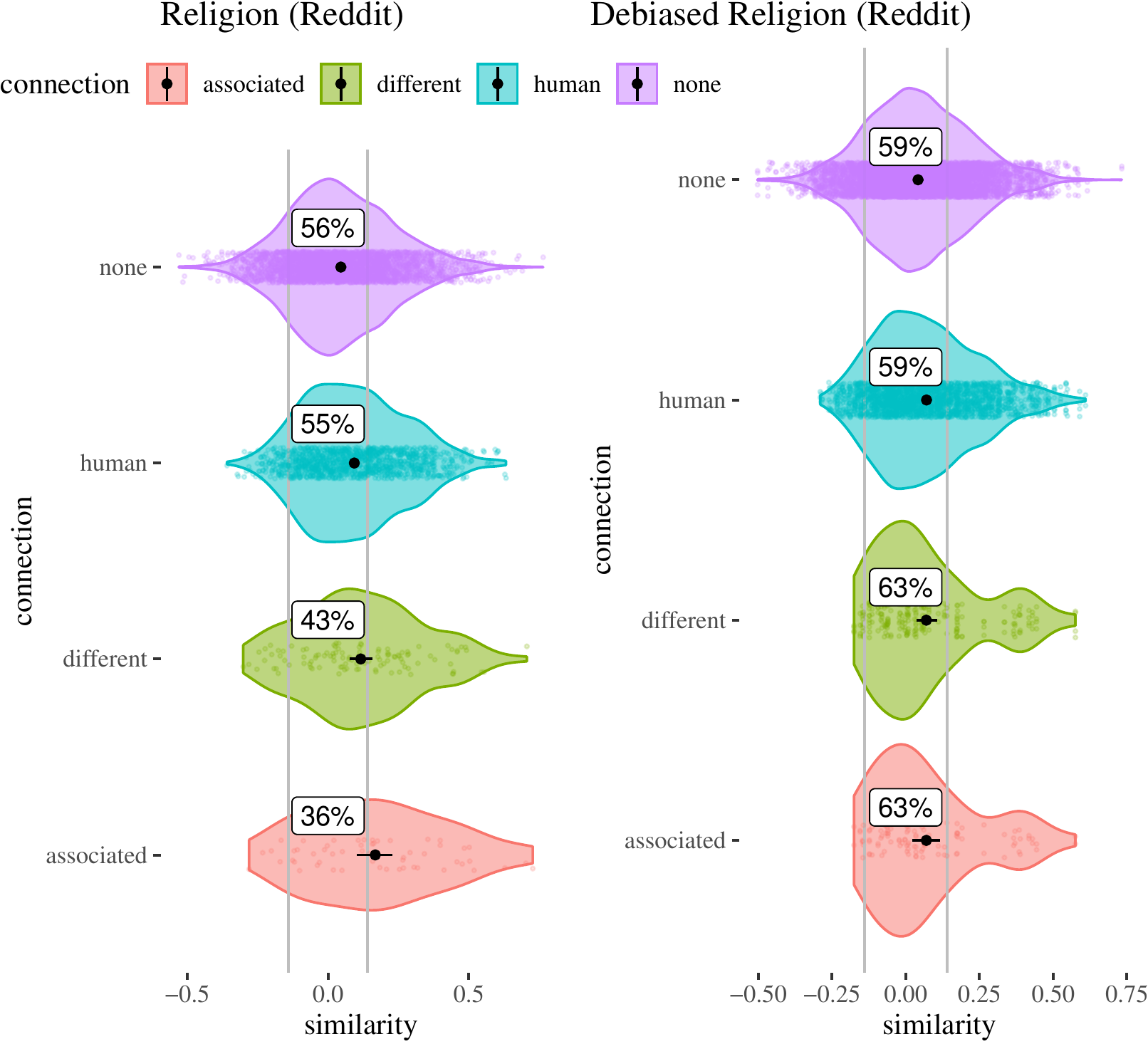} \end{center}
\caption{Empirical distributions of cosine similarities before and after  debiasing  for  the Religion word list  used in  the original paper.}
\label{fig:empiricalDebiased1}
\end{figure}

\begin{figure}[H]

\begin{center}\includegraphics[width=0.95\linewidth]{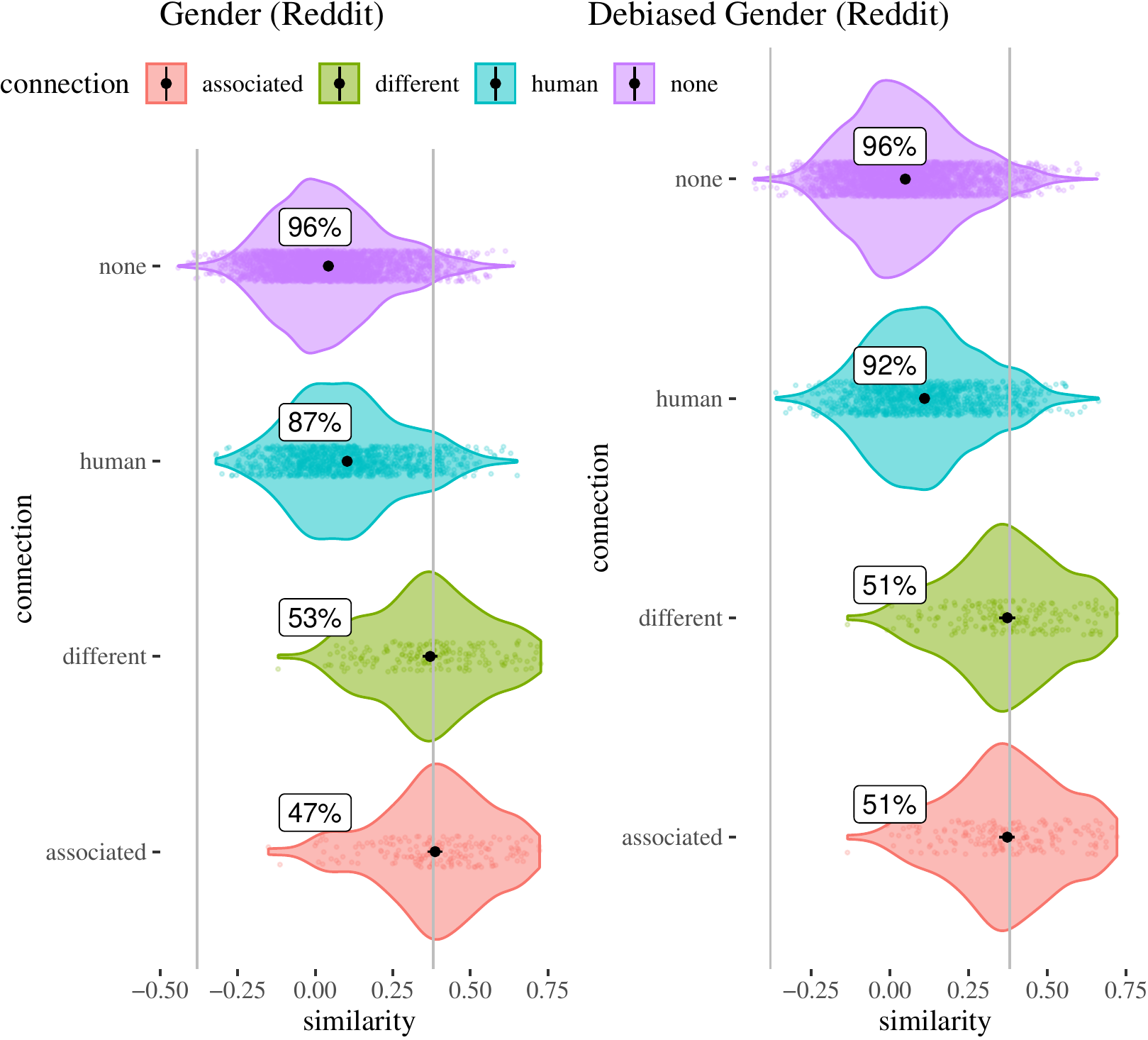} \end{center}
\caption{Empirical distributions of cosine similarities before and after  debiasing for  the Gender word list used in  the original paper.}
\label{fig:empiricalDebiased2}
\end{figure}

\begin{figure}[H]

\begin{center}\includegraphics[width=0.95\linewidth]{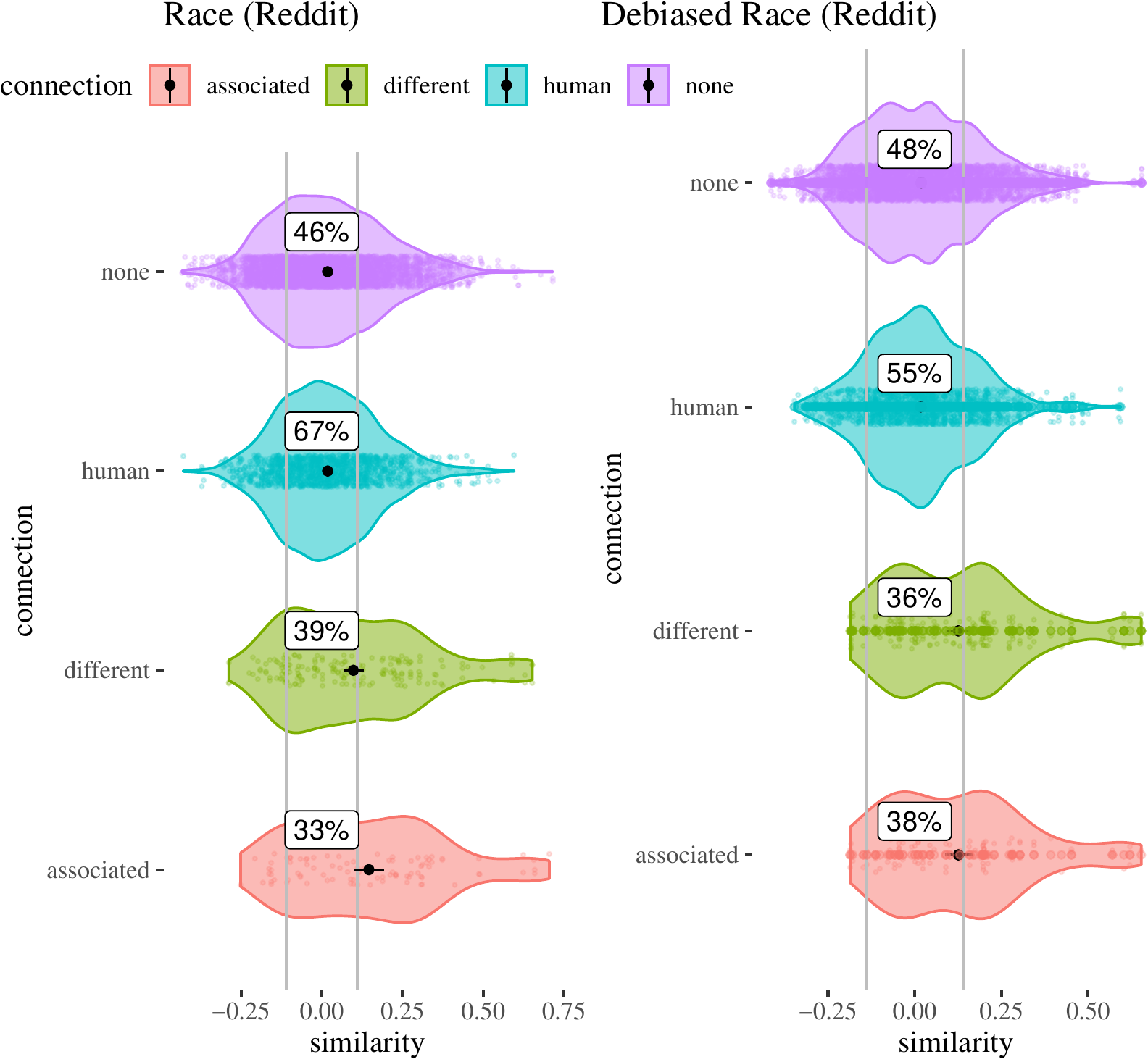} \end{center}
\caption{Empirical distributions of cosine similarities before and after  debiasing for the Race word list used in  the original paper.}
\label{fig:empiricalDebiased3}
\end{figure}

In Figures \ref{fig:empiricalDebiased1}-\ref{fig:empiricalDebiased3} we
inspect the empirical distributions for the debiased embeddings.
Comparing the results to the original embedding, one may notice that for
the Religion group, the neutral and human distribution has changed
slightly. Before within the ``correct'' cosine similarity boundaries,
there were 56\% of neutral and 55\% of human word lists. After the
debiasing, the values changed to 59\% (for neutral) and 59\% (for human).
The different and associated word lists were more influenced. The
general shape of both distributions is less stretched. Before debiasing
43\% of the different word lists and 35\% of the associated word lists
were within the accepted boundaries. After the embedding manipulation,
the percentage has increased for both lists to 63\%. Visualization for
Gender group illustrates almost no change for the neutral and human word
lists before and after debiasing. The values for different and
associated word lists are also barely impacted by the embedding
modification. In the Race group, the percentage within the boundaries
for neutral and associated word lists has increased. The opposite
happened for human and different word lists, where the percentage of
``correct'' cosine similarity dropped from 67\% to 55\% (human) and from
39\% to 36\% (different).

\hypertarget{related-works-and-conclusions}{%
\section{Related works and
conclusions}\label{related-works-and-conclusions}}

\label{sec:related}

There are a few related papers, whose discussion goes beyond the scope
of this paper:

\begin{itemize}
\item
  {[}21{]} employ Bayesian methods to estimate uncertainty in NLP tasks,
  but they apply their Bayesian Neural Networks-based method to
  sentiment analysis or named entity recognition, not to bias.
\item
  {[}2{]} correctly argues that a bias estimate should not be expressed
  as a single number without taking into account that the estimate is
  made using a sample of data and therefore has intrinsic uncertainty.
  The authors suggest using Bernstein bounds to gauge the uncertainty in
  terms of confidence intervals. We do not discuss this approach
  extensively, as we think that confidence intervals are quite
  problematic for several reasons, among others the confusing
  interpretability. We do not think that Bernstein bounds provide the
  best solution to the problem. Applying this method to a popular
  WinoBias dataset leads to the conclusion that more than 11903 samples
  are needed to claim a 95\% confidence interval for a bias estimate.
  This amount vastly exceeds the existing word lists for bias
  estimation. We propose a more realistic Bayesian method. Our
  conclusion is still that the word lists are sometimes too small, but
  at least they allow for gauging uncertainty as we go on to improve our
  methodology and extend the lists gradually.
\item
  {[}20{]} criticize some existing bias metrics such as \textsf{MAC} or
  \textsf{WEAT} on the grounds of them not satisfying some general
  formal principles, such as magnitude-comparability, and they propose a
  modification, called \textsf{SAME}.
\item
  {[}14{]} develop a generalization of \textsf{WEAT} meant to apply to
  sets of sentences, \textsf{SEAT}, which basically applies
  \textsf{WEAT} to vector representations of sentences. The authors,
  however, still pre-average and play the game of finding a
  single-number metric, so our remarks apply.
\item
  {[}7{]} introduce the Contextualized Embedding Association Test
  Intersectional, meant to apply to dynamic word embeddings and,
  importantly develop methods for intersectional bias detection. The
  measure is a generalization of the WEAT method. The authors do inspect
  a distribution of effect sizes that arises from the consideration of
  various possible contexts, but they continue to standardize the
  difference in averaged means and use a single-number summary: the
  weighted mean of the effect sizes thus understood. The method,
  admittedly, deserves further evaluation which goes beyond the scope of
  this paper.
\end{itemize}

\vspace{2mm}

To summarize, a Bayesian data analysis with hierarchical models of
cosine distances between protected words, control group words, and
stereotypical attributes provides more modest and realistic assessment
of the uncertainty involved. It reveals that much complexity is hidden
when one instead chases single bias metrics present in the literature. After introducing the method, we applied it to multiple word embeddings
and results of supposed debiasing, putting forward some general
observations that are not exactly in line with the usual picture painted
in terms of \textsf{WEAT} or \textsf{MAC} (and the problem generalizes
to any approach that focuses on chasing a single numeric metric): the
word list sizes and sample sizes used in the studies are usually small.
Posterior density intervals are fairly wide. Often the differences
between associated, different and neutral human predicates, are not very
impressive. Also, a preliminary inspection suggests that the
desirability of changes obtained by the usual debiasing methods is
debatable. The tools that we propose, however, allow for a more
fine-grained and multi-level evaluation of bias and debiasing in
language models without losing modesty about the uncertainties involved. The short, general, and somewhat disappointing lesson here is this: things
are complicated. Instead of chasing single-number metrics, we should
rather devote attention to more nuanced analysis.

\newpage

\hypertarget{references}{%
\section*{References}\label{references}}
\addcontentsline{toc}{section}{References}

\hypertarget{refs}{}
\begin{CSLReferences}{0}{0}
\leavevmode\vadjust pre{\hypertarget{ref-Bolukbasi2016man}{}}%
\CSLLeftMargin{{[}1{]} }%
\CSLRightInline{Tolga Bolukbasi, Kai-Wei Chang, James Y. Zou, Venkatesh
Saligrama, and Adam Kalai. 2016. Man is to computer programmer as woman
is to homemaker? Debiasing word embeddings. \emph{CoRR} abs/1607.06520,
(2016). Retrieved from \url{http://arxiv.org/abs/1607.06520}}

\leavevmode\vadjust pre{\hypertarget{ref-Ethayarajh2020Bernstein}{}}%
\CSLLeftMargin{{[}2{]} }%
\CSLRightInline{Kawin Ethayarajh. 2020. Is your classifier actually
biased? Measuring fairness under uncertainty with bernstein bounds.
\emph{CoRR} abs/2004.12332, (2020). Retrieved from
\url{https://arxiv.org/abs/2004.12332}}

\leavevmode\vadjust pre{\hypertarget{ref-Garg2017hundredYears}{}}%
\CSLLeftMargin{{[}3{]} }%
\CSLRightInline{Nikhil Garg, Londa Schiebinger, Dan Jurafsky, and James
Zou. 2017. Word embeddings quantify 100 years of gender and ethnic
stereotypes. \emph{Proceedings of the National Academy of Sciences} 115,
(November 2017).
DOI:https://doi.org/\href{https://doi.org/10.1073/pnas.1720347115}{10.1073/pnas.1720347115}}

\leavevmode\vadjust pre{\hypertarget{ref-Garg2018years}{}}%
\CSLLeftMargin{{[}4{]} }%
\CSLRightInline{Nikhil Garg, Londa Schiebinger, Dan Jurafsky, and James
Zou. 2018. Word embeddings quantify 100 years of gender and ethnic
stereotypes. \emph{Proceedings of the National Academy of Sciences} 115,
16 (April 2018), E3635--E3644.
DOI:https://doi.org/\href{https://doi.org/10.1073/pnas.1720347115}{10.1073/pnas.1720347115}}

\leavevmode\vadjust pre{\hypertarget{ref-Gonen2019lipstick}{}}%
\CSLLeftMargin{{[}5{]} }%
\CSLRightInline{Hila Gonen and Yoav Goldberg. 2019. Lipstick on a pig:
{D}ebiasing methods cover up systematic gender biases in word embeddings
but do not remove them. In \emph{Proceedings of the 2019 conference of
the north {A}merican chapter of the association for computational
linguistics: Human language technologies, volume 1 (long and short
papers)}, Association for Computational Linguistics, Minneapolis,
Minnesota, 609--614.
DOI:https://doi.org/\href{https://doi.org/10.18653/v1/N19-1061}{10.18653/v1/N19-1061}}

\leavevmode\vadjust pre{\hypertarget{ref-gordon2012reporting}{}}%
\CSLLeftMargin{{[}6{]} }%
\CSLRightInline{Jonathan Gordon and Benjamin Durme. 2013. Reporting bias
and knowledge acquisition. In \emph{AKBC 2013 - Proceedings of the 2013
Workshop on Automated Knowledge Base Construction, Co-located with CIKM
2013}, 25--30.
DOI:https://doi.org/\href{https://doi.org/10.1145/2509558.2509563}{10.1145/2509558.2509563}}

\leavevmode\vadjust pre{\hypertarget{ref-Guo2021CEAT}{}}%
\CSLLeftMargin{{[}7{]} }%
\CSLRightInline{Wei Guo and Aylin Caliskan. 2021. Detecting emergent
intersectional biases: Contextualized word embeddings contain a
distribution of human-like biases. In \emph{Proceedings of the 2021
{AAAI}/{ACM} conference on {AI}, ethics, and society}, {ACM}.
DOI:https://doi.org/\href{https://doi.org/10.1145/3461702.3462536}{10.1145/3461702.3462536}}

\leavevmode\vadjust pre{\hypertarget{ref-Hoekstra2014Misinterpretation}{}}%
\CSLLeftMargin{{[}8{]} }%
\CSLRightInline{Rink Hoekstra, Richard D. Morey, Jeffrey N. Rouder, and
Eric-Jan Wagenmakers. 2014. Robust misinterpretation of confidence
intervals. \emph{Psychonomic Bulletin \& Review} 21, 5 (October 2014),
1157--1164.
DOI:https://doi.org/\href{https://doi.org/10.3758/s13423-013-0572-3}{10.3758/s13423-013-0572-3}}

\leavevmode\vadjust pre{\hypertarget{ref-Caliskan2017semanticsBiases}{}}%
\CSLLeftMargin{{[}9{]} }%
\CSLRightInline{Aylin Caliskan Islam, Joanna J. Bryson, and Arvind
Narayanan. 2016. Semantics derived automatically from language corpora
necessarily contain human biases. \emph{CoRR} abs/1608.07187, (2016).
Retrieved from \url{http://arxiv.org/abs/1608.07187}}

\leavevmode\vadjust pre{\hypertarget{ref-JohnsonValueFree}{}}%
\CSLLeftMargin{{[}10{]} }%
\CSLRightInline{Gabbrielle Johnson. forthcoming. Are algorithms
value-free? Feminist theoretical virtues in machine learning.
\emph{Journal Moral Philosophy} (forthcoming).}

\leavevmode\vadjust pre{\hypertarget{ref-kruschke2015bayesian}{}}%
\CSLLeftMargin{{[}11{]} }%
\CSLRightInline{John Kruschke. 2015. \emph{Doing bayesian data analysis
(second edition)}. Academic Press, Boston.}

\leavevmode\vadjust pre{\hypertarget{ref-Lauscher2019multidimensional}{}}%
\CSLLeftMargin{{[}12{]} }%
\CSLRightInline{Anne Lauscher and Goran Glavas. 2019. Are we
consistently biased? Multidimensional analysis of biases in
distributional word vectors. \emph{CoRR} abs/1904.11783, (2019).
Retrieved from \url{http://arxiv.org/abs/1904.11783}}

\leavevmode\vadjust pre{\hypertarget{ref-Manzini2019blackToCriminal}{}}%
\CSLLeftMargin{{[}13{]} }%
\CSLRightInline{Thomas Manzini, Yao Chong Lim, Yulia Tsvetkov, and Alan
W Black. 2019. Black is to criminal as caucasian is to police: Detecting
and removing multiclass bias in word embeddings. Retrieved from
\url{https://arxiv.org/abs/1904.04047}}

\leavevmode\vadjust pre{\hypertarget{ref-may-etal-2019-measuring}{}}%
\CSLLeftMargin{{[}14{]} }%
\CSLRightInline{Chandler May, Alex Wang, Shikha Bordia, Samuel R.
Bowman, and Rachel Rudinger. 2019. On measuring social biases in
sentence encoders. In \emph{Proceedings of the 2019 conference of the
north {A}merican chapter of the association for computational
linguistics: Human language technologies, volume 1 (long and short
papers)}, Association for Computational Linguistics, Minneapolis,
Minnesota, 622--628.
DOI:https://doi.org/\href{https://doi.org/10.18653/v1/N19-1063}{10.18653/v1/N19-1063}}

\leavevmode\vadjust pre{\hypertarget{ref-statrethinkingbook2020}{}}%
\CSLLeftMargin{{[}15{]} }%
\CSLRightInline{Richard McElreath. 2020. \emph{Statistical rethinking: A
bayesian course with examples in r and stan, 2nd edition} (2nd ed.). CRC
Press. Retrieved from
\url{http://xcelab.net/rm/statistical-rethinking/}}

\leavevmode\vadjust pre{\hypertarget{ref-Mikolov2013efficient}{}}%
\CSLLeftMargin{{[}16{]} }%
\CSLRightInline{Tomas Mikolov, Kai Chen, Greg Corrado, and Jeffrey Dean.
2013. Efficient estimation of word representations in vector space.
DOI:https://doi.org/\href{https://doi.org/10.48550/ARXIV.1301.3781}{10.48550/ARXIV.1301.3781}}

\leavevmode\vadjust pre{\hypertarget{ref-Morey2015confidenceFallacy}{}}%
\CSLLeftMargin{{[}17{]} }%
\CSLRightInline{Richard Morey, Rink Hoekstra, Jeffrey Rouder, Michael
Lee, and EJ Wagenmakers. 2015. The fallacy of placing confidence in
confidence intervals. \emph{Psychonomic Bulletin \& Review} (September
2015).}

\leavevmode\vadjust pre{\hypertarget{ref-Nissim2020fair}{}}%
\CSLLeftMargin{{[}18{]} }%
\CSLRightInline{Malvina Nissim, Rik van Noord, and Rob van der Goot.
2020. Fair is better than sensational: Man is to doctor as woman is to
doctor. \emph{Computational Linguistics} 46, 2 (June 2020), 487--497.
DOI:https://doi.org/\href{https://doi.org/10.1162/coli_a_00379}{10.1162/coli\_a\_00379}}

\leavevmode\vadjust pre{\hypertarget{ref-Nosek2002harvesting}{}}%
\CSLLeftMargin{{[}19{]} }%
\CSLRightInline{Brian A. Nosek, Mahzarin R. Banaji, and Anthony G.
Greenwald. 2002. Harvesting implicit group attitudes and beliefs from a
demonstration web site. \emph{Group Dynamics: Theory, Research, and
Practice} 6, 1 (2002), 101--115.
DOI:https://doi.org/\href{https://doi.org/10.1037/1089-2699.6.1.101}{10.1037/1089-2699.6.1.101}}

\leavevmode\vadjust pre{\hypertarget{ref-schruxf6der2021evaluating}{}}%
\CSLLeftMargin{{[}20{]} }%
\CSLRightInline{Sarah Schröder, Alexander Schulz, Philip Kenneweg,
Robert Feldhans, Fabian Hinder, and Barbara Hammer. 2021. Evaluating
metrics for bias in word embeddings. Retrieved from
\url{https://arxiv.org/abs/2111.07864}}

\leavevmode\vadjust pre{\hypertarget{ref-DBLP:journalsux2fcorrux2fabs-1811-07253}{}}%
\CSLLeftMargin{{[}21{]} }%
\CSLRightInline{Yijun Xiao and William Yang Wang. 2018. Quantifying
uncertainties in natural language processing tasks. \emph{CoRR}
abs/1811.07253, (2018). Retrieved from
\url{http://arxiv.org/abs/1811.07253}}

\end{CSLReferences}

\newpage
\appendix

\hypertarget{appendix}{%
\section{Appendix}\label{appendix}}

\label{sec:appendix}

\hypertarget{visualizations}{%
\subsection{Visualizations}\label{visualizations}}

\label{appendix:visualizations}

\begin{center}\includegraphics{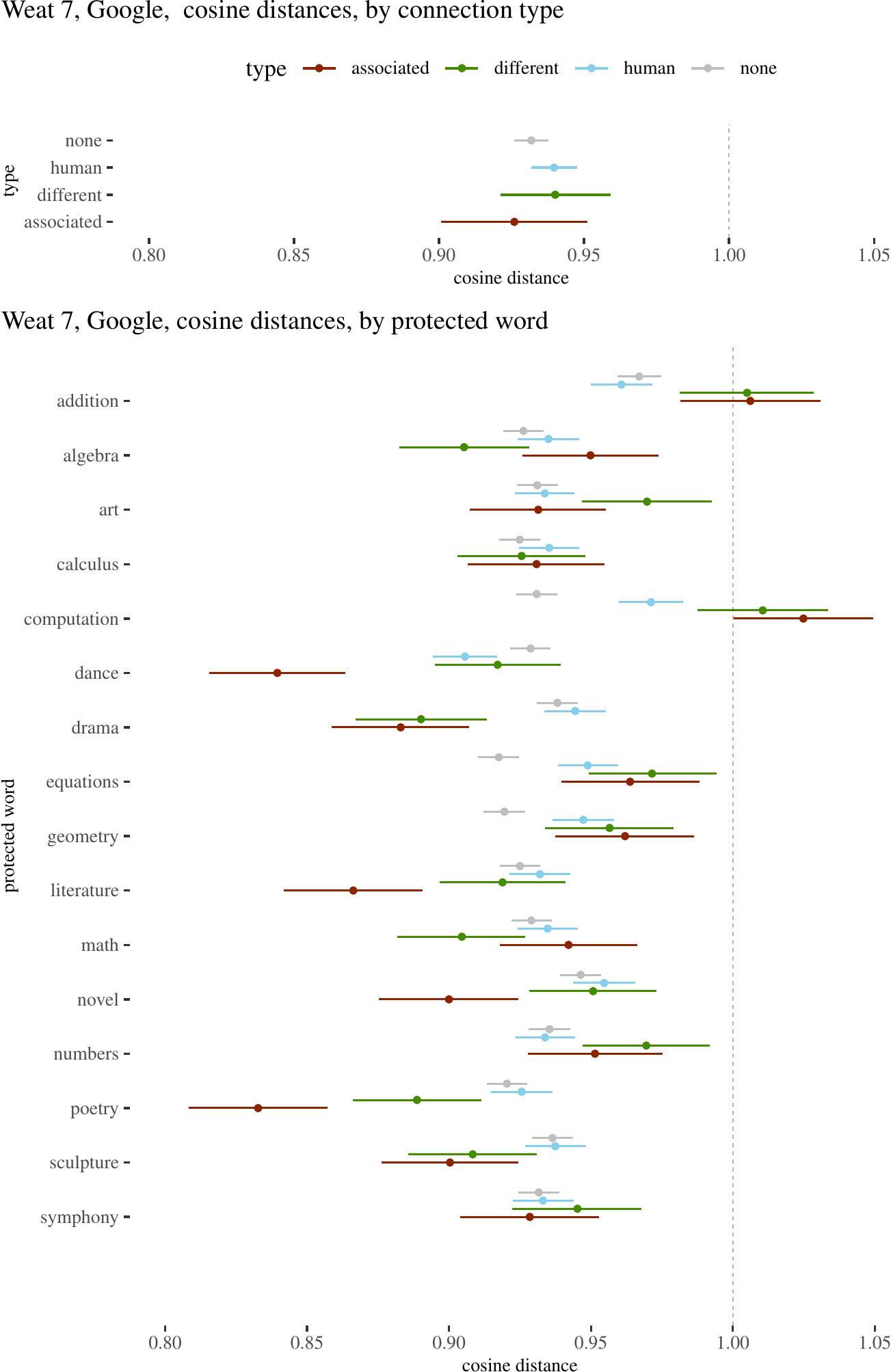} \end{center}

\pagebreak

\begin{center}\includegraphics[width=1.1\linewidth]{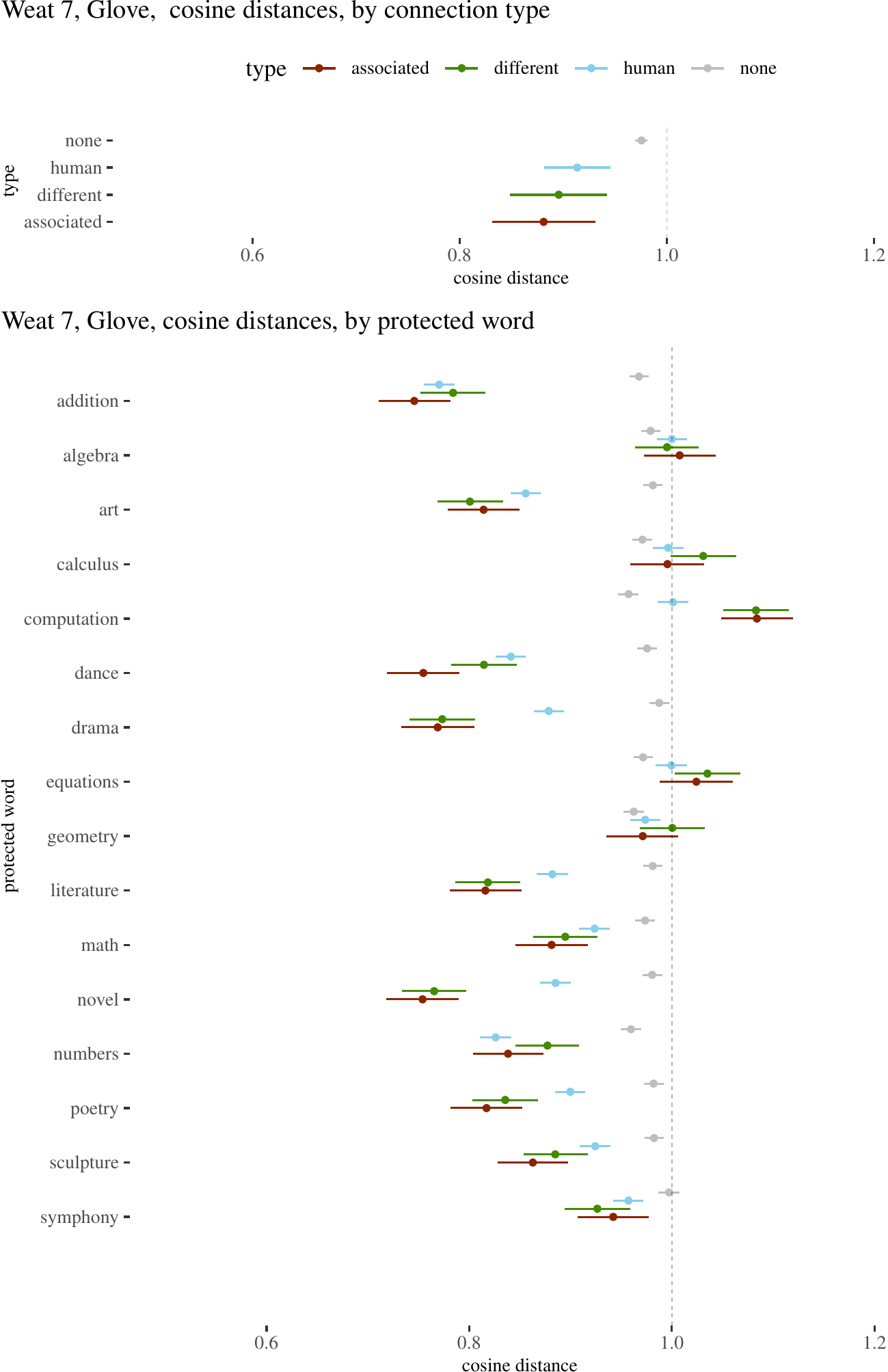} \end{center}

\begin{center}\includegraphics[width=1.1\linewidth]{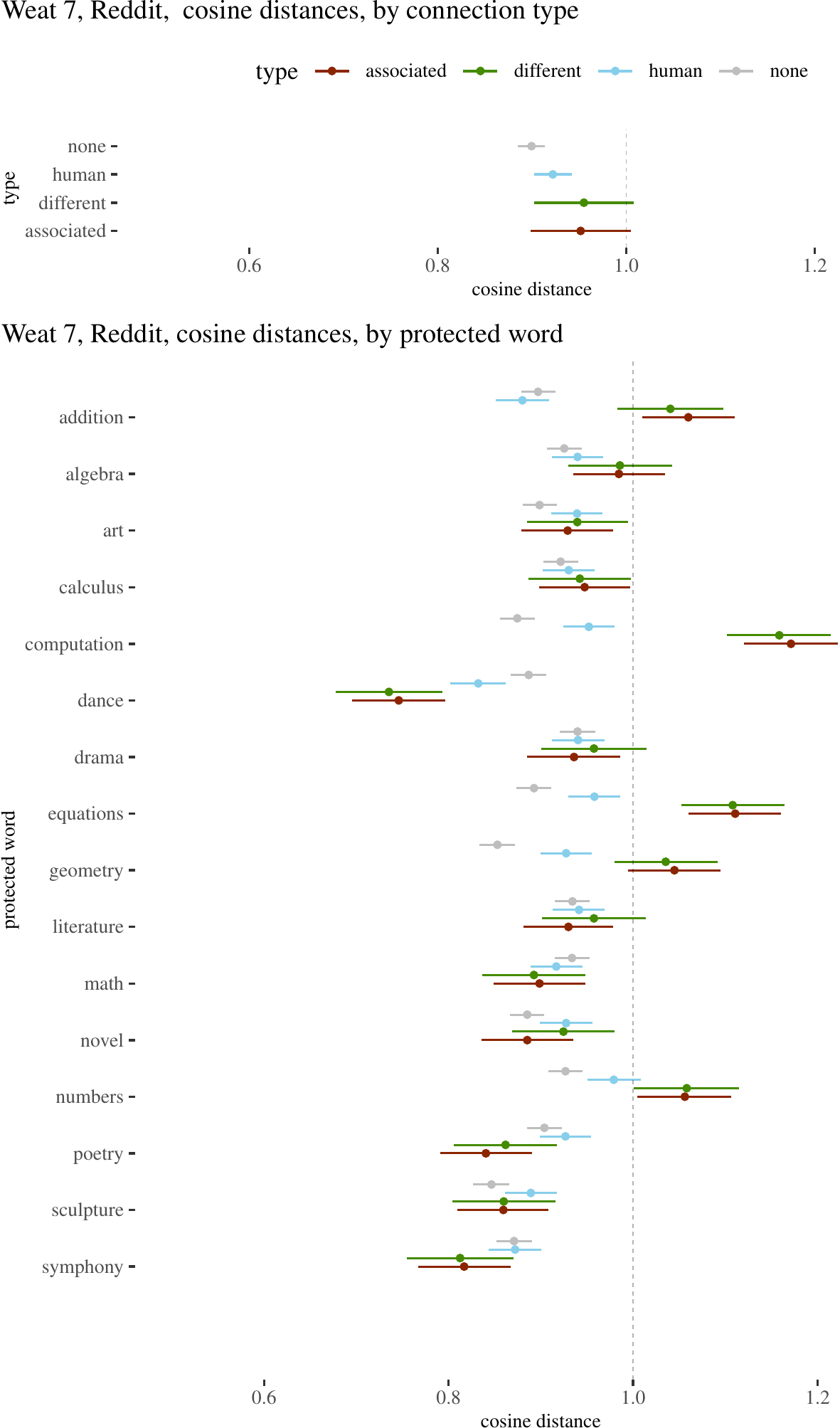} \end{center}

\begin{center}\includegraphics[width=1.1\linewidth]{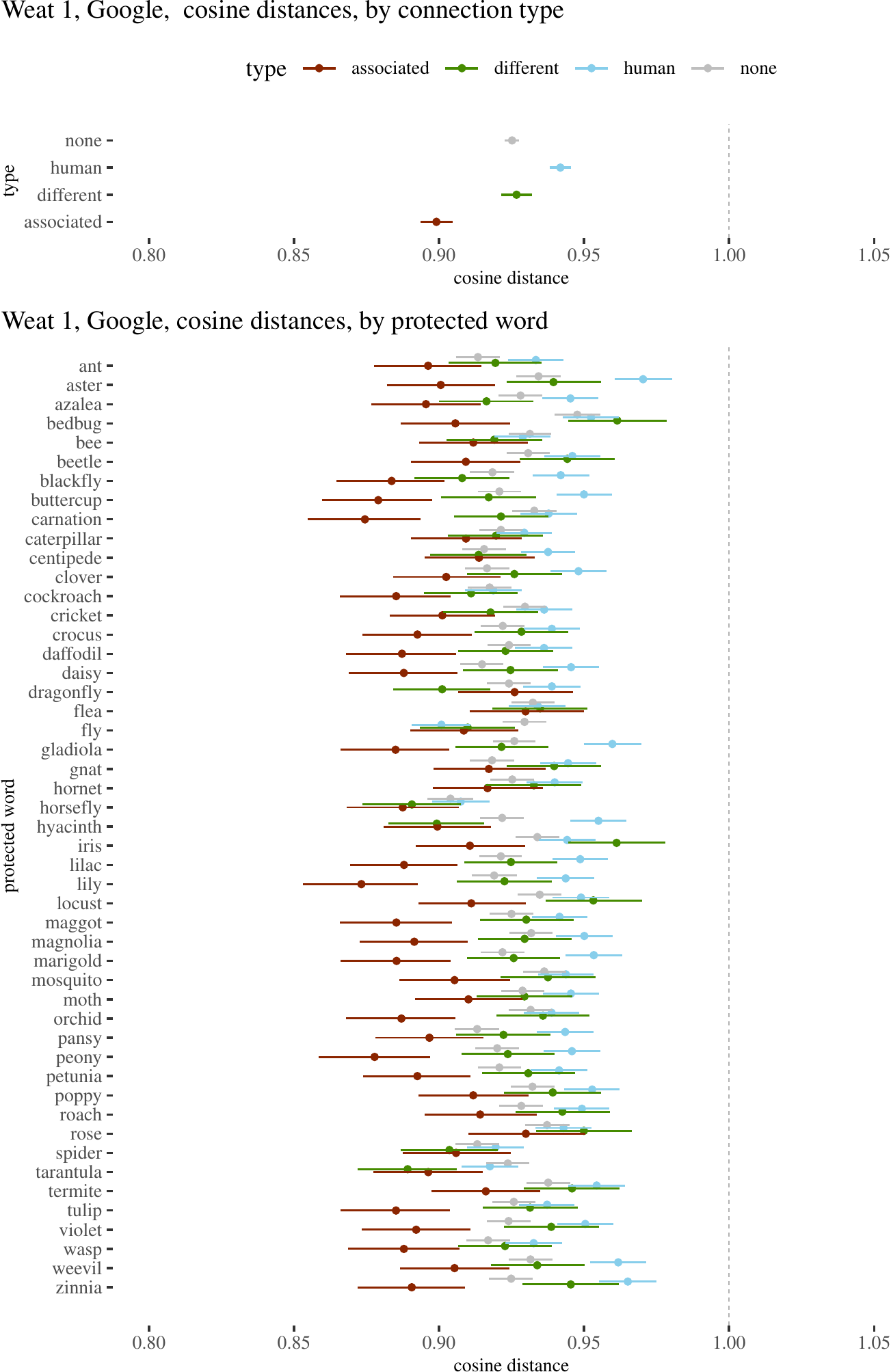} \end{center}

\begin{center}\includegraphics[width=1.1\linewidth]{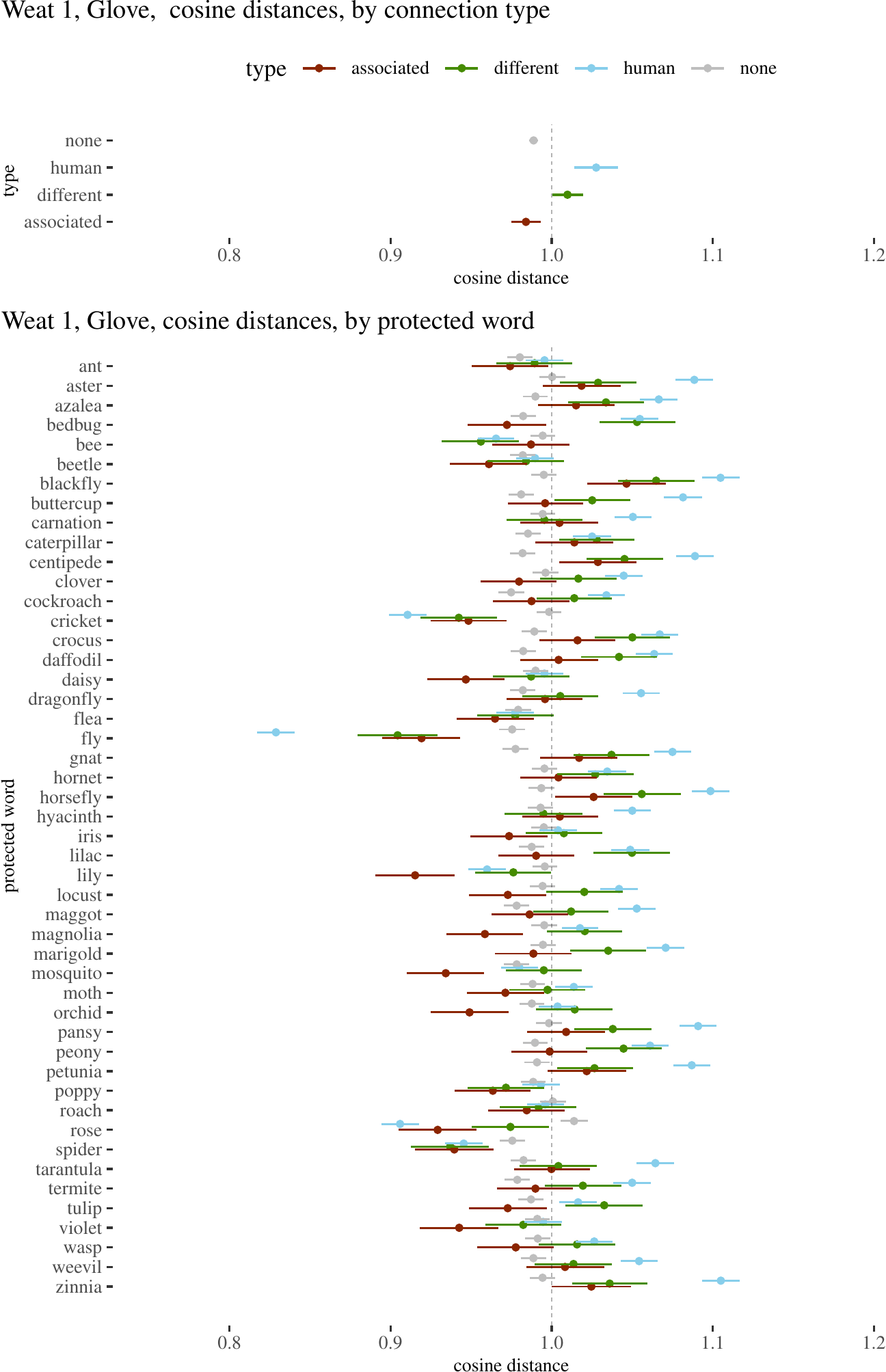} \end{center}

\begin{center}\includegraphics[width=1.1\linewidth]{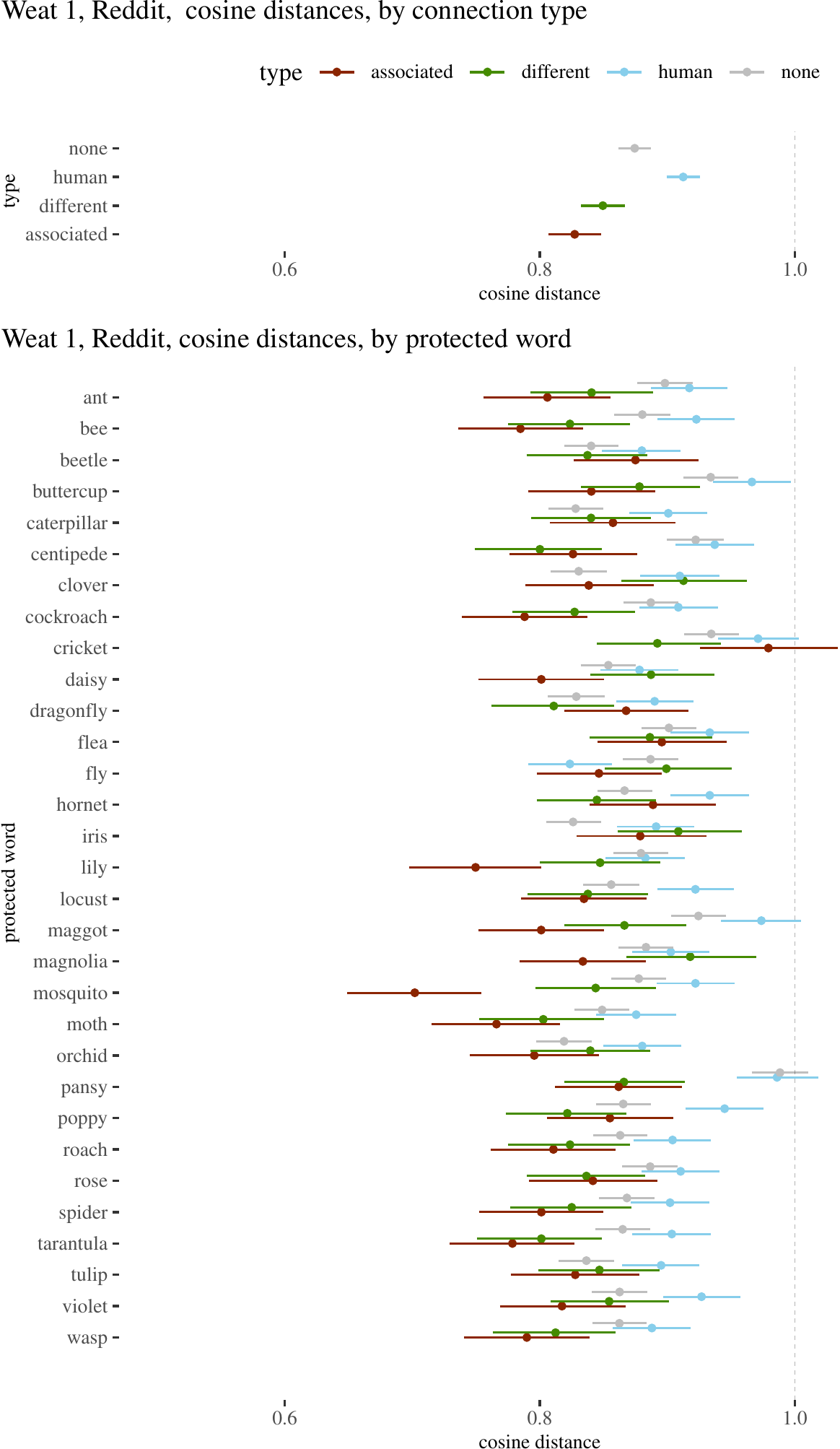} \end{center}

\begin{center}\includegraphics[width=1.1\linewidth]{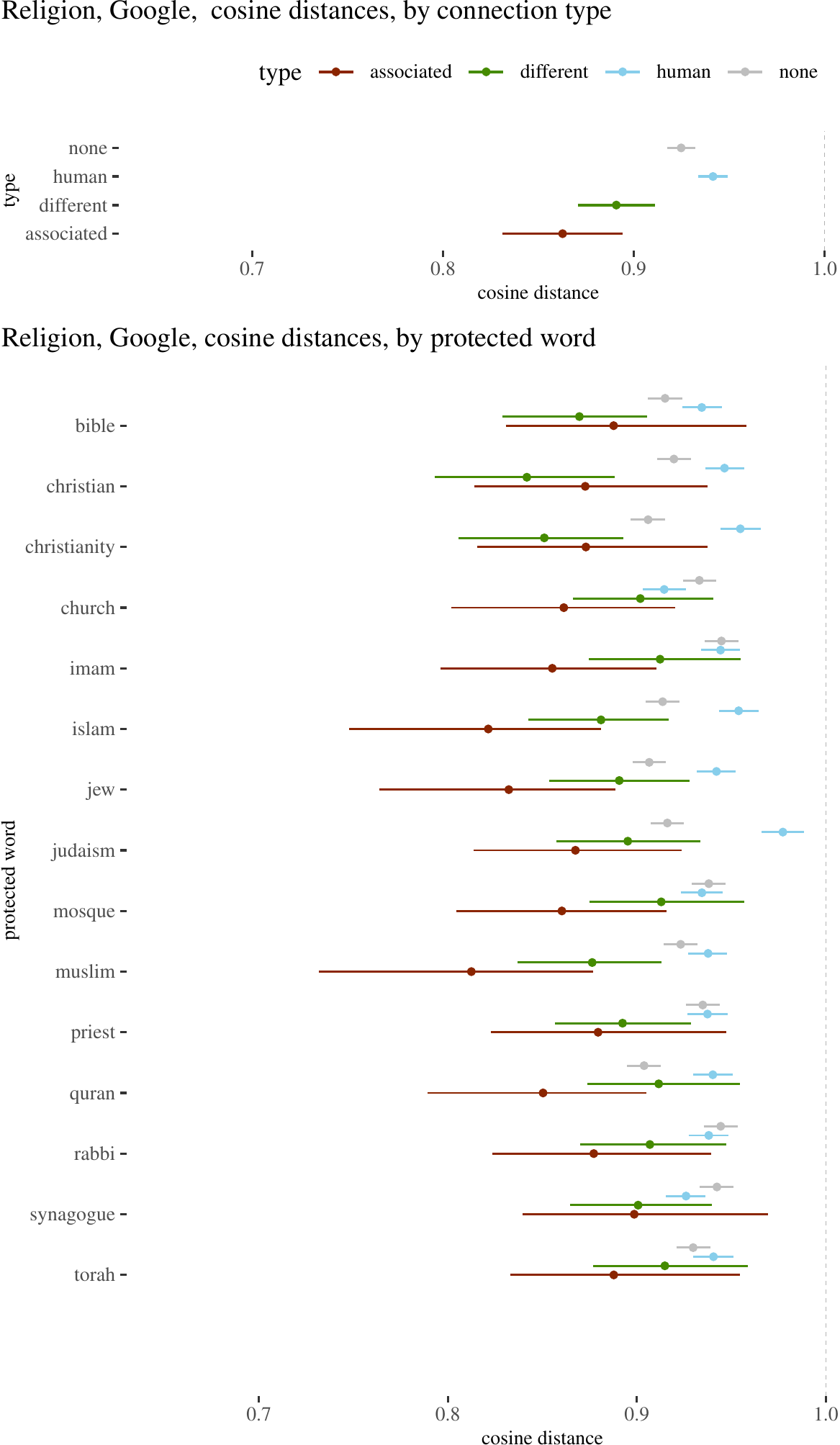} \end{center}

\begin{center}\includegraphics[width=1.1\linewidth]{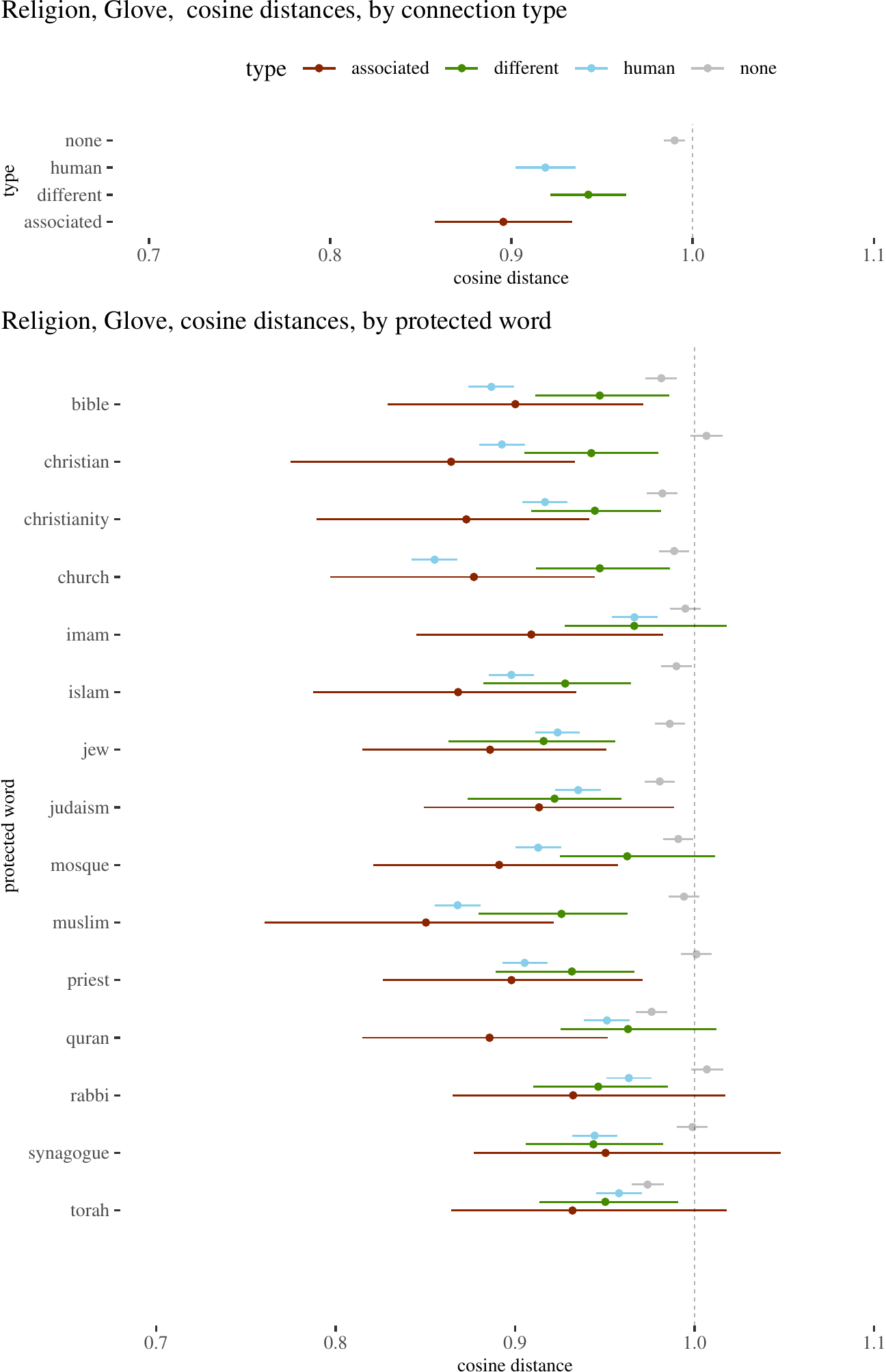} \end{center}

\begin{center}\includegraphics[width=1.1\linewidth]{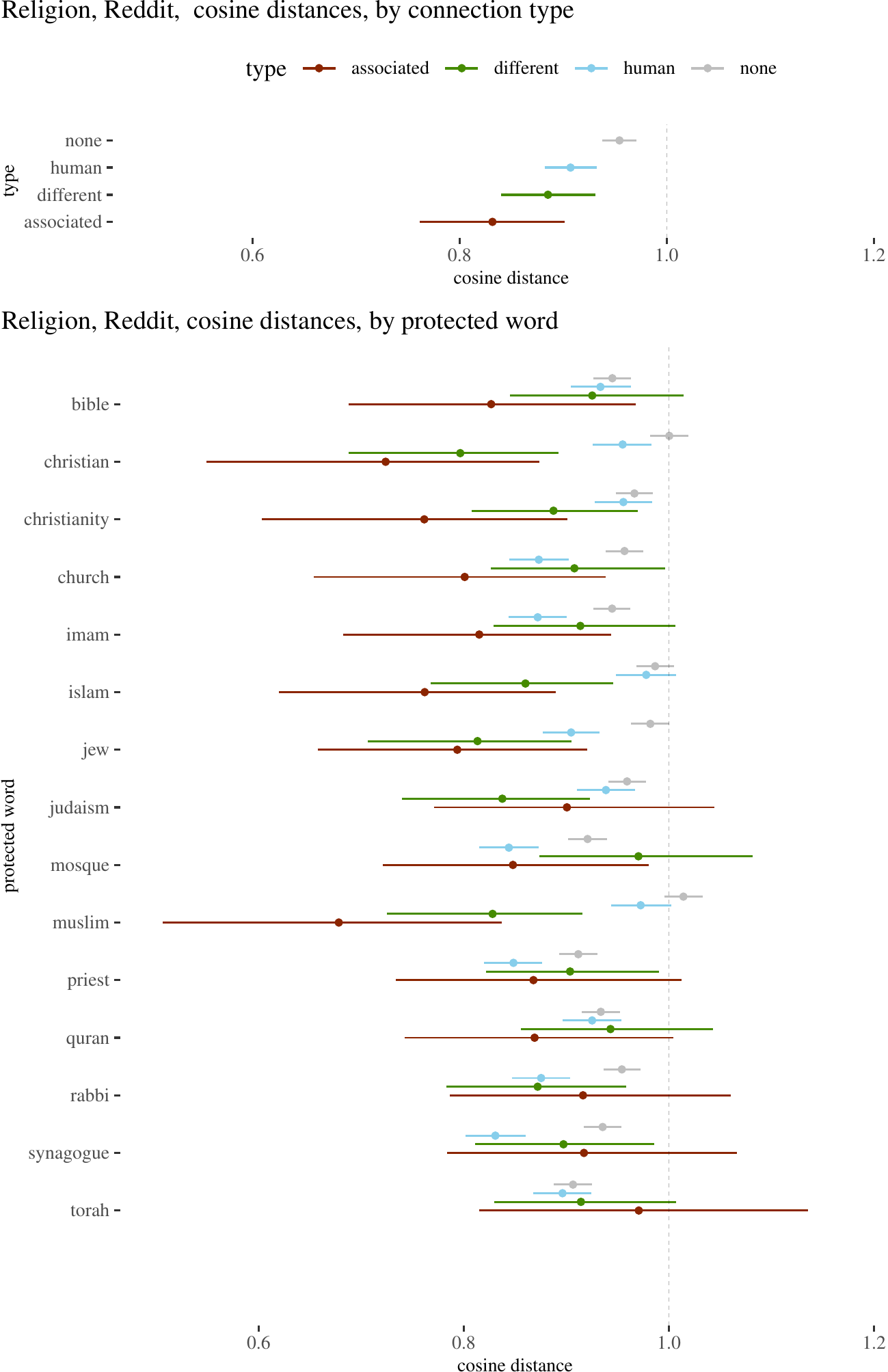} \end{center}

\begin{center}\includegraphics[width=1.1\linewidth]{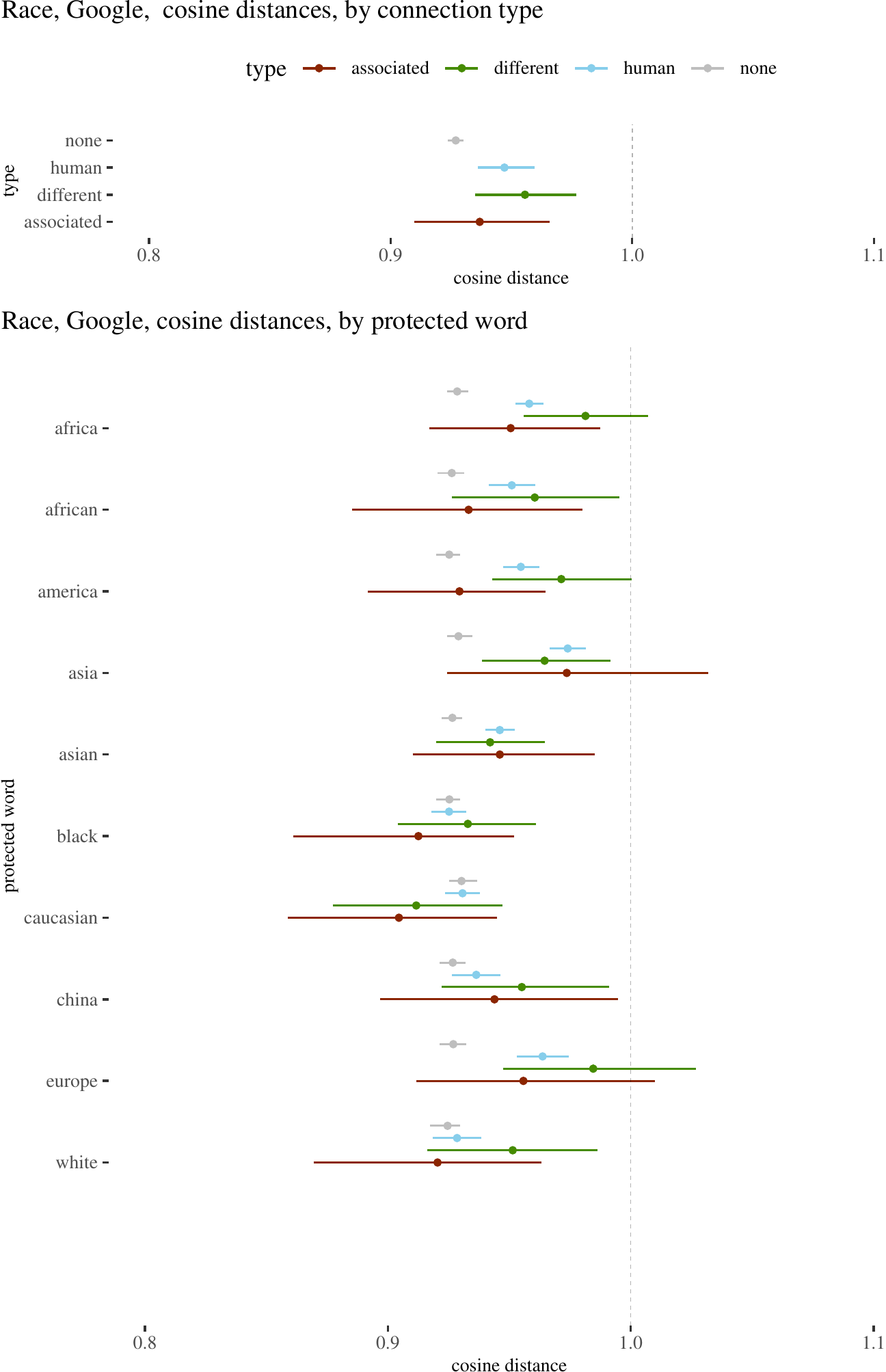} \end{center}

\begin{center}\includegraphics[width=1.1\linewidth]{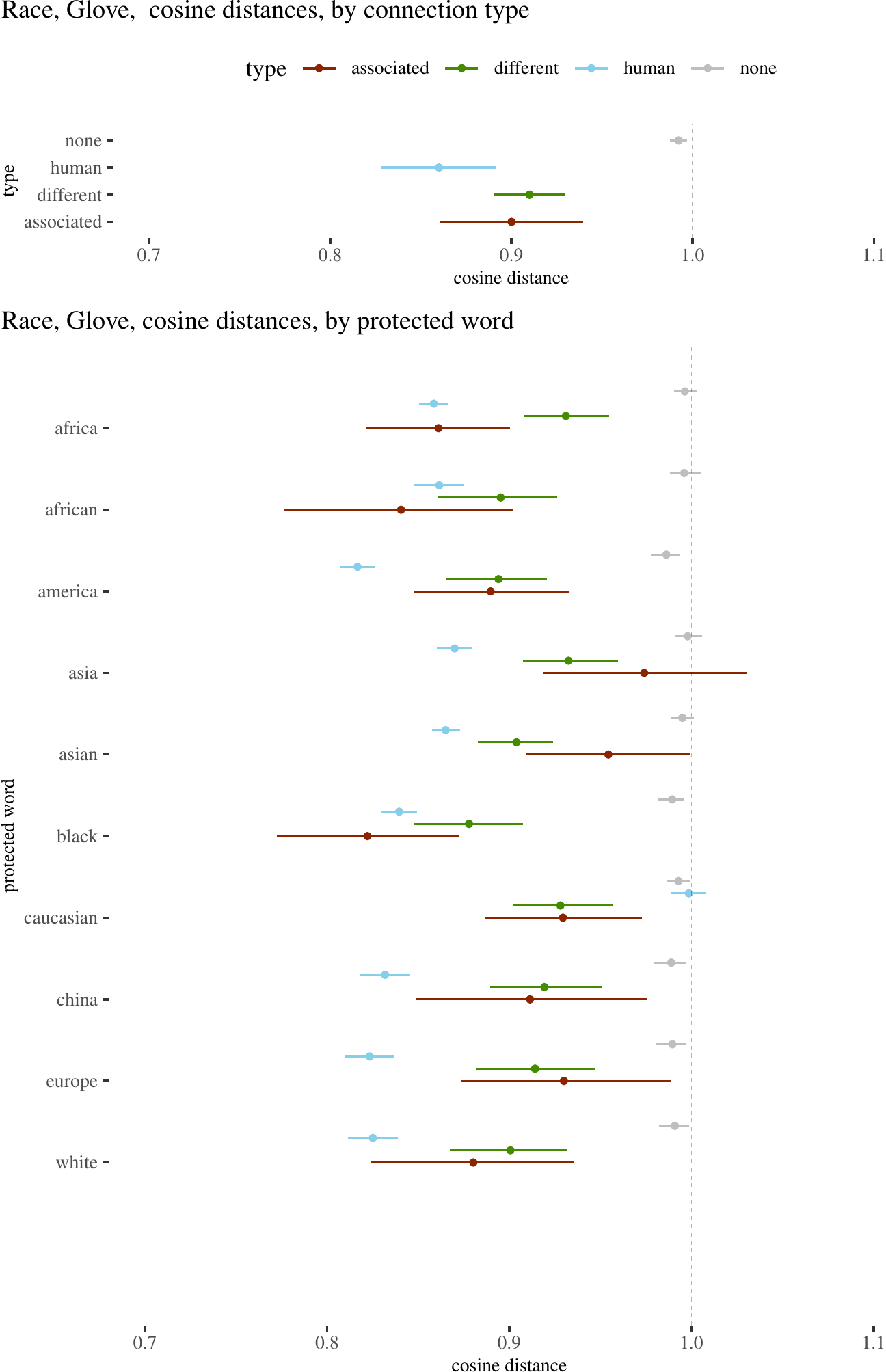} \end{center}

\begin{center}\includegraphics[width=1.1\linewidth]{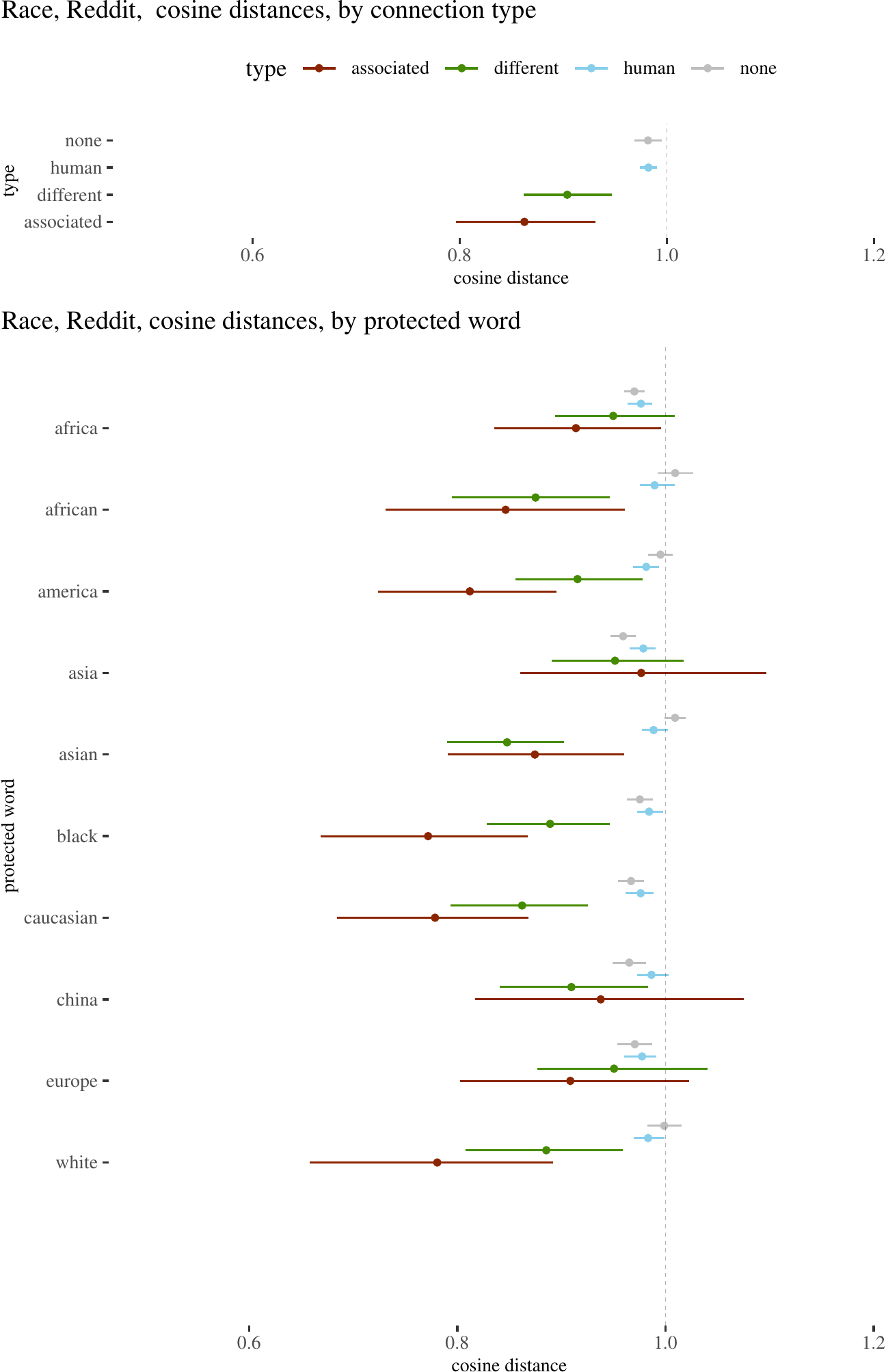} \end{center}

\begin{center}\includegraphics[width=1.1\linewidth]{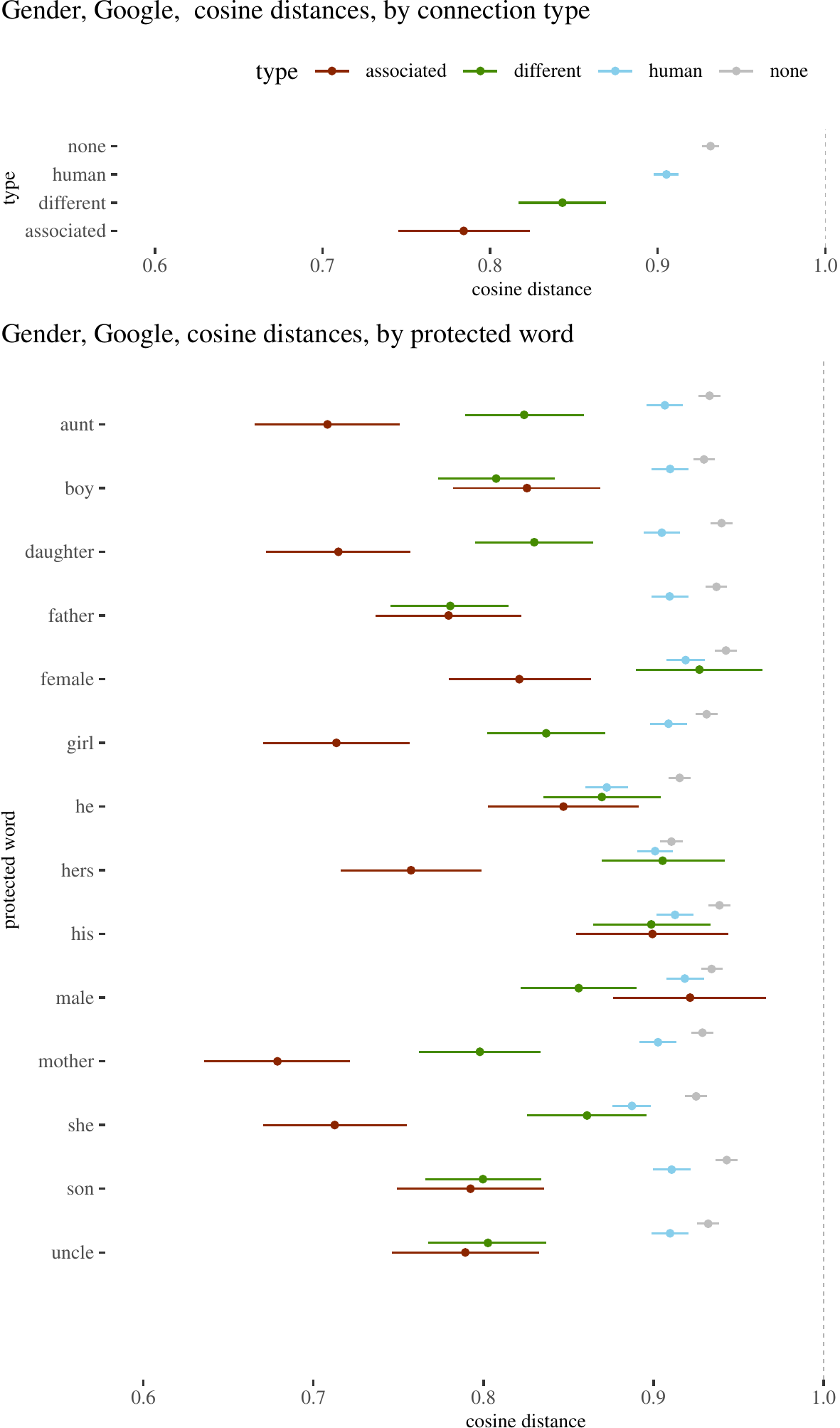} \end{center}

\begin{center}\includegraphics[width=1.1\linewidth]{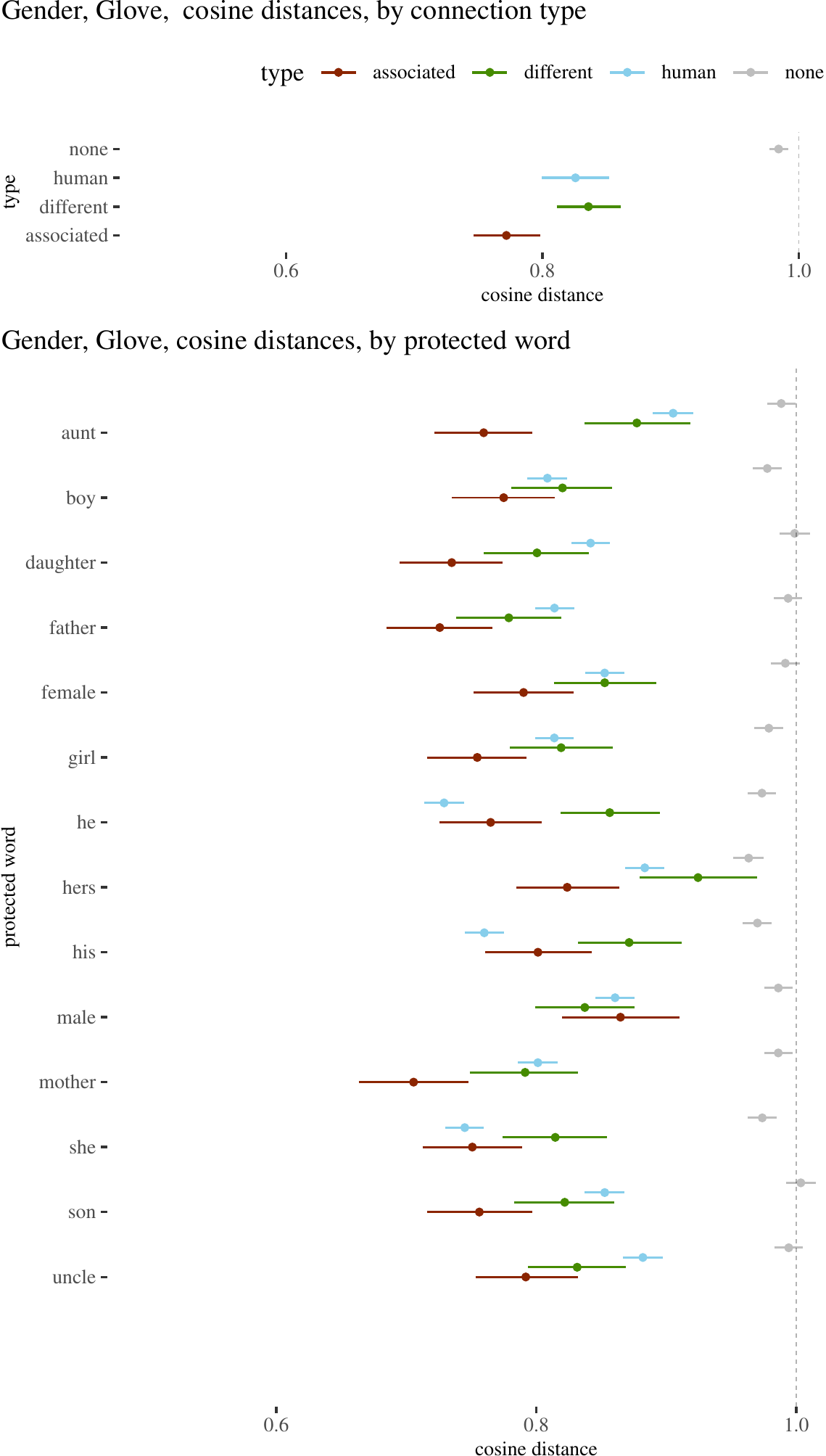} \end{center}

\begin{center}\includegraphics[width=1.1\linewidth]{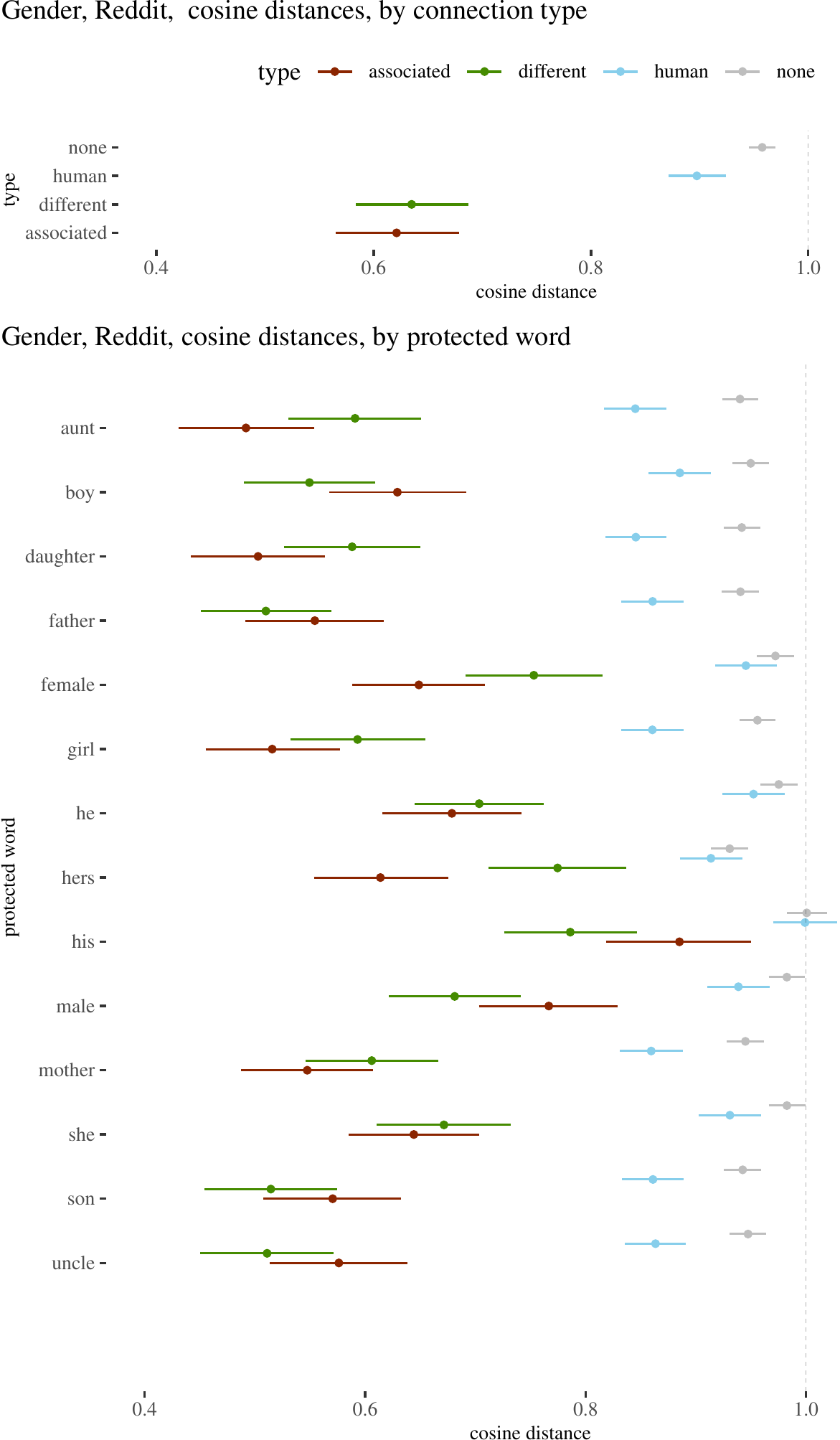} \end{center}

\begin{center}\includegraphics[width=1.1\linewidth]{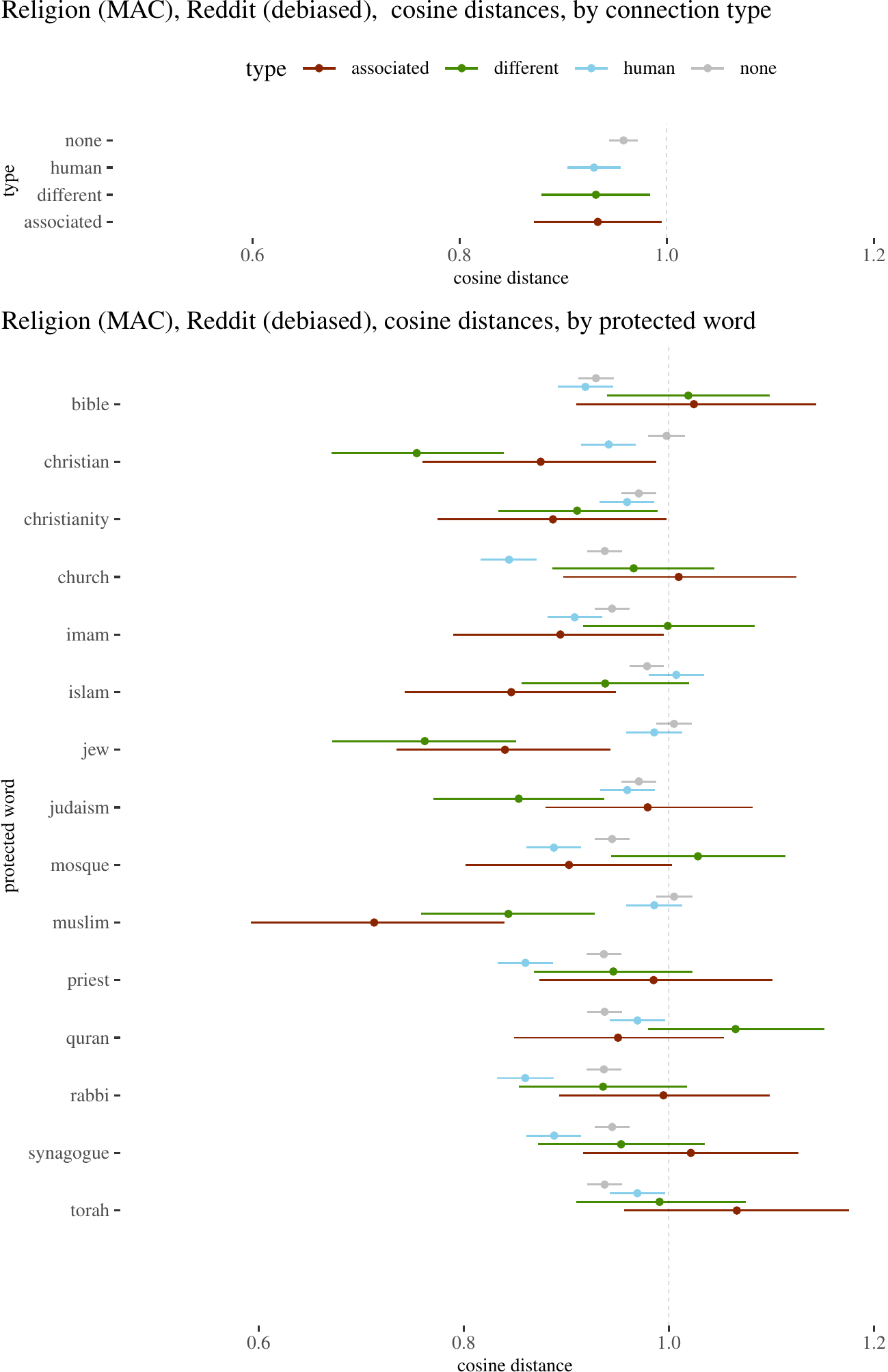} \end{center}

\begin{center}\includegraphics[width=1.1\linewidth]{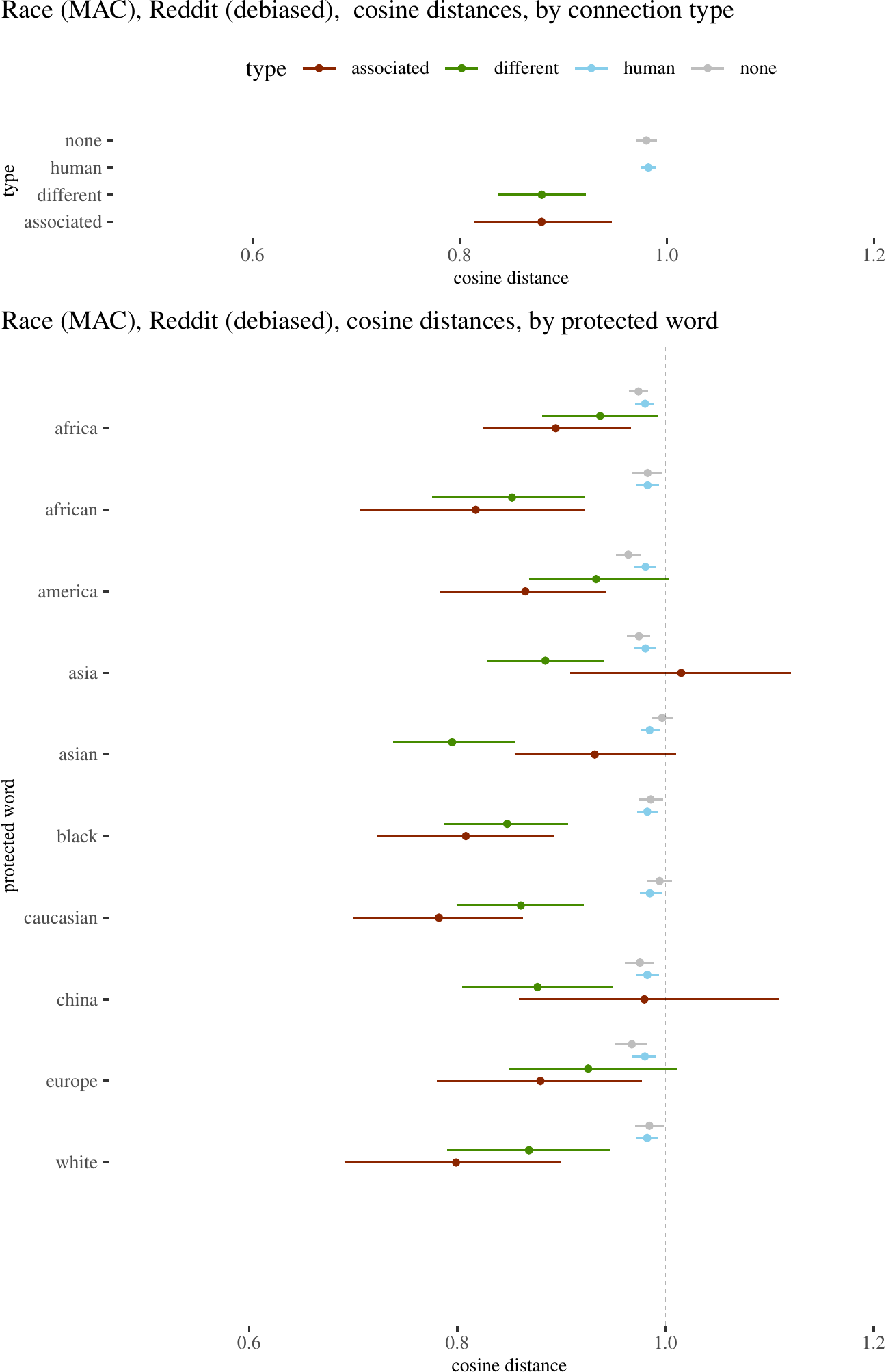} \end{center}

\begin{center}\includegraphics[width=1.1\linewidth]{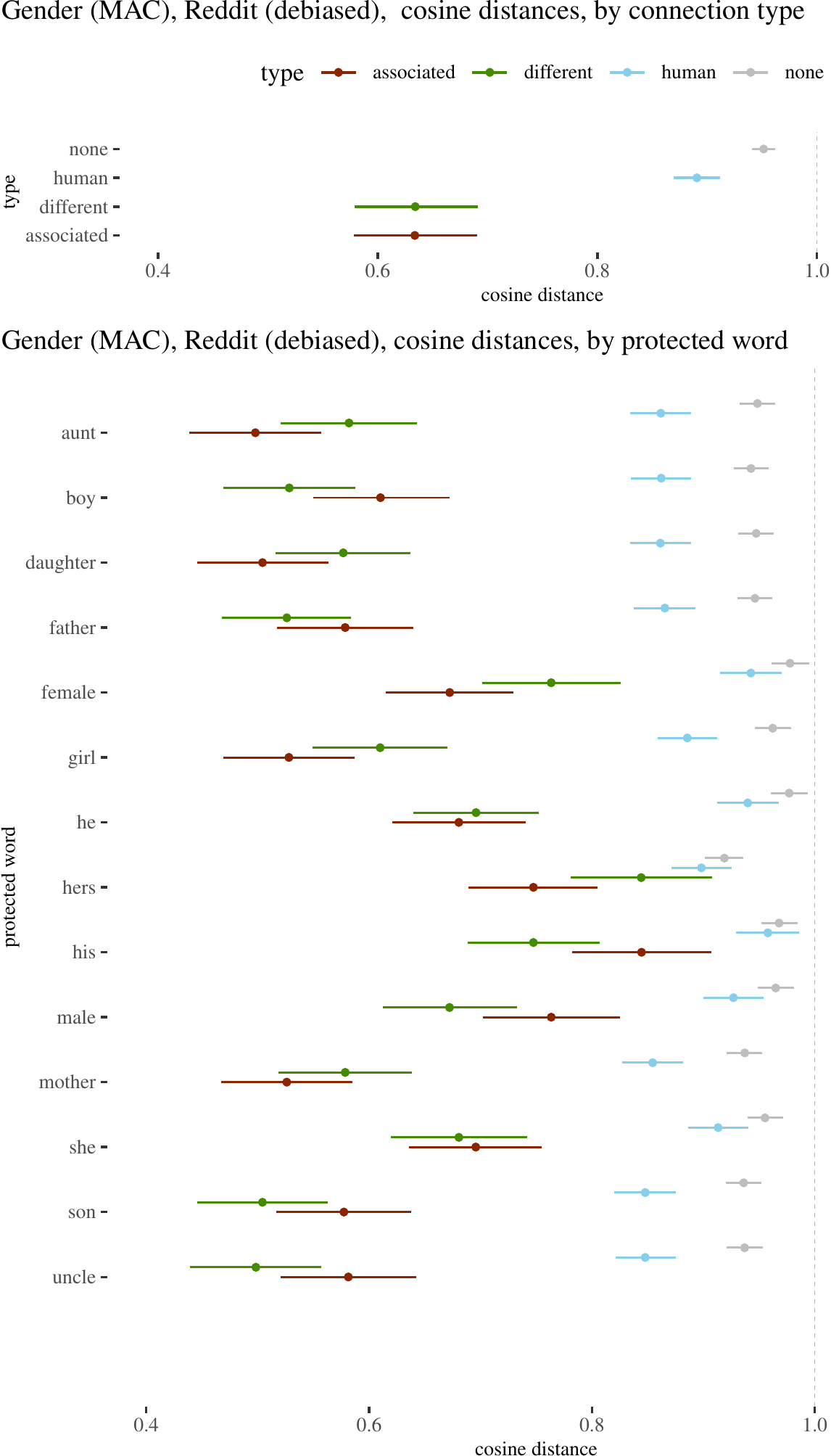} \end{center}

\hypertarget{posterior-predictive-checks}{%
\subsection{Posterior predictive
checks}\label{posterior-predictive-checks}}

\label{appendix:posterior}

\begin{center}\includegraphics[width=0.85\linewidth,height=1\textheight]{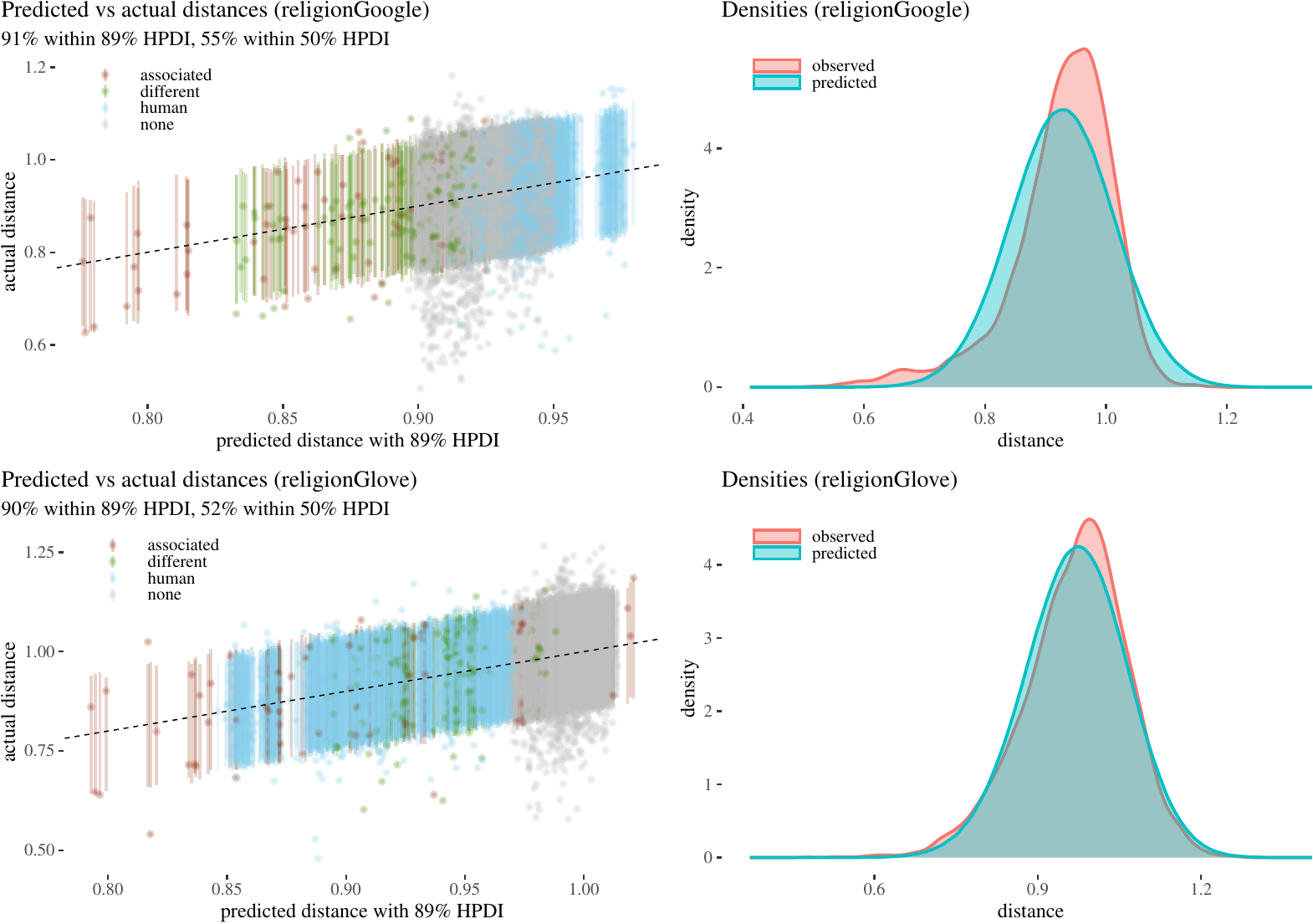} \end{center}

\begin{center}\includegraphics[width=0.85\linewidth,height=1\textheight]{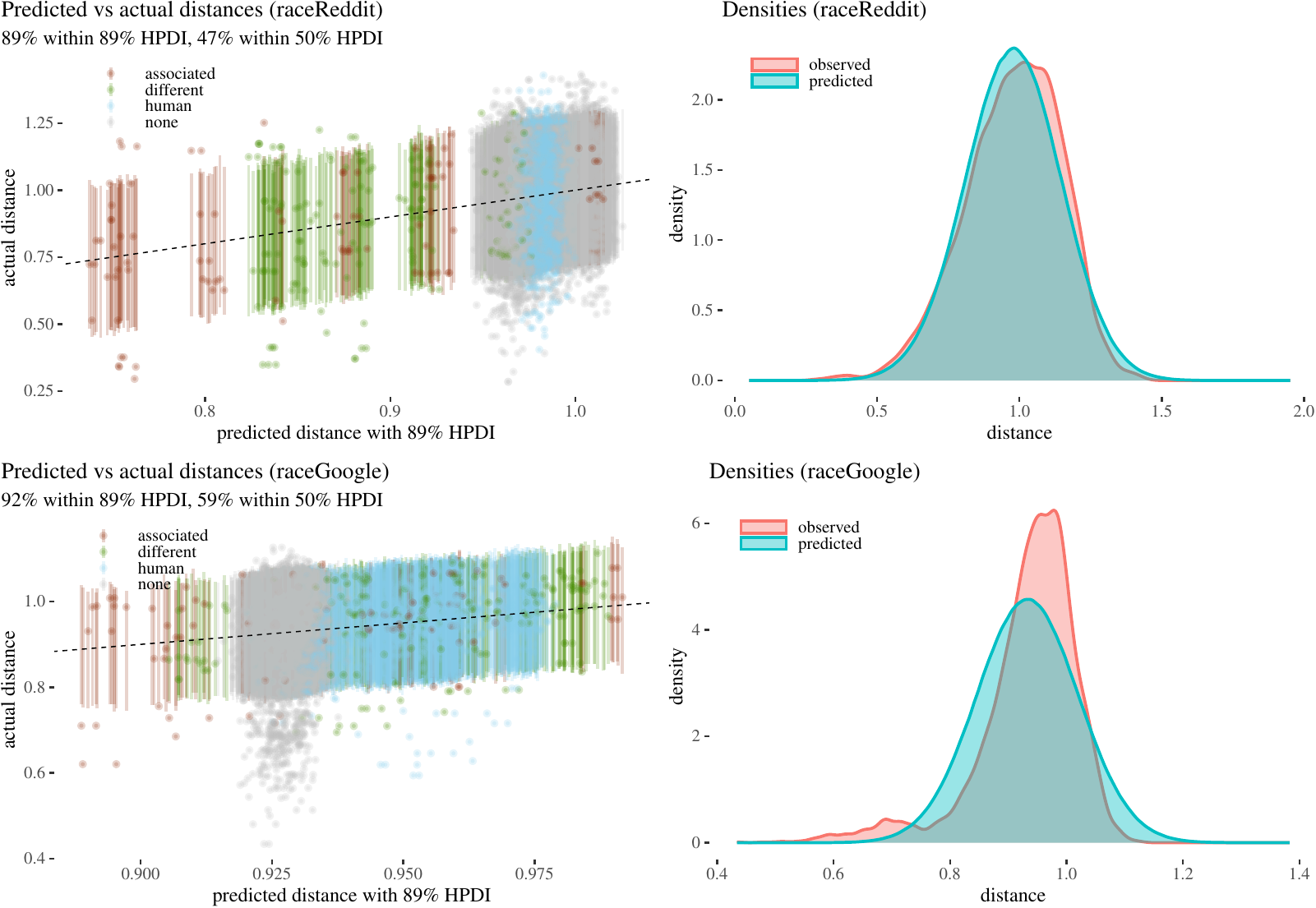} \end{center}

\begin{center}\includegraphics[width=0.87\linewidth,height=1\textheight]{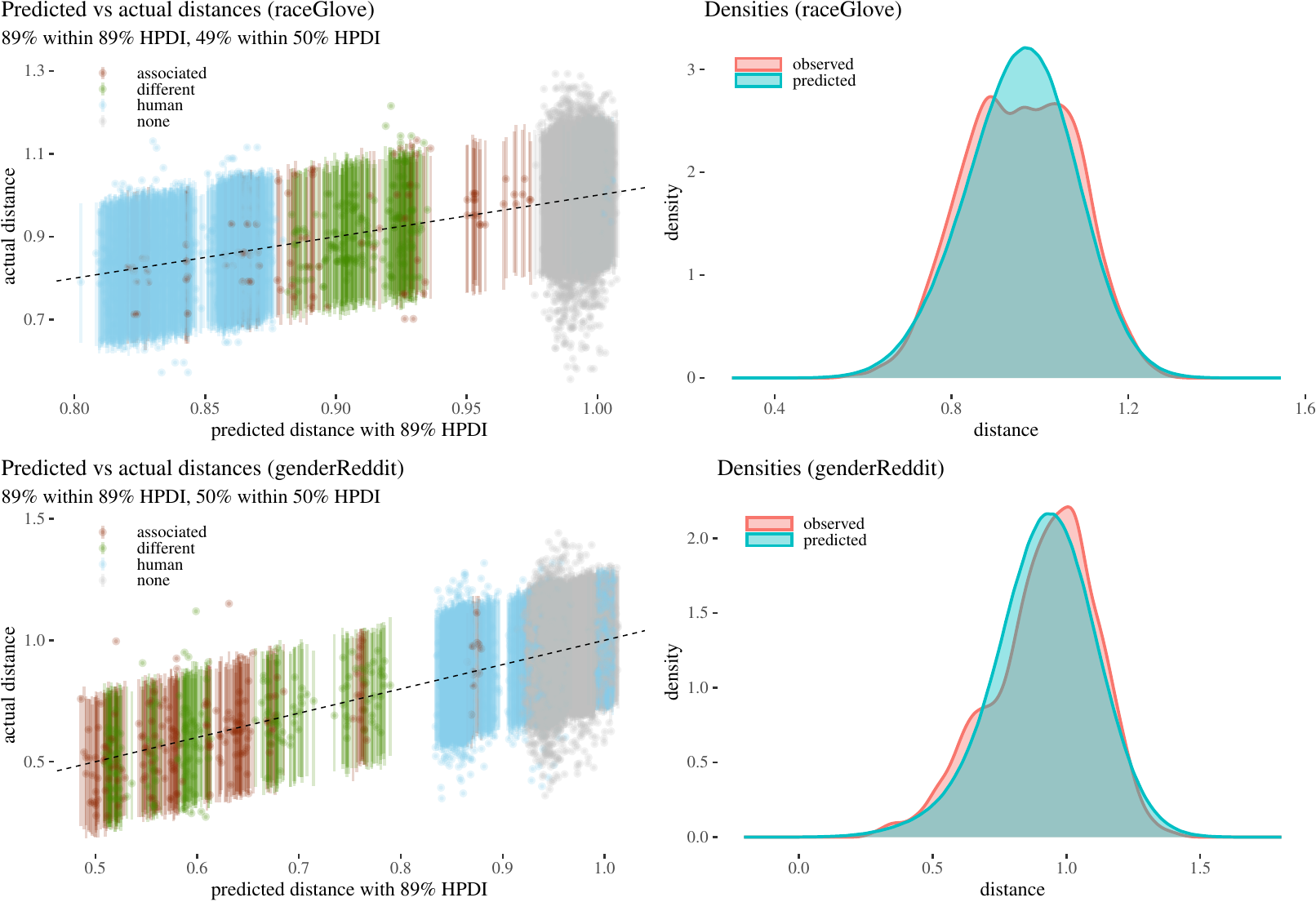} \end{center}

\begin{center}\includegraphics[width=0.89\linewidth,height=1\textheight]{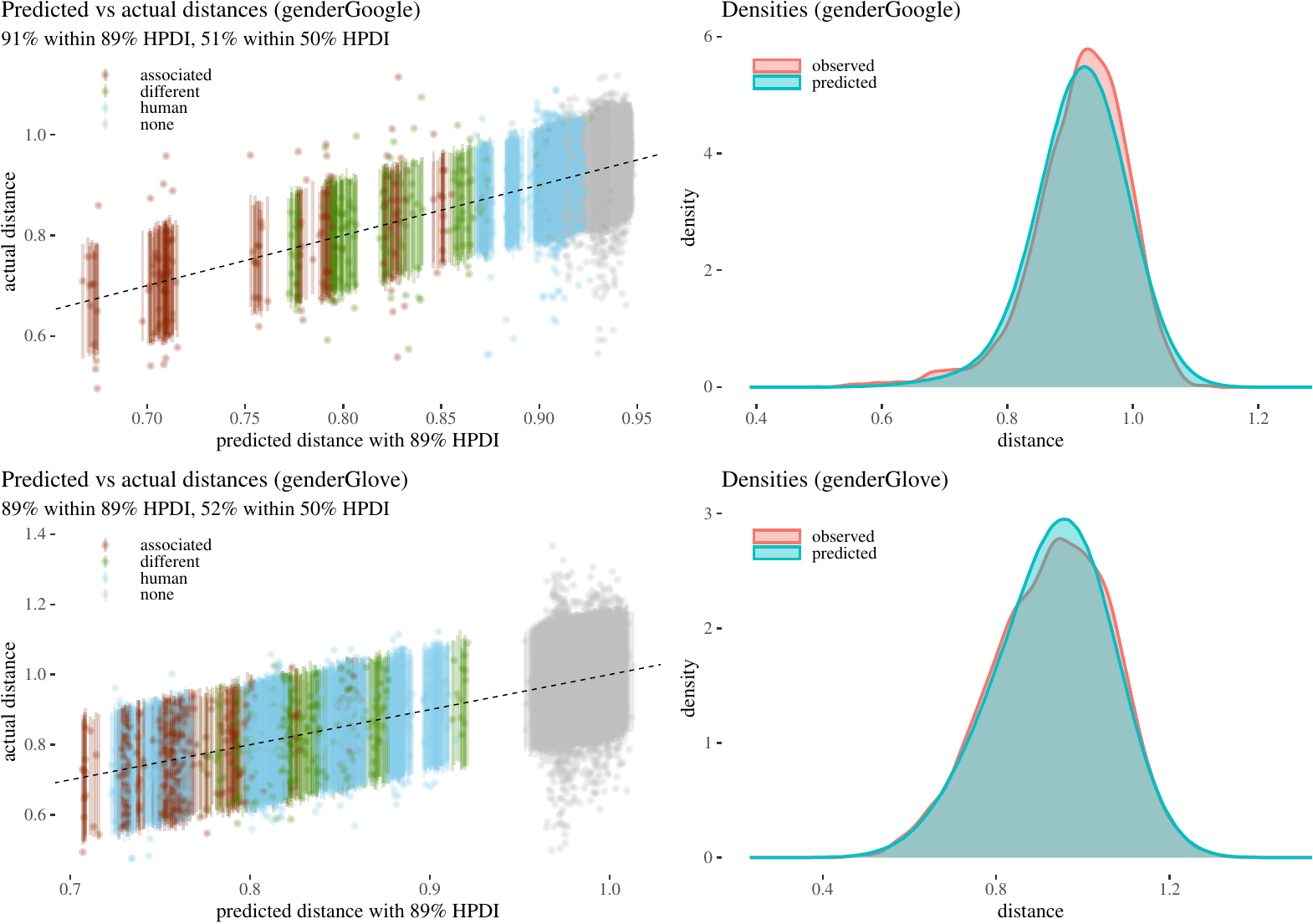} \end{center}

\hypertarget{word-lists}{%
\subsection{Word lists}\label{word-lists}}

\label{appendix:word}

\hypertarget{lists-used-in-previous-research}{%
\subsubsection{Lists used in previous
research}\label{lists-used-in-previous-research}}

\label{appendix:manzini_word_lists} The lists from {[}13{]} are
available here:

\begin{itemize}
\item
  \textbf{Gender:}
  \url{https://github.com/TManzini/DebiasMulticlassWordEmbedding/blob/master/Debiasing/data/vocab/gender_attributes_optm.json}
\item
  \textbf{Race:}
  \url{https://github.com/TManzini/DebiasMulticlassWordEmbedding/blob/master/Debiasing/data/vocab/race_attributes_optm.json}
\item
  \textbf{Religion:}
  \url{https://github.com/TManzini/DebiasMulticlassWordEmbedding/blob/master/Debiasing/data/vocab/religion_attributes_optm.json}
\end{itemize}

The lists are:

\begin{itemize}
\tightlist
\item
  \textbf{Gender:}
\end{itemize}

Pairs of protected attributes: {[}``he'', ``she''{]}, {[}``his'',
``hers''{]}, {[}``son'', ``daughter''{]}, {[}``father'', ``mother''{]},
{[}``male'', ``female''{]}, {[}``boy'', ``girl''{]}, {[}``uncle'',
``aunt''{]}

Lists of stereotypes per gender: man: {[}``manager'', ``executive'',
``doctor'', ``lawyer'', ``programmer'', ``scientist'', ``soldier'',
``supervisor'', ``rancher'', ``janitor'', ``firefighter'',
``officer''{]}, woman: {[}``secretary'', ``nurse'', ``clerk'',
``artist'', ``homemaker'', ``dancer'', ``singer'', ``librarian'',
``maid'', ``hairdresser'', ``stylist'', ``receptionist'',
``counselor''{]} \}

\begin{itemize}
\tightlist
\item
  \textbf{Race:}
\end{itemize}

Sets of protected attributes: {[}``black'', ``caucasian'', ``asian''{]},
{[}``african'', ``caucasian'', ``asian''{]}, {[}``black'', ``white'',
``asian''{]}, {[}``africa'', ``america'', ``asia''{]}, {[}``africa'',
``america'', ``china''{]}, {[}``africa'', ``europe'', ``asia''{]}

Lists of stereotypes per race: ``caucasian'': {[}``manager'',
``executive'', ``redneck'', ``hillbilly'', ``leader'', ``farmer''{]},
``asian'' : {[}``doctor'', ``engineer'', ``laborer'', ``teacher''{]},
``black'' : {[}``slave'', ``musician'', ``runner'', ``criminal'',
``homeless''{]}

\begin{itemize}
\tightlist
\item
  \textbf{Religion:}
\end{itemize}

Sets of protected attributes: {[}``judaism'', ``christianity'',
``islam''{]}, {[}``jew'', ``christian'', ``muslim''{]},
{[}``synagogue'', ``church'', ``mosque''{]}, {[}``torah'', ``bible'',
``quran''{]}, {[}``rabbi'', ``priest'', ``imam''{]}

Lists of stereotypes per race: ``jew'' : {[}``greedy'', ``cheap'',
``hairy'', ``liberal''{]}, ``christian'' : {[}``judgemental'',
``conservative'', ``familial''{]}, ``muslim'' : {[}``violent'',
``terrorist'', ``dirty'', ``uneducated''{]}

\hypertarget{custom-lists-used-in-this-paper}{%
\subsubsection{\texorpdfstring{Custom lists used in this paper
\label{app:custom}}{Custom lists used in this paper }}\label{custom-lists-used-in-this-paper}}

\begin{itemize}
\tightlist
\item
  \textbf{Neutral:}
\end{itemize}

{[}`ballpark', `glitchy', `billy', `dallas', `rip', `called',
`outlooks', `floater', `rattlesnake', `exports', `recursion',
`shortfall', `corrected', `solutions', `diagnostic', `patently',
`flops', `approx', `percents', `lox', `hamburger', `engulfed',
`households', `north', `playtest', `replayability', `glottal',
`parable', `gingers', `anachronism', `organizing', `reach', `shtick',
`eleventh', `cpu', `ranked', `irreversibly', `ponce', `velociraptor',
`defects', `puzzle', `smasher', `northside', `heft', `observation',
`rectum', `mystical', `telltale', `remnants', `inquiry', `indisputable',
`boatload', `lessening', `uselessness', `observes', `fictitious',
`repatriation', `duh', `attic', `schilling', `charges', `chatter',
`pad', `smurfing', `worthiness', `definitive', `neat', `homogenized',
`lexicon', `nationalized', `earpiece', `specializations', `lapse',
`concludes', `weaving', `apprentices', `fri', `militias',
`inscriptions', `gouda', `lift', `laboring', `adaptive', `lecture',
`hogging', `thorne', `fud', `skews', `epistles', `tagging', `crud',
`two', `rebalanced', `payroll', `damned', `approve', `reason',
`formally', `releasing', `muddled', `mineral', `shied', `capital',
`nodded', `escrow', `disconnecting', `marshals', `winamp', `forceful',
`lowes', `sip', `pencils', `stomachs', `goff', `cg', `backyard',
`uprooting', `merging', `helpful', `eid', `trenchcoat', `airlift',
`frothing', `pulls', `volta', `guinness', `viewership', `eruption',
`peeves', `goat', `goofy', `disbanding', `relented', `ratings',
`disputed', `vitamins', `singled', `hydroxide', `telegraphed',
`mercantile', `headache', `muppets', `petal', `arrange', `donovan',
`scrutinized', `spoil', `examiner', `ironed', `maia', `condensation',
`receipt', `solider', `tattooing', `encoded', `compartmentalize',
`lain', `gov', `printers', `hiked', `resentment', `revisionism',
`tavern', `backpacking', `pestering', `acknowledges', `testimonies',
`parlance', `hallucinate', `speeches', `engaging', `solder',
`perceptive', `microbiology', `reconnaissance', `garlic', `neutrals',
`width', `literaly', `guild', `despicable', `dion', `option',
`transistors', `chiropractic', `tattered', `consolidating', `olds',
`garmin', `shift', `granted', `intramural', `allie', `cylinders',
`wishlist', `crank', `wrongly', `workshop', `yesterday', `wooden',
`without', `wheel', `weather', `watch', `version', `usually', `twice',
`tomato', `ticket', `text', `switch', `studio', `stick', `soup',
`sometimes', `signal', `prior', `plant', `photo', `path', `park',
`near', `menu', `latter', `grass', `clock'{]}

\begin{itemize}
\tightlist
\item
  \textbf{Human-related:}
\end{itemize}

{[}`wear', `walk', `visitor', `toy', `tissue', `throw', `talk', `sleep',
`eye', `enjoy', `blogger', `character', `candidate', `breakfast',
`supper', `dinner', `eat', `drink', ``carry'', ``run'', ``cast'',
``ask'', ``awake'', ``ear'', ``nose'', ``lunch'', ``coalition'',
``policies'', ``restaurant'', ``stood'', ``assumed'', ``attend'',
``swimming'', ``trip'', ``door'', ``determine'', ``gets'', ``leg'',
``arrival'', ``translated'', ``eyes'', ``step'', ``whilst'',
``translation'', ``practices'', ``measure'', ``storage'', ``window'',
``journey'', ``interested'', ``tries'', ``suggests'', ``allied'',
``cinema'', ``finding'', ``restoration'', ``expression'',``visitors'',
``tell'', ``visiting'', ``appointment'', ``adults'', ``bringing'',
``camera'', ``deaths'', ``filmed'', ``annually'', ``plane'', ``speak'',
``meetings'', ``arm'', ``speaking'', ``touring'', ``weekend'',
``accept'', ``describe'', ``everyone'', ``ready'', ``recovered'',
``birthday'', ``seeing'', ``steps'', ``indicate'', ``anyone'',
``youtube''{]}

\end{document}